\documentclass[letterpaper,10pt,conference]{ieeeconf}
\usepackage[T1]{fontenc}
\usepackage{graphicx}
\usepackage{xcolor}
\usepackage{amsmath,amssymb}
\usepackage{array,booktabs,colortbl}
\usepackage{url}
\usepackage{placeins}
\usepackage{multicol}
\usepackage[hidelinks,unicode]{hyperref}
\IEEEoverridecommandlockouts
\definecolor{reportblue}{RGB}{35,78,119}
\definecolor{tablehead}{RGB}{233,240,246}
\definecolor{tablerow}{RGB}{246,249,251}
\definecolor{reviewyellow}{RGB}{255,245,191}
\definecolor{reviewred}{RGB}{255,222,218}
\colorlet{reviewyellow}{white}
\DeclareFontShape{T1}{ptm}{b}{it}{<->ssub * ptm/b/n}{}
\DeclareFontShape{T1}{ptm}{bx}{it}{<->ssub * ptm/b/n}{}
\newlength{\reportfigurebefore}
\newlength{\reportfigureafter}
\newlength{\reporttablebefore}
\newlength{\reporttableafter}
\makeatletter
\newif\ifreportinflow
\let\reportoriginalmakecaption\@makecaption
\long\def\@makecaption#1#2{\reportoriginalmakecaption{#1}{#2}%
\ifx\@captype\@IEEEtablestring\else\ifreportinflow\else\par\vspace{3pt}\fi\fi}
\newsavebox{\reportblockbox}
\newcounter{reportblocknumber}
\newcommand{\reportneedspace}[1]{\par\begingroup\dimen@=\pagegoal\advance\dimen@ by -\pagetotal\ifdim\dimen@<#1\relax\newpage\fi\endgroup}
\newenvironment{reportfigure}{\par\addvspace{\reportfigurebefore}\stepcounter{reportblocknumber}\begin{lrbox}{\reportblockbox}\begin{minipage}{\textwidth}\centering\reportinflowtrue\def\@captype{figure}}{\end{minipage}\end{lrbox}\typeout{REPORTBLOCK:\arabic{reportblocknumber}:\the\dimexpr\ht\reportblockbox+\dp\reportblockbox\relax}\noindent\usebox{\reportblockbox}\par\addvspace{\reportfigureafter}}
\newenvironment{reporttable}{\par\addvspace{\reporttablebefore}\stepcounter{reportblocknumber}\begin{lrbox}{\reportblockbox}\begin{minipage}{\textwidth}\centering\def\@captype{table}}{\end{minipage}\end{lrbox}\typeout{REPORTBLOCK:\arabic{reportblocknumber}:\the\dimexpr\ht\reportblockbox+\dp\reportblockbox\relax}\noindent\usebox{\reportblockbox}\par\addvspace{\reporttableafter}}
\makeatother
\title{\LARGE\bfseries Genie Sim Benchmark:\\Real-to-Sim Fidelity and Evaluation\\[4pt]
{\normalsize\mdseries Data-Driven Simulation Fidelity and Real-World Robot Evaluation}}
\author{\parbox{\textwidth}{\centering\normalsize Di Yang$^{*}$\quad Da Huang$^{*}$\quad Yang Lv$^{*}$\quad Fan Yang\quad Yi Lu\quad Zheyuan Xing\\[2pt]Linting Guo\quad Jia'en Shen\quad Wenjun Sun\quad Maoqing Yao$^{\dagger}$\quad Qian Wang$^{\dagger}$\\[5pt]
\scriptsize * First authors\quad $\dagger$ Corresponding authors}}
\makeatletter
\newcommand{\reportopening}{%
\begin{minipage}{\textwidth}
\begin{abstract}
We evaluate Genie Sim Benchmark through paired real--simulation tasks and complementary system-identification studies. Across 22 robot--task pairs spanning 16 task names,\allowbreak{} task-specific \ensuremath{\pi_{0.5}} policies fine-tuned on real-robot data achieve a descriptive Score correlation of 0.9686 \allowbreak{}(MAE 0.0405; RMSE 0.0520),\allowbreak{} with simulation--real differences from \ensuremath{-}0.110 to +0.080. G2 crsB experiments assess long-horizon closed-loop position and joint-side torque agreement,\allowbreak{} while five representative object cases examine identification and cross-engine contact transfer. A separate 32.6503 s teleoperation trajectory,\allowbreak{} excluded from mechanical identification and servo-gain search,\allowbreak{} evaluates external position targets through simulated motion-controller command shaping and a calibrated position servo. Across both arms,\allowbreak{} per-joint position RMSE spans 0.00614--0.03522 rad and current-derived torque RMSE spans 0.73545--4.00099 N\ensuremath{\cdot}m. Native PhysX adaptations recover three recorded interaction tasks,\allowbreak{} with case-dependent alignment gains. These results support task-specific simulation evaluation through joint assessment of task outcomes and physical-response consistency.
\end{abstract}
\par\vspace{6pt}
\begin{minipage}{\textwidth}
\makeatletter\def\@captype{figure}\makeatother
\centering
\renewcommand{\thefigure}{1}
\makebox[.48\linewidth][l]{\small\bfseries\color{reportblue}Robot-level consistency}\hfill\makebox[.48\linewidth][l]{\small\bfseries\color{reportblue}Benchmark correspondence}\par\smallskip

\includegraphics[width=.48\linewidth,height=.43\textheight,keepaspectratio]{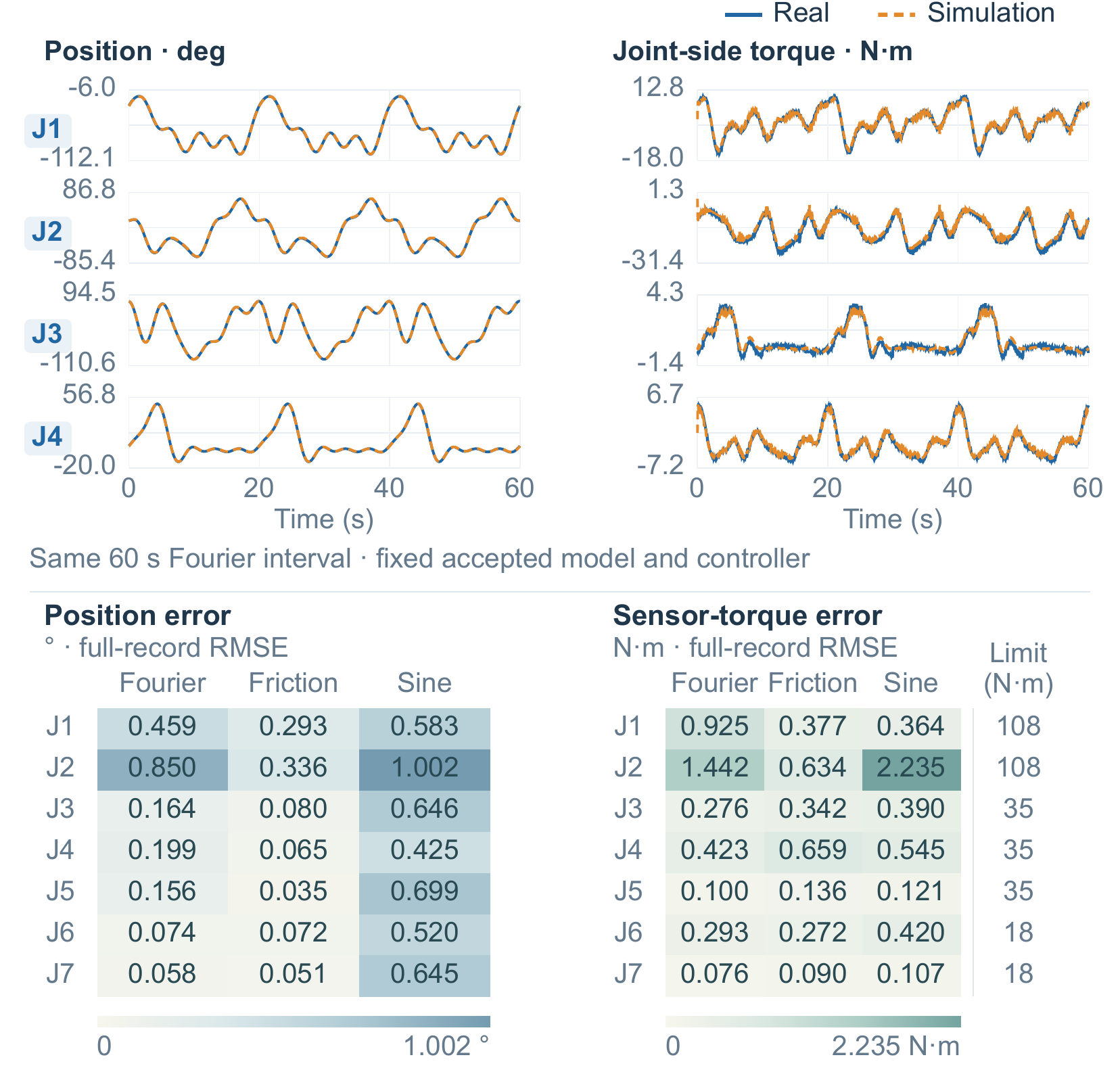}\hfill
\includegraphics[width=.48\linewidth,height=.43\textheight,keepaspectratio]{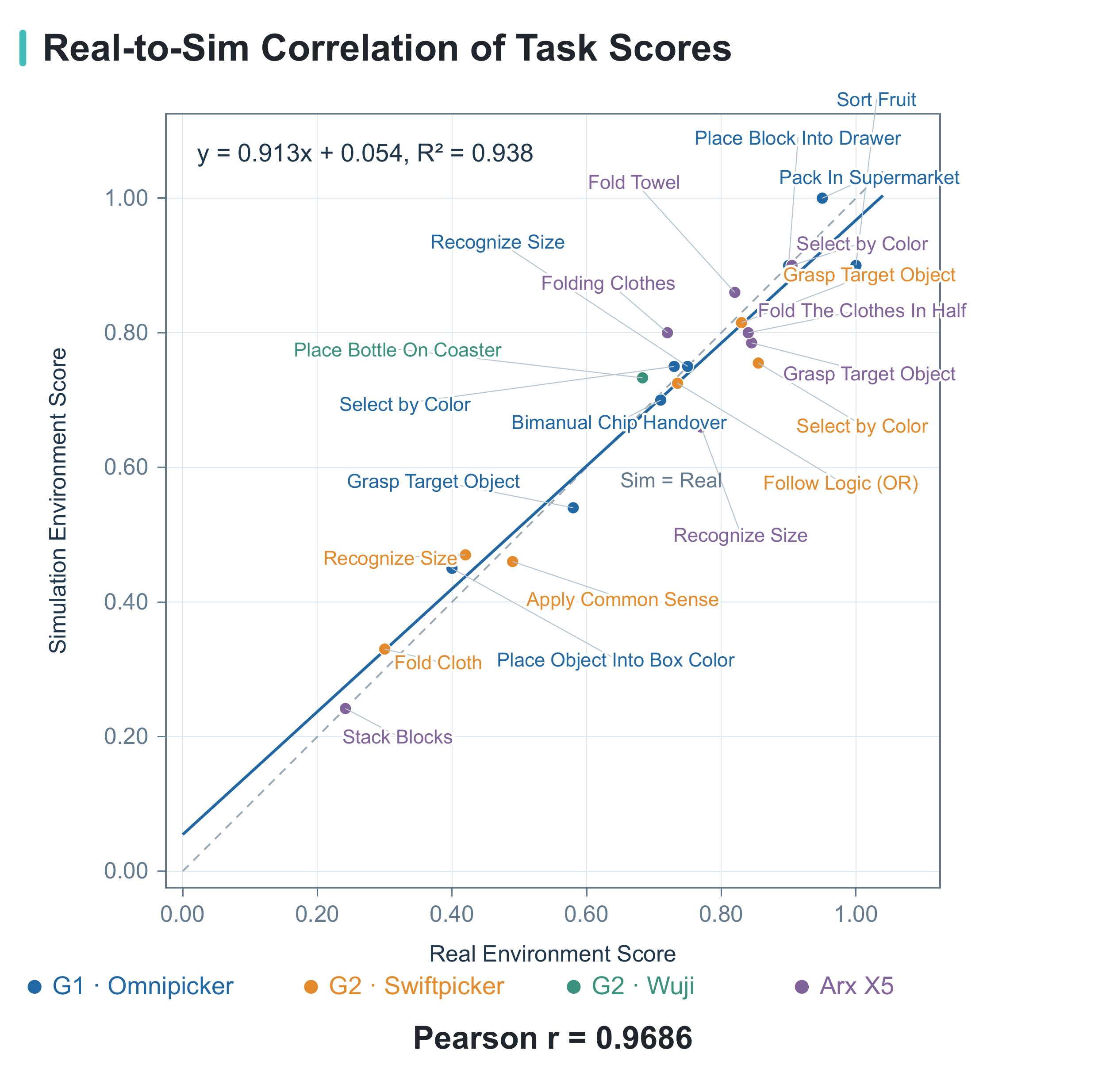}
\caption{Robot- and task-level real--sim evidence. Left: 60 s J1--J4 closed-loop replay and seven-joint RMSE across three excitation families; errors are uncentred and torque limits are URDF references,\allowbreak{} not hardware ratings \allowbreak{}(\hyperref[sec:4.1]{Section~\ref*{sec:4.1}}; \hyperref[fig:5]{Figure~5}). Right: correspondence across 22 final task-Score pairs. Pearson r = 0.9686,\allowbreak{} MAE 0.0405 and RMSE 0.0520 \allowbreak{}(\hyperref[tab:B1]{Table~B1}). Before/after comparisons: \hyperref[tab:B2]{Table~B2}.}
\label{fig:1}
\phantomsection\label{html:benchmark-scatter}
\end{minipage}
\end{minipage}}
\IEEEaftertitletext{\reportopening}
\makeatother
\newcommand{\reportnotePageSeven}{%
\par\addvspace{6pt}\phantomsection\label{formal:reading-07}
\noindent\colorbox{reviewyellow}{\parbox{\dimexpr\linewidth-2\fboxsep\relax}{\normalsize\fontsize{10}{12.5}\selectfont \textbf{Reading the parameter groups.} These parameter groups describe complementary contributions to the simulated execution chain. The regressor represents configuration-dependent link dynamics,\allowbreak{} armature represents acceleration-proportional drive effects,\allowbreak{} and friction represents velocity-dependent resistance. Keeping them separate makes the estimated model directly usable for replay and control. Fourier reconstruction inspects their combined response during coupled motion,\allowbreak{} while the independent teleoperation recording evaluates identified mechanics within the calibrated command-to-motion chain.}}
\par\addvspace{7pt}}

\newcommand{\reportnotePageNine}{%
\par\addvspace{6pt}\phantomsection\label{formal:reading-09}
\noindent\colorbox{reviewyellow}{\parbox{\dimexpr\linewidth-2\fboxsep\relax}{\normalsize\fontsize{10}{12.5}\selectfont \textbf{Reading the evidence together.} The paired task evaluation measures correspondence of outcomes,\allowbreak{} while the robot and object studies expose the physical responses behind those outcomes. Their complementary views make simulation agreement inspectable at the task,\allowbreak{} motion and contact levels,\allowbreak{} linking benchmark results to concrete execution and interaction evidence.}}
\par\addvspace{7pt}}

\newcommand{\reportnotePageThirteen}{%
\par\addvspace{6pt}\phantomsection\label{formal:reading-13}
\noindent\colorbox{reviewyellow}{\parbox{\dimexpr\linewidth-2\fboxsep\relax}{\normalsize\fontsize{10}{13}\selectfont \textbf{Joint and excitation-family views}\par\vspace{3pt}{\setlength{\multicolsep}{0pt}\setlength{\columnsep}{14pt}\begin{multicols}{2}\raggedcolumns The two tables organize the identification archive along complementary axes. The per-joint view makes the distribution of residual error across the arm visible,\allowbreak{} whereas the excitation-family view shows how that distribution changes with the imposed motion. Friction sweeps,\allowbreak{} sine excitation and coupled Fourier motion therefore remain separate entries instead of being collapsed into a single apparent accuracy. The RMS and R\textsuperscript{2} columns are read together: one measures error in torque units,\allowbreak{} while the other describes agreement relative to the variation in the recorded signal. Maximum error further identifies departures that an average can soften. Together,\allowbreak{} these views connect the compact parameter summary in the main text to the complete recording-level evidence behind the following diagnostic figures.\end{multicols}}}}
\par\addvspace{7pt}}

\newcommand{\reportnotePageFourteen}{%
\par\addvspace{6pt}\phantomsection\label{formal:reading-14}
\noindent\colorbox{reviewyellow}{\parbox{\dimexpr\linewidth-2\fboxsep\relax}{\normalsize\fontsize{10}{13}\selectfont \textbf{Interpreting the recording overlays}\par\vspace{3pt}{\setlength{\multicolsep}{0pt}\setlength{\columnsep}{14pt}\begin{multicols}{2}\raggedcolumns Each panel pairs the recorded and identified-model torque over the displayed source interval. Their common time axis makes repeated periods,\allowbreak{} reversals and waveform shape directly comparable. The panels retain the driven joints and background-posture variants used by the source experiments,\allowbreak{} so the reader can inspect a collection of motions rather than one selected instant. The overlays and their numeric annotations answer related questions: the centred curves expose temporal structure,\allowbreak{} while the Own RMS labels retain the scoring convention stated in the caption. Reading both avoids attributing a constant display alignment to improved absolute torque calibration. All shown joints and the complete metrics remain part of the comparison.\end{multicols}}}}
\par\addvspace{7pt}}

\newcommand{\reportnotePageFifteen}{%
\par\addvspace{6pt}\phantomsection\label{formal:reading-15}
\noindent\colorbox{reviewyellow}{\parbox{\dimexpr\linewidth-2\fboxsep\relax}{\normalsize\fontsize{10}{13}\selectfont \textbf{Connecting configuration and response.} \hyperref[tab:C3]{Table~C3} and \hyperref[fig:C4]{Figure~C4} form a paired view of this execution chain: the table retains each arm's gains and errors,\allowbreak{} while the curves show when error occurs. The recorded inputs and accepted settings remain fixed throughout the replay.}}
\par\addvspace{7pt}}

\newcommand{\reportnotePageNineteen}{%
\par\addvspace{6pt}\phantomsection\label{formal:reading-19}
\noindent\colorbox{reviewyellow}{\parbox{\dimexpr\linewidth-2\fboxsep\relax}{\normalsize\fontsize{10}{13}\selectfont \textbf{Reading the search protocol.} The objectives use different observation channels,\allowbreak{} so their magnitudes are interpreted within each case. Compare each candidate with its own baseline under the stated configuration. This keeps contact-model adaptation,\allowbreak{} trajectory alignment and completion checks distinct while retaining a common route from observation to replay.}}
\par\addvspace{7pt}}

\newcommand{\reportnotePageTwenty}{%
\par\addvspace{6pt}\phantomsection\label{formal:reading-20}
\noindent\colorbox{reviewyellow}{\parbox{\dimexpr\linewidth-2\fboxsep\relax}{\normalsize\fontsize{10}{13}\selectfont \textbf{Connecting alignment and observation coverage}\par\vspace{3pt}{\setlength{\multicolsep}{0pt}\setlength{\columnsep}{14pt}\begin{multicols}{2}\raggedcolumns The coverage table identifies which observations support each trajectory metric. Dense command streams drive complete replays,\allowbreak{} while object error is evaluated where the recorded feature is available. Missing depth for most shuttlecock annotations therefore does not become a fabricated three-dimensional path. Keeping observation counts next to replay durations makes the measurement basis visible. Across engines,\allowbreak{} completion and feature alignment remain separate evidence: a model can execute the interaction while following a different object path. Reading the tables together preserves both outcomes and gives the trajectory figures a concrete source,\allowbreak{} time range and unit.\end{multicols}}}}
\par\addvspace{7pt}}

\newcommand{\reportnotePageTwentyFour}{%
\par\addvspace{6pt}\phantomsection\label{formal:reading-24}
\noindent\colorbox{reviewyellow}{\parbox{\dimexpr\linewidth-2\fboxsep\relax}{\normalsize\fontsize{10}{13}\selectfont \textbf{Reading recovery alongside alignment}\par\vspace{3pt}{\setlength{\multicolsep}{0pt}\setlength{\columnsep}{14pt}\begin{multicols}{2}\raggedcolumns The trajectory panels make a distinction that a completion label alone cannot show. With shared native settings,\allowbreak{} both parameter sets can carry out a handover,\allowbreak{} while their predicted feature paths remain distinguishable. The real points provide the same visual anchor for judging each path; residual panels retain errors only at the available annotations. In particular,\allowbreak{} smooth simulated curves do not increase the number of real observations. The supplementary matrices that follow widen this view from individual trajectories to the recorded case--stack comparisons. They preserve increases as well as reductions in error and keep the original transfer study separate from the adapted native comparisons. Together,\allowbreak{} these presentations connect the recovered interaction to its measured path instead of reducing the result to a single success indicator.\end{multicols}}}}
\par\addvspace{7pt}}

\newcommand{\reportnotePageTwentySix}{%
\par\addvspace{6pt}\phantomsection\label{formal:reading-26}
\noindent\colorbox{reviewyellow}{\parbox{\dimexpr\linewidth-2\fboxsep\relax}{\normalsize\fontsize{10}{13}\selectfont \textbf{Cross-reading the two matrices.} Cross-read the matrices by object,\allowbreak{} retaining each panel's units and phase. Their complementary view distinguishes gripper feedback from object-path agreement and keeps unsuccessful contacts visible even when an angle-based error is small.}}
\par\addvspace{7pt}}

\begin{document}
\maketitle
\thispagestyle{plain}
\pagestyle{plain}
\phantomsection\label{html:benchmark-overview}
\section*{Real--Sim Consistency Across Benchmark Tasks}

Genie Sim Benchmark connects simulation evaluation with real-robot performance through paired manipulation tasks. Across 22 robot--task pairs spanning 16 task names,\allowbreak{} the evaluated Scores show strong real--simulation correspondence \allowbreak{}(Pearson r = 0.9686),\allowbreak{} with signed simulation--real differences from \ensuremath{-}0.110 to +0.080. Together with robot- and object-level physical validation,\allowbreak{} these results establish an evidence-led basis for assessing simulation fidelity at both the task and interaction levels.

\phantomsection\label{html:benchmark-validation-objective}
The report begins with this task-level correspondence,\allowbreak{} then develops the Real2Sim pipeline and system-identification methods that support it. \hyperref[sec:2]{Section~\ref*{sec:2}} details the evaluation protocol and task coverage; Sections~\ref{sec:3}--\ref{sec:4} examine execution and contact consistency. Complete Scores,\allowbreak{} identification comparisons and test quantities are retained in \hyperref[tab:B1]{Tables~B1}--B2.

\begin{table*}[!t]
\centering
\renewcommand{\thetable}{1}
\caption{Real2Sim evaluation coverage for the four robot configurations illustrated in \hyperref[fig:2]{Figure~2}. Tasks are distinct evaluated task names; robot columns count completed tasks for each configuration. \hyperref[fig:2]{Figure~2} presents representative paired scenes.}
\label{tab:1}
{\fontsize{8.2}{10.1}\selectfont
\setlength{\tabcolsep}{3pt}
\renewcommand{\arraystretch}{1.13}
\begin{tabular}{>{\raggedright\arraybackslash}p{\dimexpr 0.2900\linewidth-2\tabcolsep\relax}>{\raggedright\arraybackslash}p{\dimexpr 0.1500\linewidth-2\tabcolsep\relax}>{\raggedright\arraybackslash}p{\dimexpr 0.1400\linewidth-2\tabcolsep\relax}>{\raggedright\arraybackslash}p{\dimexpr 0.1400\linewidth-2\tabcolsep\relax}>{\raggedright\arraybackslash}p{\dimexpr 0.1400\linewidth-2\tabcolsep\relax}>{\raggedright\arraybackslash}p{\dimexpr 0.1400\linewidth-2\tabcolsep\relax}}
\toprule
\rowcolor{tablehead}\textbf{Task family} & \textbf{Tasks} & \textbf{G1 Omnipicker} & \textbf{G2 Swiftpicker} & \textbf{G2 Wuji} & \textbf{Arx X5} \\
\midrule
Selection \& reasoning & 5 & 3 & 5 & 0 & 3 \\
\rowcolor{tablerow}Placement \& sorting & 5 & 4 & 0 & 1 & 0 \\
Bimanual handover & 1 & 1 & 0 & 0 & 0 \\
\rowcolor{tablerow}Stacking & 1 & 0 & 0 & 0 & 1 \\
Textile manipulation & 4 & 0 & 1 & 0 & 3 \\
\bottomrule
\end{tabular}
}
\end{table*}

\phantomsection\label{html:introduction}
\section{Introduction}
\label{sec:1}

A simulation benchmark is useful only when its task outcomes remain informative about real robots.

Matching joint and gripper behavior addresses only one part of the reality gap. Object geometry,\allowbreak{} contact response,\allowbreak{} rotational resistance and initial state can determine whether a grasp,\allowbreak{} handover or placement succeeds. We organize the report around Genie Sim Benchmark: which task families are covered,\allowbreak{} which execution and interaction errors can change their outcomes,\allowbreak{} and whether frozen real--simulation evaluations preserve task scores and failure stages. \hyperref[sec:2]{Section~\ref*{sec:2}} presents benchmark coverage,\allowbreak{} the paired-evaluation protocol,\allowbreak{} reported scores and failure boundaries. Sections~\ref{sec:3} and~\ref{sec:4} then explain the Real2Sim pipeline and the robot/object identification methods supporting this evaluation.

We examine 22 final Score pairs and 14 G2/Arx before/after system-identification comparisons. Robot-identification replay and five G2 object recordings provide complementary evidence on execution and contact response. The evaluation compares real and simulated task Scores and physical responses,\allowbreak{} and examines how these discrepancies change after system identification \cite{ref1,ref4,ref5}.

\phantomsection\label{html:benchmark}
\section{Genie Sim Benchmark: Protocol,\allowbreak{} Coverage and Real--Sim Evaluation}
\label{sec:2}

\subsection{Genie Sim Benchmark: platform and capability dimensions}
\label{sec:2.1}
\phantomsection\label{html:genie-sim-benchmark-introduction}

\phantomsection\label{html:benchmark-positioning}
Genie Sim Benchmark is an embodied-model evaluation platform built around a systematic,\allowbreak{} multidimensional assessment framework \cite{ref27}. It uses high-fidelity simulation evaluation closely aligned with real-robot performance to help developers identify model strengths and weaknesses and guide continuous improvement. The platform reports 78 simulation tasks across four capability dimensions. Per-dimension results,\allowbreak{} completed substeps and failure-stage diagnostics help developers localize limitations and prioritize model and training-data improvements.

\paragraph{Platform capability dimensions}
\phantomsection\label{html:benchmark-capability-dimensions}

\textbf{Instruction Following.} Colour,\allowbreak{} quantity,\allowbreak{} shape,\allowbreak{} size,\allowbreak{} category,\allowbreak{} logic and common-sense instructions test whether a model selects the intended target and executes the requested action.

\textbf{Spatial Reasoning.} Absolute position,\allowbreak{} relative relations,\allowbreak{} size and ordinal ordering,\allowbreak{} stacking and spatial alignment test the interpretation of object arrangements.

\textbf{Robustness.} Changes in background,\allowbreak{} lighting,\allowbreak{} material,\allowbreak{} robot initial state,\allowbreak{} camera pose,\allowbreak{} image quality and instruction wording probe performance under perturbations.

\textbf{General Manipulation.} Door opening,\allowbreak{} pouring,\allowbreak{} scooping,\allowbreak{} stocking,\allowbreak{} sorting,\allowbreak{} table clearing and bimanual coordination test the composition of atomic skills in multi-step tasks.

\phantomsection\label{html:benchmark-platform-design}
The platform description organizes its 78 tasks into these four capability dimensions and records completed substeps and failure stages. It describes diverse samples,\allowbreak{} domain randomization,\allowbreak{} separate training and test sets,\allowbreak{} and in-distribution and out-of-distribution tests as evaluation design mechanisms.

\subsection{Real2Sim task selection and embodiment coverage}
\label{sec:2.2}

\phantomsection\label{html:benchmark-task-relation}
To examine real--simulation consistency across robot embodiments and end effectors,\allowbreak{} the evaluation covers G1 Omnipicker,\allowbreak{} G2 Swiftpicker,\allowbreak{} G2 Wuji and Arx X5.

\begin{figure*}[!t]
\centering
\renewcommand{\thefigure}{2}
\includegraphics[width=\linewidth,height=.42\textheight,keepaspectratio]{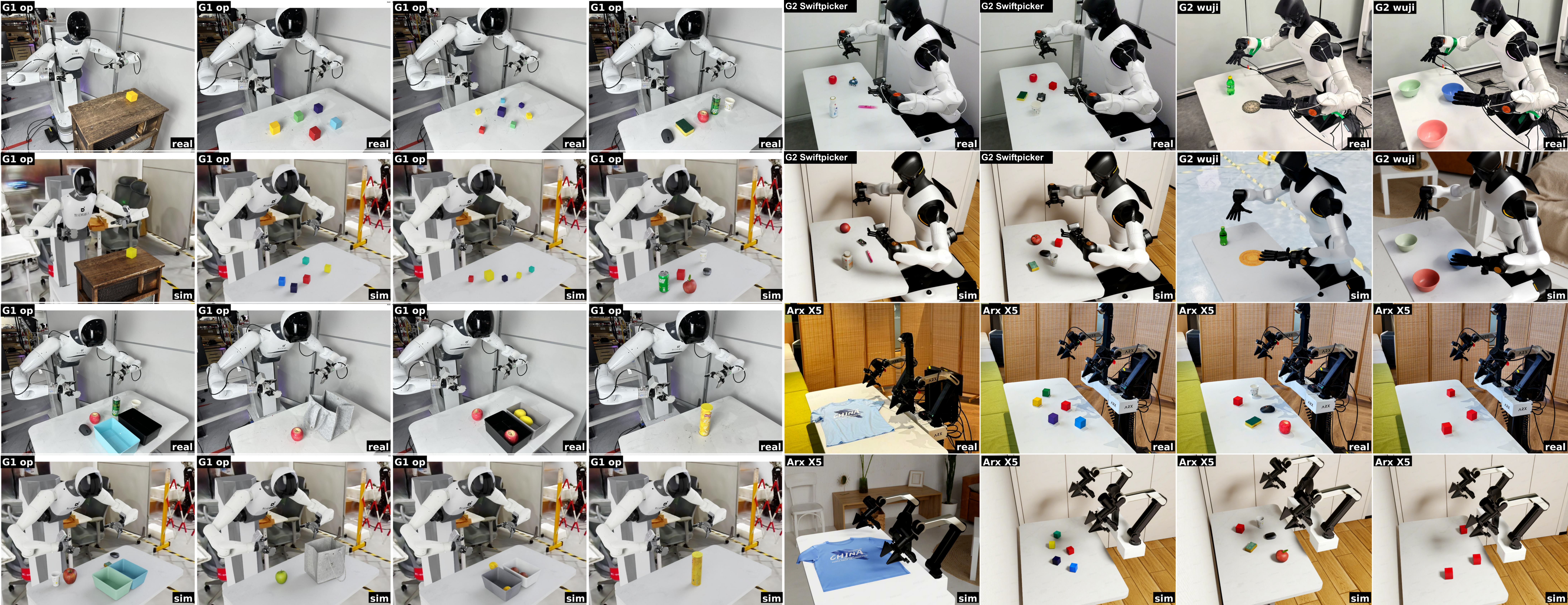}
\caption{Representative scenes from the Real2Sim experiments summarized in \hyperref[tab:1]{Table~1}: sixteen scene pairs across G1 Omnipicker,\allowbreak{} G2 Swiftpicker,\allowbreak{} G2 Wuji and Arx X5 configurations. Rows 1 and 3 are real scenes; rows 2 and 4 are their simulation counterparts. Corresponding columns preserve the paired object arrangements and interaction context.}
\label{fig:2}
\phantomsection\label{html:benchmark-paired-scenes}
\end{figure*}

\phantomsection\label{html:benchmark-task-selection}
The newly added G2 and Arx tasks cover rigid-object manipulation and deformable textile manipulation. \hyperref[fig:2]{Figure~2} presents representative real--simulation scene pairs for all four configurations,\allowbreak{} illustrating the robot embodiments,\allowbreak{} object arrangements and interaction contexts used in the experiments.

\phantomsection\label{html:benchmark-family-analysis}
The evaluation spans five task families. \textbf{Selection \& reasoning} tests instruction-conditioned target selection and grasp execution. \textbf{Placement \& sorting} tests object transport and release into containers,\allowbreak{} drawers or onto supports. \textbf{Bimanual handover} examines coordinated transfer between hands,\allowbreak{} while \textbf{Stacking} examines alignment and stable placement. \textbf{Textile manipulation} extends the evaluation to towels and clothing,\allowbreak{} whose deformation changes the object configuration during manipulation. Together,\allowbreak{} these tasks assess real--simulation correspondence across target selection,\allowbreak{} rigid-object contact and deformable-object interaction.

\phantomsection\label{html:benchmark-scene-coverage}
\hyperref[tab:1]{Table~1} summarizes the complete task coverage; \hyperref[fig:2]{Figure~2} illustrates representative scenes from this evaluation. Per-task Scores and test quantities are provided in \hyperref[tab:B1]{Table~B1},\allowbreak{} and the matched identification comparisons are reported in \hyperref[tab:B2]{Table~B2}.

\subsection{Scene and evaluation specification}
\label{sec:2.3}

\phantomsection\label{html:benchmark-paired-test-protocol}
Paired layouts and scoring. Simulation evaluation uses a fixed Genie Sim experimental branch. Each simulation layout corresponds to a real-robot test layout instance by instance. Evaluation criteria,\allowbreak{} scene layouts,\allowbreak{} test conditions and test settings are aligned between the two environments. Test personnel participate throughout real-robot testing,\allowbreak{} assign the scores manually and adjudicate failures and timeouts. Simulation scores are assigned automatically using the aligned evaluation criteria.

\phantomsection\label{html:benchmark-identification-protocol}
Identification comparison. G2 Swiftpicker,\allowbreak{} G2 Wuji and Arx X5 have simulation Scores reported before and after system identification. G1 Omnipicker data are taken from the experiments reported in Genie Sim 3.0 \cite{ref28}; we did not perform system identification specifically for G1 in this study. All newly added experiments underwent system identification. The 22 final pairs use the G1 reference Scores and the post-identification Scores of the other configurations; the 14 before/after comparisons revisit existing robot--task pairs \allowbreak{}(\hyperref[tab:B1]{Tables~B1}--B2). \hyperref[sec:4]{Section~\ref*{sec:4}} details representative robot and object studies; the X5 study retains its simplified-model scope.

The evaluator distinguishes stage-averaged Score from its separate E2E field. Task families and instruction modes can use different scored stages. We retain the final results table's Score label and report \ensuremath{\Delta} = Sim \ensuremath{-} Real.

\phantomsection\label{html:benchmark-score-definition}
Within a fixed layout of S scored stages and N recorded episodes,\allowbreak{} \ensuremath{\mathrm{Score}=\frac{1}{NS}\sum_i\sum_s c_{is}} before implementation rounding. The evaluator rounds instruction-level stage means,\allowbreak{} weights them by recorded episode counts and rounds the result to four decimals. Full-instruction selection scores Follow and PickUpOnGripper; subtask mode can use a different layout. A separate E2E field is one when the last recorded scored entry is one. Score,\allowbreak{} E2E and binary task success are distinct; the formula does not pool heterogeneous layouts.

\subsection{Experimental coverage and settings}
\label{sec:2.4}

The current results contain 22 robot--task pairs across G1 Omnipicker \allowbreak{}(8),\allowbreak{} G2 Swiftpicker \allowbreak{}(6),\allowbreak{} G2 Wuji \allowbreak{}(1) and Arx X5 \allowbreak{}(7). The other 14 pairs have before/after simulation Scores; these are repeated conditions of existing tasks,\allowbreak{} not additional task coverage.

\phantomsection\label{html:benchmark-rollout-counts}
Real and simulation evaluations use the same number of trials. Experimental settings are summarized in \hyperref[tab:B1]{Table~B1}.

\phantomsection\label{html:benchmark-training-inference}
\par\smallskip\noindent\colorbox{reviewyellow}{\parbox{\dimexpr\linewidth-2\fboxsep\relax}{\normalsize Model training and inference. The newly added experiments use the \ensuremath{\pi_{0.5}} \cite{ref29} baseline with supervised fine-tuning \allowbreak{}(SFT) on real-robot data collected for the corresponding task. The resulting task-specific models are evaluated by inference in both the real and simulated environments. The paired study tests whether a real-data-trained model exhibits corresponding task performance in simulation and on the physical robot.}}\par\smallskip

\subsection{Paired Scores and identification comparison}
\label{sec:2.5}

\phantomsection\label{html:benchmark-agreement-metrics}
Absolute Score agreement. Across the 22 final pairs in \hyperref[tab:B1]{Table~B1},\allowbreak{} MAE = 0.0405,\allowbreak{} RMSE = 0.0520 and mean signed gap = -0.0068. With \ensuremath{\Delta}\textsubscript{i} = Sim\textsubscript{i} \ensuremath{-} Real\textsubscript{i},\allowbreak{} these are mean\allowbreak{}(|\ensuremath{\Delta}\textsubscript{i}|),\allowbreak{} \ensuremath{\surd}mean\allowbreak{}(\ensuremath{\Delta}\textsubscript{i}\textsuperscript{2}) and mean\allowbreak{}(\ensuremath{\Delta}\textsubscript{i}),\allowbreak{} respectively. All 22 pairs have equal weight \allowbreak{}(\hyperref[tab:B1]{Table~B1}). These describe task-specific Score differences,\allowbreak{} not success-rate differences or uncertainty intervals.

\phantomsection\label{html:benchmark-identification-analysis}
Paired identification results. Across the 14 G2/Arx comparisons,\allowbreak{} task-level MAE decreases from 0.0554 to 0.0443 \allowbreak{}(20.0\%),\allowbreak{} and RMSE decreases from 0.0653 to 0.0551 \allowbreak{}(15.6\%). The mean signed gap shifts from -0.0004 to -0.0086. G2 Swiftpicker MAE decreases from 0.0583 to 0.0392; G2 Wuji decreases from 0.0750 to 0.0500. Arx MAE decreases from 0.0500 to 0.0479,\allowbreak{} while RMSE remains approximately 0.0601. These equal-weight task comparisons use the same Real reference \allowbreak{}(\hyperref[tab:B2]{Table~B2}).

Arx X5 shows mixed effects of system identification: some task gaps decrease,\allowbreak{} others remain unchanged,\allowbreak{} and size recognition worsens. \hyperref[tab:B2]{Table~B2} provides the full task-level comparisons.

Before/after comparison. Among the 14 G2/Arx pairs,\allowbreak{} eight have smaller absolute Sim--Real gaps after identification or calibration,\allowbreak{} one has a larger gap,\allowbreak{} and five are unchanged. G2 Swiftpicker has four improvements and two unchanged pairs; G2 Wuji improves; Arx X5 has three improvements,\allowbreak{} one deterioration and three unchanged pairs. \hyperref[tab:B2]{Table~B2} retains all conditions,\allowbreak{} including the worsened Arx size result.

\phantomsection\label{html:completed-real2sim-conclusion}
Conclusion for the completed Real2Sim experiments. The evaluated G2 and Arx tasks already show strong real--simulation correspondence before identification. Across these 14 pairs,\allowbreak{} correspondence improves overall after identification: Pearson r increases from 0.9518 to 0.9669,\allowbreak{} while MAE and RMSE decrease by 20.0\% and 15.6\%,\allowbreak{} respectively \allowbreak{}(\hyperref[tab:B2]{Table~B2}). Across all 22 final robot--task pairs,\allowbreak{} Pearson r = 0.9686 \allowbreak{}(\hyperref[tab:B1]{Table~B1}).

\phantomsection\label{html:benchmark-predictivity-finding}
Across the 22 final robot--task pairs,\allowbreak{} real and simulated Scores show strong consistency: Pearson r = 0.9686,\allowbreak{} MAE = 0.0405 and RMSE = 0.0520 \allowbreak{}(\hyperref[tab:B1]{Table~B1}). This agreement supports using simulation as a reference for real-robot performance across the evaluated embodiments,\allowbreak{} end effectors and task families.

\phantomsection\label{html:pipeline}
\section{Genie Sim \& Real2Sim Pipeline}
\label{sec:3}

Real2Sim connects benchmark task construction with physical-response validation: define paired tasks,\allowbreak{} instantiate robot and interacting assets,\allowbreak{} reproduce recorded commands,\allowbreak{} identify robot and object response,\allowbreak{} and evaluate simulation against real observations and task Scores. This report presents complementary evidence from G2 crsB system identification,\allowbreak{} five signature-asset object studies and 22 paired robot--task evaluations.

\subsection{Real2Sim construction and execution}
\label{sec:3.1}

The object-replay study uses the full G2 model,\allowbreak{} initializing robot and mimic joints from real feedback and applying 22 joint-position targets and two gripper-position targets. Commands originate from HAL and end-effector command messages; the H5 action channel is not substituted. Recorded targets are held at approximately 100 Hz while physics advances in 1 ms substeps. Objects remain free rigid bodies throughout integration: there is no grasp attachment,\allowbreak{} frame-wise object-pose overwrite or inverse-kinematics rewrite.

\begin{table*}[!t]
\centering
\renewcommand{\thetable}{2}
\caption{Reported execution stacks in the object study.}
\label{tab:2}
{\fontsize{8.2}{10.1}\selectfont
\setlength{\tabcolsep}{3pt}
\renewcommand{\arraystretch}{1.13}
\begin{tabular}{>{\raggedright\arraybackslash}p{\dimexpr 0.1500\linewidth-2\tabcolsep\relax}>{\raggedright\arraybackslash}p{\dimexpr 0.2100\linewidth-2\tabcolsep\relax}>{\raggedright\arraybackslash}p{\dimexpr 0.2600\linewidth-2\tabcolsep\relax}>{\raggedright\arraybackslash}p{\dimexpr 0.3800\linewidth-2\tabcolsep\relax}}
\toprule
\rowcolor{tablehead}\textbf{Stack} & \textbf{Version / path} & \textbf{Model route} & \textbf{Interpretation} \\
\midrule
MuJoCo & 3.10.0 CPU & Frozen native G2 + task model & Source fit; soft contact settings \\
\rowcolor{tablerow}Newton / MJWarp & MuJoCo-Warp 3.10.0 CUDA; Newton 1.4.0 & Frozen MJB via SolverMuJoCo & Same contact family; GPU repeat variation \\
PhysX & Isaac Sim 6.0.0 RC \ensuremath{\cdot} CPU TGS & USD built from frozen numeric arrays & Original hard-contact transfer; updated native contact/damping adaptations \allowbreak{}(\hyperref[sec:D]{Appendix~\ref*{sec:D}}) \\
\bottomrule
\end{tabular}
}
\end{table*}

\begin{figure*}[!t]
\centering
\renewcommand{\thefigure}{3}
\includegraphics[width=\linewidth,height=.42\textheight,keepaspectratio]{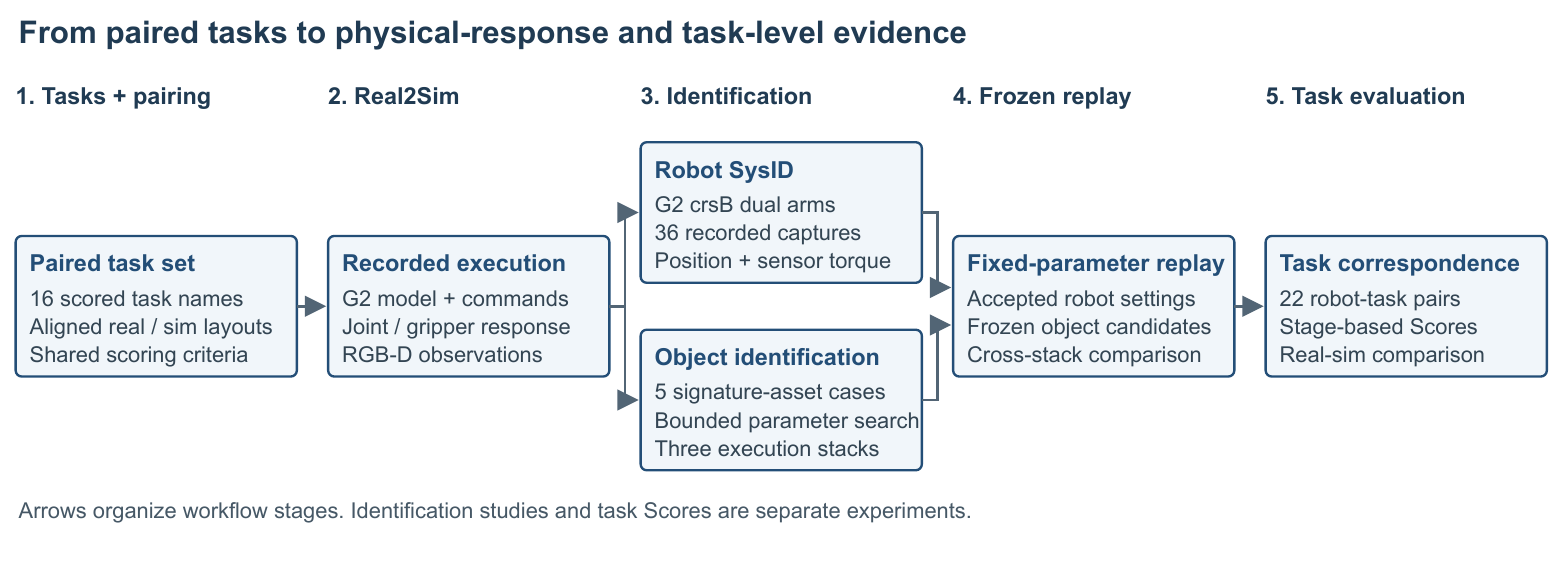}
\caption{Implemented Real2Sim workflow and corresponding evidence. Robot identification,\allowbreak{} object replay and paired task evaluation provide complementary measurements; arrows organize the process rather than denote a single end-to-end experiment.}
\label{fig:3}
\phantomsection\label{html:system-architecture}
\end{figure*}

\begin{figure*}[!t]
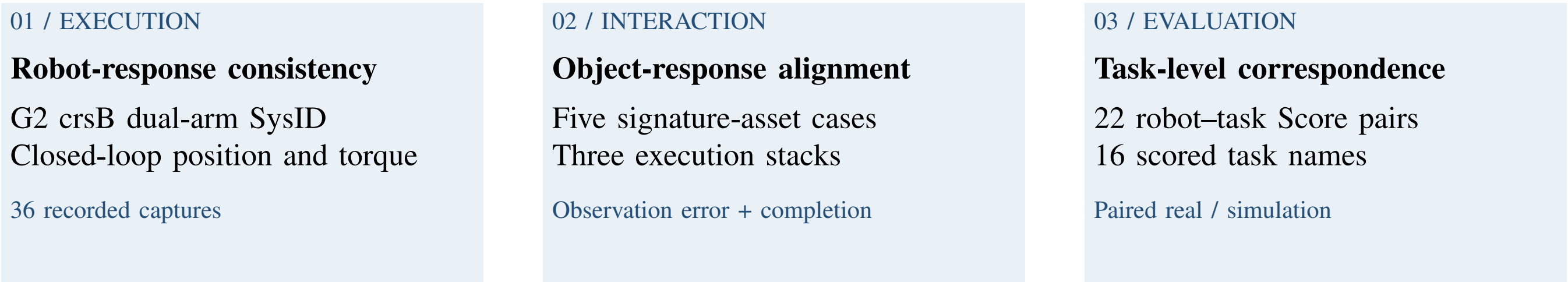

\centering
\renewcommand{\thefigure}{4}
\colorbox{tablehead}{\parbox[t][84pt][t]{.296\linewidth}{\raggedright {\footnotesize\color{reportblue}01 / EXECUTION}\\[4pt]\textbf{Robot-response consistency}\\[4pt]G2 crsB dual-arm SysID\newline{} Closed-loop position and torque\\[5pt]{\footnotesize\color{reportblue}36 recorded captures}}}\hfill\colorbox{tablehead}{\parbox[t][84pt][t]{.296\linewidth}{\raggedright {\footnotesize\color{reportblue}02 / INTERACTION}\\[4pt]\textbf{Object-response alignment}\\[4pt]Five signature-asset cases\newline{} Three execution stacks\\[5pt]{\footnotesize\color{reportblue}Observation error + completion}}}\hfill\colorbox{tablehead}{\parbox[t][84pt][t]{.296\linewidth}{\raggedright {\footnotesize\color{reportblue}03 / EVALUATION}\\[4pt]\textbf{Task-level correspondence}\\[4pt]22 robot--task Score pairs\newline{} 16 scored task names\\[5pt]{\footnotesize\color{reportblue}Paired real / simulation}}}
\caption{Three complementary evidence layers: robot execution,\allowbreak{} object interaction and task outcomes. Each layer has its own metrics and experimental units. Author-created schematic; quantitative results appear in Sections~\ref{sec:2} and~\ref{sec:4}.}
\label{fig:4}
\end{figure*}

Newton in this study uses SolverMuJoCo / MuJoCo-Warp to load the frozen MJB. MuJoCo and MJWarp share contact semantics; PhysX provides a distinct contact implementation. \hyperref[tab:2]{Table~2} identifies the three execution stacks used for the object comparisons.

\subsection{Fixed-parameter replay and evidence alignment}
\label{sec:3.2}

The object study fixes robot execution settings while comparing candidate object parameters and execution stacks. Command response,\allowbreak{} object behavior,\allowbreak{} stage events and task Score are measured separately. Five signature-asset recordings examine representative interaction families; 22 benchmark pairs evaluate task outcomes. The excitation-family robot diagnostics compare joint-side sensor torque with sensor-observer predictions. The independent teleoperation replay instead compares simulated motor-side joint torque with current-derived joint torque. External position targets pass through simulated motion-controller \allowbreak{}(MC) command shaping and the calibrated position servo,\allowbreak{} with mechanical parameters and servo gains fixed. Each study retains its own experimental units and denominator. The object replay retains its own G2 simulation priors and Swiftpicker settings; \hyperref[tab:D1]{Table~D1} records the object-replay and excitation-family robot-diagnostic configurations.

\subsection{What is measured?}
\label{sec:3.3}

We distinguish \emph{real-observation alignment},\allowbreak{} \emph{cross-engine trajectory agreement},\allowbreak{} and \emph{physical task completion}. The observation error is RMSE = \ensuremath{\surd}[mean\allowbreak{}(\ensuremath{\Vert}p\textsubscript{sim} \ensuremath{-} p\textsubscript{obs}\ensuremath{\Vert}\textsuperscript{2})],\allowbreak{} measured in mm for 3D points or px for image points. It is evaluated at the valid observation timestamps. Neither the feature points nor their visibility are common across tasks,\allowbreak{} so their errors are not pooled into a single accuracy score.

RGB-D centres are model-assisted observations rather than motion-capture ground truth. Native simulated contact forces use different readback semantics,\allowbreak{} and no real contact-force reference is supplied. The ``held-out'' time windows are later segments of the same recording; whole-trajectory feasibility gates and development inspection mean they are not blind independent tests.

\phantomsection\label{html:sysid}
\section{System Identification}
\label{sec:4}

Benchmark fidelity depends on two separately testable responses: robot command execution and object interaction under a fixed robot and scene configuration. The independent G2 crsB teleoperation replay applies external position targets through simulated MC command shaping and a calibrated position servo,\allowbreak{} evaluating position and current-derived joint torque with fixed dynamics and gains. The excitation-family sensor diagnostics and G2--Swiftpicker object replays retain their own frozen configurations \allowbreak{}(Appendices C and D). These studies provide complementary execution- and interaction-level evidence under their respective configurations.

\phantomsection\label{html:robot-sysid}
\subsection{Robot Dual-Arm System Identification: G2 crsB}
\label{sec:4.1}

The G2 crsB study identifies both seven-joint arms using friction,\allowbreak{} sine and Fourier experiments. Sections~\ref{sec:4.1.1}--\ref{sec:4.1.3} describe the model,\allowbreak{} staged estimation and recorded-data reconstruction; \hyperref[sec:4.1.4]{Section~\ref*{sec:4.1.4}} tests an unused teleoperation recording. These robot-level evaluations do not establish an asset-linked benchmark benefit.

\begin{figure*}[!t]
\centering
\renewcommand{\thefigure}{5}
\includegraphics[width=\linewidth,height=.42\textheight,keepaspectratio]{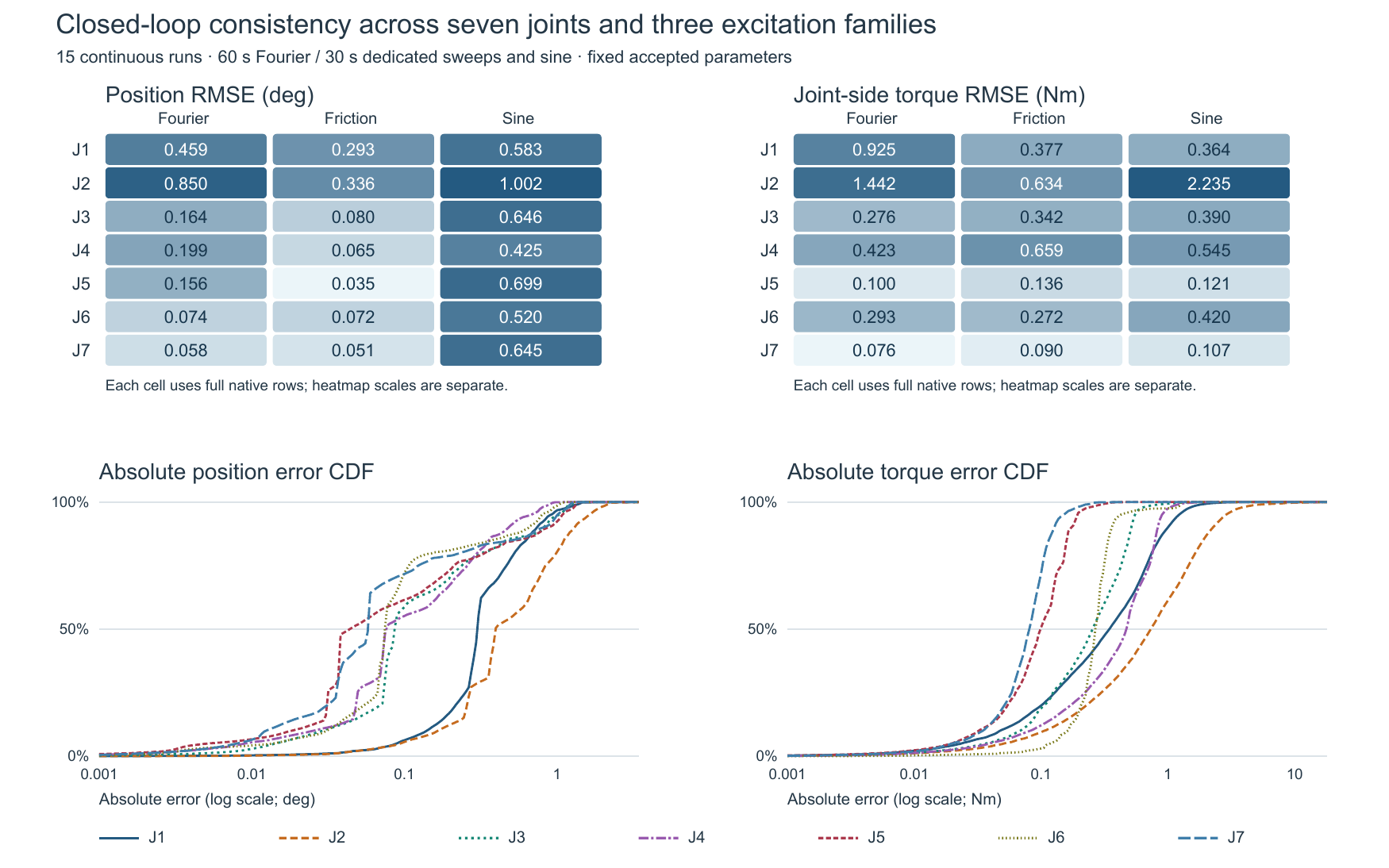}
\caption{Long-horizon closed-loop consistency with frozen parameters. Top: seven-joint RMSE across three excitation families,\allowbreak{} with separate position/torque scales. Bottom: absolute-error CDFs,\allowbreak{} equally weighted by family per joint; logarithmic error axes include the full observed upper tail. Full native rows score each cell; time samples are correlated. Torque is uncentred,\allowbreak{} and the J2 mismatch remains visible. Separate joint-side sensor diagnostics appear in \hyperref[sec:C.4]{Appendix~\ref*{sec:C.4}}.}
\label{fig:5}
\phantomsection\label{html:g2-consistency-overview}
\end{figure*}

The earlier Simulation Closed-Loop Validation in \hyperref[fig:5]{Figure~5} comprises one 60 s coupled Fourier run,\allowbreak{} seven 30 s single-joint friction runs and seven 30 s single-joint sine runs,\allowbreak{} all for the left arm. The 15 runs yield 21 joint--excitation cells \allowbreak{}(J1--J7 \ensuremath{\times} Fourier/friction/sine),\allowbreak{} with only the designated driven joint scored in each single-joint run. Position RMSE spans 0.035\ensuremath{^{\circ}}--1.002\ensuremath{^{\circ}} \allowbreak{}(median 0.293\ensuremath{^{\circ}}),\allowbreak{} and all runs complete without numerical warnings.

Position tracking and torque agreement are separate evidence. Common torque residuals and reversal/cross-frequency failures remain in \hyperref[sec:C]{Appendix~\ref*{sec:C}}. These are same-recording diagnostics under stated control/posture assumptions,\allowbreak{} not independent benchmark task validation.

\subsubsection{Dynamics model and identifiable parameter groups}
\label{sec:4.1.1}
\phantomsection\label{html:g2-dynamics-model}

We identify the seven active joints of each arm under a fixed-base model. Joint position,\allowbreak{} velocity and acceleration are \ensuremath{\boldsymbol{q}},\allowbreak{} \ensuremath{\boldsymbol{\dot{q}}} and \ensuremath{\boldsymbol{\ddot{q}}} \ensuremath{\in} \ensuremath{\mathbb R}\textsuperscript{7},\allowbreak{} in rad,\allowbreak{} rad/s and rad/s\textsuperscript{2}. The observation \ensuremath{\boldsymbol{\tau _{I}}} \ensuremath{\in} \ensuremath{\mathbb R}\textsuperscript{7} is motor current converted to joint-side torque; joint-side and end-effector force sensors do not enter the fit.

\begin{equation}
\boldsymbol{\tau}_{I,k}=\mathbf Y_k\boldsymbol\pi+\operatorname{diag}(\ddot{\mathbf q}_k)\mathbf J_a+\boldsymbol\tau_f(\dot{\mathbf q}_k)+\mathbf b_\tau+\boldsymbol\epsilon_k
\end{equation}
Here j indexes joints J1--J7 and k indexes time samples. The jth torque component is \ensuremath{\tau _{I,j,k}} = \ensuremath{p_{j}}\ensuremath{K_{t,j}}\ensuremath{n_{j}}\ensuremath{i_{j,k}},\allowbreak{} where \ensuremath{i_{j,k}} is current in A,\allowbreak{} \ensuremath{K_{t,j}} the fixed motor torque constant in N\ensuremath{\cdot}m/A,\allowbreak{} \ensuremath{n_{j}} the reduction ratio,\allowbreak{} and \ensuremath{p_{j}} \ensuremath{\in} \{\ensuremath{-}1,\allowbreak{}1\} the joint-coordinate sign. \ensuremath{\boldsymbol{Y_{k}}} = \ensuremath{\boldsymbol{Y}}\allowbreak{}(\ensuremath{\boldsymbol{q_{k}}},\allowbreak{}\ensuremath{\boldsymbol{\dot{q}_{k}}},\allowbreak{}\ensuremath{\boldsymbol{\ddot{q}_{k}}}) \ensuremath{\in} \ensuremath{\mathbb R}\textsuperscript{7\ensuremath{\times}70} is the Pinocchio rigid-body torque regressor,\allowbreak{} \ensuremath{\boldsymbol{\pi }} \ensuremath{\in} \ensuremath{\mathbb R}\textsuperscript{7}\ensuremath{{}^0} the link inertial parameter vector,\allowbreak{} and diag forms a diagonal matrix. \ensuremath{\boldsymbol{J_{a}}} \ensuremath{\in} \ensuremath{\mathbb R}\textsuperscript{7} is armature in kg\ensuremath{\cdot}m\textsuperscript{2},\allowbreak{} \ensuremath{\boldsymbol{b_{\tau }}} \ensuremath{\in} \ensuremath{\mathbb R}\textsuperscript{7} a constant current-to-torque offset in N\ensuremath{\cdot}m,\allowbreak{} and \ensuremath{\boldsymbol{\epsilon _{k}}} \ensuremath{\in} \ensuremath{\mathbb R}\textsuperscript{7} the torque residual. The friction vector has components \ensuremath{\tau _{f,j,k}} = \ensuremath{f_{c,j}} tanh\allowbreak{}(\ensuremath{\dot{q}_{j,k}}/C\ensuremath{_0}) + \ensuremath{f_{v,j}}\ensuremath{\dot{q}_{j,k}}; \ensuremath{f_{c,j}} and \ensuremath{f_{v,j}} are nonnegative Coulomb and viscous coefficients,\allowbreak{} in N\ensuremath{\cdot}m and N\ensuremath{\cdot}m\ensuremath{\cdot}s/rad. C\ensuremath{_0} = 0.001 rad/s is the smoothing scale.

Armature denotes inertia reflected to the joint coordinate,\allowbreak{} including rotor and approximately acceleration-proportional transmission effects. It is installed directly in the simulator without another reduction-ratio-squared factor. The bias is a calibration/compensation term,\allowbreak{} separate from link inertia.

\begin{equation}
\begin{aligned}\boldsymbol\pi_i=[&m_i,h_{x,i},h_{y,i},h_{z,i},I_{xx,i},I_{xy,i},\\&I_{yy,i},I_{xz,i},I_{yz,i},I_{zz,i}]^T\end{aligned}
\end{equation}
Here i indexes the seven links,\allowbreak{} and \ensuremath{\boldsymbol{\pi _{i}}} contains the ten standard inertial parameters of link i: mass m\textsubscript{i} \allowbreak{}(kg),\allowbreak{} three first moments \ensuremath{h_{x,i}},\allowbreak{} \ensuremath{h_{y,i}} and \ensuremath{h_{z,i}} \allowbreak{}(kg\ensuremath{\cdot}m),\allowbreak{} and six independent entries of the symmetric inertia tensor \ensuremath{\boldsymbol{I_{O,i}}} \allowbreak{}(kg\ensuremath{\cdot}m\textsuperscript{2}). The first-moment vector is \ensuremath{\boldsymbol{h_{i}}} = m\textsubscript{i}\ensuremath{\boldsymbol{c_{i}}},\allowbreak{} where \ensuremath{\boldsymbol{c_{i}}} is the centre of mass relative to the link-local origin; x,\allowbreak{} \ensuremath{\boldsymbol{y}} and z denote the local coordinate axes. \ensuremath{\boldsymbol{I_{O,i}}} is about that origin,\allowbreak{} expressed in the same axes,\allowbreak{} and the superscript T denotes transpose. The full \ensuremath{\boldsymbol{\pi }} concatenates these seven link blocks. Thus the model has 70 inertial,\allowbreak{} seven armature,\allowbreak{} fourteen friction and seven bias unknowns: 98 scalars per arm.

These 98 coordinates are not all independently observable from joint torque. Rigid-body regressors contain structural dependencies,\allowbreak{} while limited excitation can further correlate inertia,\allowbreak{} armature and friction columns. A simultaneous fit therefore cannot generally recover a unique full physical parameter vector. We separate the experiments and regularize physically feasible estimates; this reduces coupling without claiming unique identification of every link property [12--16].

Both dynamic stages require each link's pseudo-inertia matrix to be positive semidefinite,\allowbreak{} constrain mass to 50\%--150\% of its original URDF/CAD reference,\allowbreak{} and enforce nonnegative armature/friction; bias signs are unrestricted.

\subsubsection{Staged excitation and constrained estimation}
\label{sec:4.1.2}
\phantomsection\label{html:g2-staged-estimation}

Each arm contributes 18 recordings: seven single-joint constant-speed friction sweeps,\allowbreak{} ten single-joint sine recordings,\allowbreak{} and one synchronous seven-joint Fourier recording. J1--J3 sine recordings each use two background postures \allowbreak{}(zero posture and J4 approximately \ensuremath{-}\ensuremath{\pi}/2); J4--J7 each use one zero-posture background. This gives 7 + \allowbreak{}(3\ensuremath{\times}2 + 4) + 1 = 18,\allowbreak{} rather than eighteen distinct trajectories per joint.

\textbf{Stage 1: friction.} Twenty speed levels span 0.0087--3.0 rad/s,\allowbreak{} with forward/reverse plateaus. Samples must be within 1\% of the prescribed speed and satisfy |\ensuremath{\ddot q}| \ensuremath{\leq} 0.15 rad/s\textsuperscript{2}. Adjacent opposite-direction plateaus are matched over their common position interval and divided into ten bins,\allowbreak{} each retaining at least five samples per direction.

\begin{equation}
\begin{aligned}v_\ell&=[\operatorname{med}(\dot q_+)-\operatorname{med}(\dot q_-)]/2,\\z_\ell&=[\operatorname{med}(\tau_{I,+})-\operatorname{med}(\tau_{I,-})]/2,\\r_\ell&=f_{c,j}\tanh(v_\ell/C_0)+f_{v,j}v_\ell-z_\ell\end{aligned}
\end{equation}
Here med denotes the median; + and \ensuremath{-} denote positive- and negative-direction samples in paired position bin \ensuremath{\ell}. \ensuremath{v_{\ell }} is the paired speed,\allowbreak{} \ensuremath{z_{\ell }} the paired torque difference,\allowbreak{} and \ensuremath{r_{\ell }} the friction-fit residual. Position matching approximately cancels gravity,\allowbreak{} static load and constant bias. Nonnegative Huber iteratively reweighted least squares yields \ensuremath{\boldsymbol{f_{j}^{(1)}}} = [\ensuremath{f_{c,j}},\allowbreak{}\ensuremath{f_{v,j}}]\ensuremath{{}^T},\allowbreak{} where the superscript \allowbreak{}(1) denotes the first-stage output,\allowbreak{} with zero prior,\allowbreak{} ridge weight 10\textsuperscript{-}\ensuremath{{}^6},\allowbreak{} and adaptive threshold \ensuremath{\delta _{f}} = 2\ensuremath{\times}1.4826\ensuremath{\times}MAD\allowbreak{}(r). MAD is the median absolute deviation of residuals from their median; \ensuremath{\delta _{f}} has torque units.

\textbf{Stage 2: armature.} Each target joint is excited at 0.5,\allowbreak{} 0.7,\allowbreak{} 1.0,\allowbreak{} 1.3 and 1.5 Hz for at least 10 s per frequency. Amplitude is limited by the safe position interval and 80\% of joint velocity/acceleration limits. Transitions and waits are excluded; continuous segments are cropped by 0.5 s at each end. Different frequencies and background postures help separate acceleration-proportional armature from configuration-dependent link inertia.

\begin{equation}
\begin{aligned}\min\quad &\frac1{N_j}\sum_k\rho_1(e_{j,k})\\&+\lambda_\pi\|\mathbf D_\pi(\boldsymbol\pi-\boldsymbol\pi^{\mathrm{URDF}})\|_2^2\\&+\lambda_f\|\mathbf D_f(\mathbf f_j-\mathbf f_j^{(1)})\|_2^2\end{aligned}
\end{equation}
Here N\textsubscript{j} counts retained time samples for target joint j,\allowbreak{} and \ensuremath{e_{j,k}} is predicted minus current-derived measured torque at sample k. Prediction follows the dynamics model above,\allowbreak{} with a separate constant nuisance offset \ensuremath{b_{j,s}^{seg}} for each continuous segment s in place of the final \ensuremath{b_{\tau ,j}}. The first term averages the Huber loss \ensuremath{\rho}\ensuremath{_1} over samples: for residual e and threshold \ensuremath{\delta} = 1 N\ensuremath{\cdot}m,\allowbreak{} it equals e\textsuperscript{2} when |e| \ensuremath{\leq} \ensuremath{\delta} and 2\ensuremath{\delta}|e| \ensuremath{-} \ensuremath{\delta}\textsuperscript{2} otherwise. The second and third terms penalize deviations from the original inertia prior \ensuremath{\boldsymbol{\pi }}\textsuperscript{URDF} and first-stage friction estimate \ensuremath{\boldsymbol{f_{j}^{(1)}}}. The friction pair is \ensuremath{\boldsymbol{f_{j}}} = [\ensuremath{f_{c,j}},\allowbreak{}\ensuremath{f_{v,j}}]\ensuremath{{}^T}; \ensuremath{\lambda _{\pi }} = 1.0 and \ensuremath{\lambda _{f}} = 0.1 are regularization weights,\allowbreak{} \ensuremath{\Vert}\ensuremath{\cdot}\ensuremath{\Vert}\ensuremath{_2} is the Euclidean norm,\allowbreak{} and \ensuremath{\boldsymbol{D_{\pi }}}/\ensuremath{\boldsymbol{D_{f}}} are diagonal inverse-prior-magnitude scaling matrices. Inertia scales are floored at 10\textsuperscript{-}\ensuremath{{}^3} times their positive-scale median; friction scales at 10\textsuperscript{-}\ensuremath{{}^3}. Each solve estimates a temporary full \ensuremath{\boldsymbol{\pi }},\allowbreak{} one \ensuremath{J_{a,j}},\allowbreak{} \ensuremath{\boldsymbol{f_{j}}} and the segment offsets; armature has no prior penalty.

After seven solves,\allowbreak{} only the ten inertial parameters of the target joint's link are retained from each temporary \ensuremath{\boldsymbol{\pi }}. These blocks form \ensuremath{\boldsymbol{\pi }}\textsuperscript{\allowbreak{}(2)},\allowbreak{} while the seven armature and fourteen friction estimates form \ensuremath{\boldsymbol{J_{a}^{(2)}}} and \ensuremath{\boldsymbol{f}}\textsuperscript{\allowbreak{}(2)}. The superscript \allowbreak{}(2) denotes the second-stage output.

\textbf{Stage 3: coupled Fourier dynamics.} All seven joints execute a five-harmonic Fourier trajectory with a 20 s period for five cycles \allowbreak{}(approximately 100 s). Coefficients are optimized for the nonzero-singular-value conditioning of the normalized rigid-body regressor,\allowbreak{} subject to safe joint position and 90\% velocity/acceleration limits. Cycle 1 is skipped,\allowbreak{} cycles 2--4 fit the model,\allowbreak{} and cycle 5 is reserved in the identification procedure. The full-recording diagnostic below deliberately scores the original recording,\allowbreak{} including its fitting intervals.

\begin{equation}
\begin{aligned}\min_{\boldsymbol\pi,\mathbf b_\tau}\quad &\frac1{N_\tau}\sum_r\rho_1(u_r)\\&+0.5\|\mathbf D_\pi(\boldsymbol\pi-\boldsymbol\pi^{(2)})\|_2^2\end{aligned}
\end{equation}
Here r indexes stacked sample--joint rows,\allowbreak{} N\textsubscript{\ensuremath{\tau}} is their training count,\allowbreak{} and \ensuremath{u_{r}} is predicted minus current-derived measured torque for row r. Prediction follows the dynamics model above,\allowbreak{} with armature \ensuremath{\boldsymbol{J_{a}^{(2)}}} and friction \ensuremath{\boldsymbol{f}}\textsuperscript{\allowbreak{}(2)} fixed at their second-stage estimates; only \ensuremath{\boldsymbol{\pi }} and \ensuremath{\boldsymbol{b_{\tau }}} are optimized. The first term is the mean Huber loss with the same 1 N\ensuremath{\cdot}m threshold. The second term regularizes inertia toward \ensuremath{\boldsymbol{\pi }}\textsuperscript{\allowbreak{}(2)} with weight 0.5. \ensuremath{\boldsymbol{D_{\pi }}} now scales by \ensuremath{\boldsymbol{\pi }}\textsuperscript{\allowbreak{}(2)},\allowbreak{} with a floor of 10\textsuperscript{-}\ensuremath{{}^3} times the positive-scale median and an absolute floor of 10\textsuperscript{-}\ensuremath{{}^8}. CLARABEL is preferred,\allowbreak{} with SCS as fallback; only optimal or optimal\_\allowbreak{}inaccurate solver statuses are accepted.

Identification data are time-sorted,\allowbreak{} deduplicated,\allowbreak{} split at gaps longer than 0.10 s and linearly resampled. General dynamic processing uses 100 Hz with a 0.11 s,\allowbreak{} second-order Savitzky--Golay velocity/acceleration window; sine processing uses 200 Hz with a 0.035 s,\allowbreak{} third-order window. These method settings are separate from the archived scoring harness and display sampling described in \hyperref[sec:C.1]{Appendix~\ref*{sec:C.1}}.

\subsubsection{Identified parameters and Fourier reconstruction}
\label{sec:4.1.3}
\phantomsection\label{html:g2-identified-results}

\hyperref[tab:3]{Table~3} reports the final identified model on the original approximately 100 s Fourier recording only. Its current-derived torque RMS retains the model's own constant offset; no new per-recording centring is fitted for this score. For an arm,\allowbreak{} RMS = \ensuremath{\surd}\allowbreak{}(\ensuremath{\sum}\textsubscript{r=1}\textsuperscript{N}\ensuremath{e_{r}}\textsuperscript{2}/N),\allowbreak{} where \ensuremath{e_{r}} is predicted minus measured joint torque and N = 6,\allowbreak{}986 scored joint-time rows \allowbreak{}(998 time rows \ensuremath{\times} seven joints). The supplied scoring rows are temporally correlated; they are neither independent trials nor plot vertices.

\begin{table}
\centering
\renewcommand{\thetable}{3}
\caption{Final-model Fourier torque reconstruction,\allowbreak{} approximately 100 s. One recording per arm; all seven joints; own-offset RMS. Same-recording diagnostics,\allowbreak{} including fitting intervals.}
\label{tab:3}
{\fontsize{8.2}{10.1}\selectfont
\setlength{\tabcolsep}{3pt}
\renewcommand{\arraystretch}{1.13}
\begin{tabular}{>{\raggedright\arraybackslash}p{\dimexpr 0.1800\linewidth-2\tabcolsep\relax}>{\raggedright\arraybackslash}p{\dimexpr 0.3900\linewidth-2\tabcolsep\relax}>{\raggedright\arraybackslash}p{\dimexpr 0.4300\linewidth-2\tabcolsep\relax}}
\toprule
\rowcolor{tablehead}\textbf{Arm} & \textbf{Torque RMS \allowbreak{}(N\ensuremath{\cdot}m)} & \textbf{Scored joint-time rows} \\
\midrule
Left & 0.7709 & 6,\allowbreak{}986 \\
\rowcolor{tablerow}Right & 0.9698 & 6,\allowbreak{}986 \\
\bottomrule
\end{tabular}
}
\end{table}

\hyperref[fig:C3]{Figure~C3} shows the full left-arm torque reconstruction. Its plotted curves use the archived recording/joint centring,\allowbreak{} while its Own RMS annotations follow \hyperref[tab:3]{Table~3}'s own-offset convention. \hyperref[tab:C1]{Tables~C1}--C2 preserve the complete joint/family breakdown. The full-recording archive does not supply a matching approximately 100 s simulated-position score; position and torque response on an unused trajectory are evaluated together in \hyperref[sec:4.1.4]{Section~\ref*{sec:4.1.4}}.

\begin{table*}[!t]
\centering
\renewcommand{\thetable}{4}
\caption{Identified joint parameters for both arms,\allowbreak{} J1--J7. Armature J\ensuremath{_a}: kg\ensuremath{\cdot}m\textsuperscript{2}; Coulomb f\_\allowbreak{}c and bias b\ensuremath{\tau}: N\ensuremath{\cdot}m; viscous f\_\allowbreak{}v: N\ensuremath{\cdot}m\ensuremath{\cdot}s/rad. Values retain the supplied four-decimal precision.}
\label{tab:4}
{\fontsize{8.2}{10.1}\selectfont
\setlength{\tabcolsep}{3pt}
\renewcommand{\arraystretch}{1.13}
\begin{tabular}{>{\raggedright\arraybackslash}p{\dimexpr 0.1000\linewidth-2\tabcolsep\relax}>{\raggedright\arraybackslash}p{\dimexpr 0.1000\linewidth-2\tabcolsep\relax}>{\raggedright\arraybackslash}p{\dimexpr 0.2000\linewidth-2\tabcolsep\relax}>{\raggedright\arraybackslash}p{\dimexpr 0.2000\linewidth-2\tabcolsep\relax}>{\raggedright\arraybackslash}p{\dimexpr 0.2000\linewidth-2\tabcolsep\relax}>{\raggedright\arraybackslash}p{\dimexpr 0.2000\linewidth-2\tabcolsep\relax}}
\toprule
\rowcolor{tablehead}\textbf{Arm} & \textbf{Joint} & \textbf{Armature J\ensuremath{_a}} & \textbf{Coulomb f\_\allowbreak{}c} & \textbf{Viscous f\_\allowbreak{}v} & \textbf{Bias b\ensuremath{\tau}} \\
\midrule
Left & J1 & 1.5165 & 5.2766 & 4.1386 & 0.0246 \\
\rowcolor{tablerow}Left & J2 & 1.5031 & 5.4307 & 2.7398 & 1.1140 \\
Left & J3 & 0.5690 & 1.8589 & 1.2645 & 0.0177 \\
\rowcolor{tablerow}Left & J4 & 0.5717 & 1.6363 & 1.9437 & 0.2242 \\
Left & J5 & 0.2869 & 1.3225 & 0.6124 & 0.0111 \\
\rowcolor{tablerow}Left & J6 & 0.2871 & 1.9650 & 0.6518 & 0.0568 \\
Left & J7 & 0.2850 & 1.5350 & 0.8837 & 0.0392 \\
\rowcolor{tablerow}Right & J1 & 1.5565 & 5.2096 & 3.7917 & -0.1226 \\
Right & J2 & 1.5503 & 4.2864 & 2.6620 & 0.2514 \\
\rowcolor{tablerow}Right & J3 & 0.5763 & 2.2484 & 1.4317 & 0.0192 \\
Right & J4 & 0.5877 & 2.1654 & 2.0767 & -0.0021 \\
\rowcolor{tablerow}Right & J5 & 0.2878 & 1.2849 & 0.6122 & -0.0168 \\
Right & J6 & 0.2856 & 1.4960 & 0.5621 & 0.0372 \\
\rowcolor{tablerow}Right & J7 & 0.2925 & 1.4512 & 0.7401 & -0.0369 \\
\bottomrule
\end{tabular}
}
\end{table*}

Both arms show armature groups near 1.5 kg\ensuremath{\cdot}m\textsuperscript{2} for J1--J2,\allowbreak{} 0.57 kg\ensuremath{\cdot}m\textsuperscript{2} for J3--J4 and 0.29 kg\ensuremath{\cdot}m\textsuperscript{2} for J5--J7. This is a plausibility check,\allowbreak{} not an independent rotor measurement. Link inertias,\allowbreak{} armature,\allowbreak{} motor friction and current-to-torque bias remain distinct from the joint-side sensor observation terms examined in \hyperref[sec:C.4]{Appendix~\ref*{sec:C.4}}. Those diagnostics retain their original simulation PD configuration in \hyperref[tab:D1]{Table~D1} and their seven-joint results in \hyperref[tab:C6]{Table~C6}.

\reportnotePageSeven

\subsubsection{Independent teleoperation validation}
\label{sec:4.1.4}
\phantomsection\label{html:g2-independent-teleoperation}
\par\smallskip\noindent\colorbox{reviewyellow}{\parbox{\dimexpr\linewidth-2\fboxsep\relax}{\normalsize A separate 32.6503 s teleoperation recording was excluded from mechanical identification and servo-gain search. External position targets pass through simulated motion-controller \allowbreak{}(MC) command shaping before an equivalent joint PD servo drives the identified MuJoCo model. The servo uses both shaped position and velocity commands: \[\begin{aligned}\tau_{\mathrm{PD},j}&=K_{p,j}(q_j^c-q_j)\\&\quad+K_{d,j}(\dot q_j^c-\dot q_j).\end{aligned}\] Here superscript c denotes the MC command and the unmarked variables are simulated states. The derivative term tracks commanded velocity; it is not zero-target damping. Mechanics and servo gains remain fixed for this recording. \hyperref[sec:C.3]{Appendix~\ref*{sec:C.3}} specifies the execution chain and gains; \hyperref[tab:C3]{Table~C3} retains both arms.}}\par\smallskip

Across both arms,\allowbreak{} per-joint position RMSE spans 0.00614--0.03522 rad and current-derived torque RMSE spans 0.73545--4.00099 N\ensuremath{\cdot}m. Scores use 3,\allowbreak{}165 measured-time samples after excluding the first 1 s of warmup. \hyperref[fig:C4]{Figure~C4} shows the complete left-arm recording; \hyperref[tab:C3]{Table~C3} retains all fourteen arm/joint results. This single-recording test evaluates the combined execution chain rather than isolating the contribution of mechanical identification.

\paragraph{Supplementary result: simplified X5 identification}

X5-related platform and simulation references include LeggedManip\_\allowbreak{}Lab \cite{ref9},\allowbreak{} DISCOVERSE \cite{ref10},\allowbreak{} and SIM1 \cite{ref11}; repository-level references do not establish which assets were reused. X5 lacks force control,\allowbreak{} so we use a simplified model for its Real2Sim parameter identification. This is a supplementary result; the detailed robot SysID study focuses on G2 crsB.

\phantomsection\label{html:object-id}
\subsection{Object Identification and Cross-Engine Diagnostics}
\label{sec:4.2}

We deliberately select five signature assets from Genie Sim Benchmark to validate representative interaction families: a kettle and gum box for grasping and carrying,\allowbreak{} a sponge for effective compliant contact,\allowbreak{} and a shuttlecock and pen for handover and rotational resistance. This is a category-level validation of how our simulator performs object identification,\allowbreak{} rather than an exhaustive,\allowbreak{} identical experiment on every benchmark asset.

\subsubsection{Representative assets and identification workflow}
\label{sec:4.2.1}

With the G2--Swiftpicker execution configuration fixed,\allowbreak{} we fit bounded effective object parameters against recorded real interactions. The original study freezes each MuJoCo candidate for transfer to MJWarp and PhysX; separate native PhysX adaptation is reported in \hyperref[sec:4.2.2]{Section~\ref*{sec:4.2.2}}. Kettle fitting adjusts geometry,\allowbreak{} COM,\allowbreak{} initial yaw and contact; other cases emphasize contact,\allowbreak{} with an additional pen-radius search. Object mass priors remain fixed.

Observation and fitting protocol. Each asset has one recorded real interaction. Observations are compared at valid timestamps. For the gum box and sponge,\allowbreak{} the reported objective combines training centre RMSE \ensuremath{e_{\mathrm{3D}}} and holding-gripper RMSE ehold: \begin{equation}
J_{\mathrm{3D}}=\left(\frac{e_{\mathrm{3D}}}{0.01\,\mathrm m}\right)^2
+0.1\left(\frac{e_{\mathrm{hold}}}{0.03\,\mathrm{rad}}\right)^2.
\end{equation} Handover fitting uses RGB error and dual-gripper feedback. Full-trajectory task gates and development inspection make the later segment a same-episode temporal check. \hyperref[tab:D2]{Table~D2} gives all six search rows,\allowbreak{} objectives,\allowbreak{} time windows and selection scope; \hyperref[tab:D3]{Table~D3} retains all 15 case--stack results.

\begin{table*}[!t]
\centering
\renewcommand{\thetable}{5}
\caption{Signature benchmark assets,\allowbreak{} observation coverage and identification scope.}
\label{tab:5}
{\fontsize{8.2}{10.1}\selectfont
\setlength{\tabcolsep}{3pt}
\renewcommand{\arraystretch}{1.13}
\begin{tabular}{>{\raggedright\arraybackslash}p{\dimexpr 0.1900\linewidth-2\tabcolsep\relax}>{\raggedright\arraybackslash}p{\dimexpr 0.3000\linewidth-2\tabcolsep\relax}>{\raggedright\arraybackslash}p{\dimexpr 0.5100\linewidth-2\tabcolsep\relax}}
\toprule
\rowcolor{tablehead}\textbf{Asset} & \textbf{Real observations} & \textbf{Search and adjusted quantities} \\
\midrule
Kettle & 260 RGB-D lid observations & Geometry,\allowbreak{} COM,\allowbreak{} yaw,\allowbreak{} contact; 3 grid rounds \\
\rowcolor{tablerow}Gum box & 127 RGB-D centre observations & Contact; 100 candidates \\
Sponge & 92 RGB-D centre observations & Effective contact; 120 MuJoCo candidates \\
\rowcolor{tablerow}Shuttlecock & 12 RGB points; 3 valid depths & Sliding/rotational contact; 81 candidates \\
Pen & 11 RGB points and valid depths & 81 contact + 42 radius candidates; rejected \\
\bottomrule
\end{tabular}
}
\end{table*}

\begin{figure*}[!t]
\centering
\renewcommand{\thefigure}{6}
\includegraphics[width=.48\linewidth,height=.43\textheight,keepaspectratio]{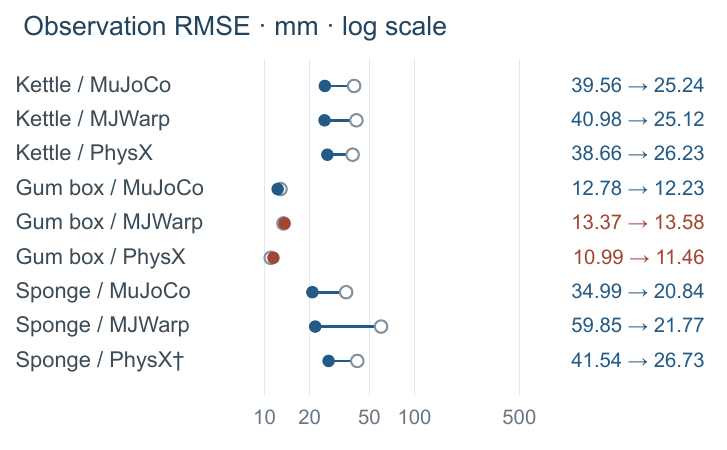}\hfill
\includegraphics[width=.48\linewidth,height=.43\textheight,keepaspectratio]{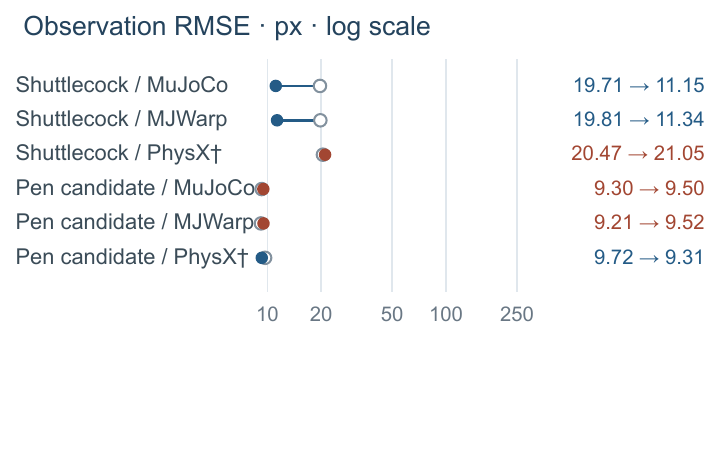}

\par\smallskip{\footnotesize \ensuremath{\circ} Before \ensuremath{\cdot} \ensuremath{\bullet} After / candidate \ensuremath{\cdot} Red: higher error. \textsuperscript{\ensuremath{\dagger}} Adapted PhysX. Source pen candidate not accepted.}

\caption{Fifteen case--stack mean pairs,\allowbreak{} compacted for the main text. PhysX\textsuperscript{\ensuremath{\dagger}} uses the matched native configurations in \hyperref[tab:D3]{Table~D3}; other rows retain source results. Hollow markers are baseline,\allowbreak{} filled markers are fitted / candidate models; red marks increased error. The source pen candidate remains unaccepted. Units and log axes differ between panels. Historical repeat ranges and failures remain in \hyperref[sec:D]{Appendix~\ref*{sec:D}}; no confidence intervals are inferred.\newline{}}
\label{fig:6}
\phantomsection\label{html:object-summary}
\end{figure*}

\subsubsection{Replay results and native PhysX recovery}
\label{sec:4.2.2}
\phantomsection\label{html:sponge-diagnosis}

\begin{figure*}[!t]
\centering
\renewcommand{\thefigure}{7}
\includegraphics[width=\linewidth,height=.32\textheight,keepaspectratio]{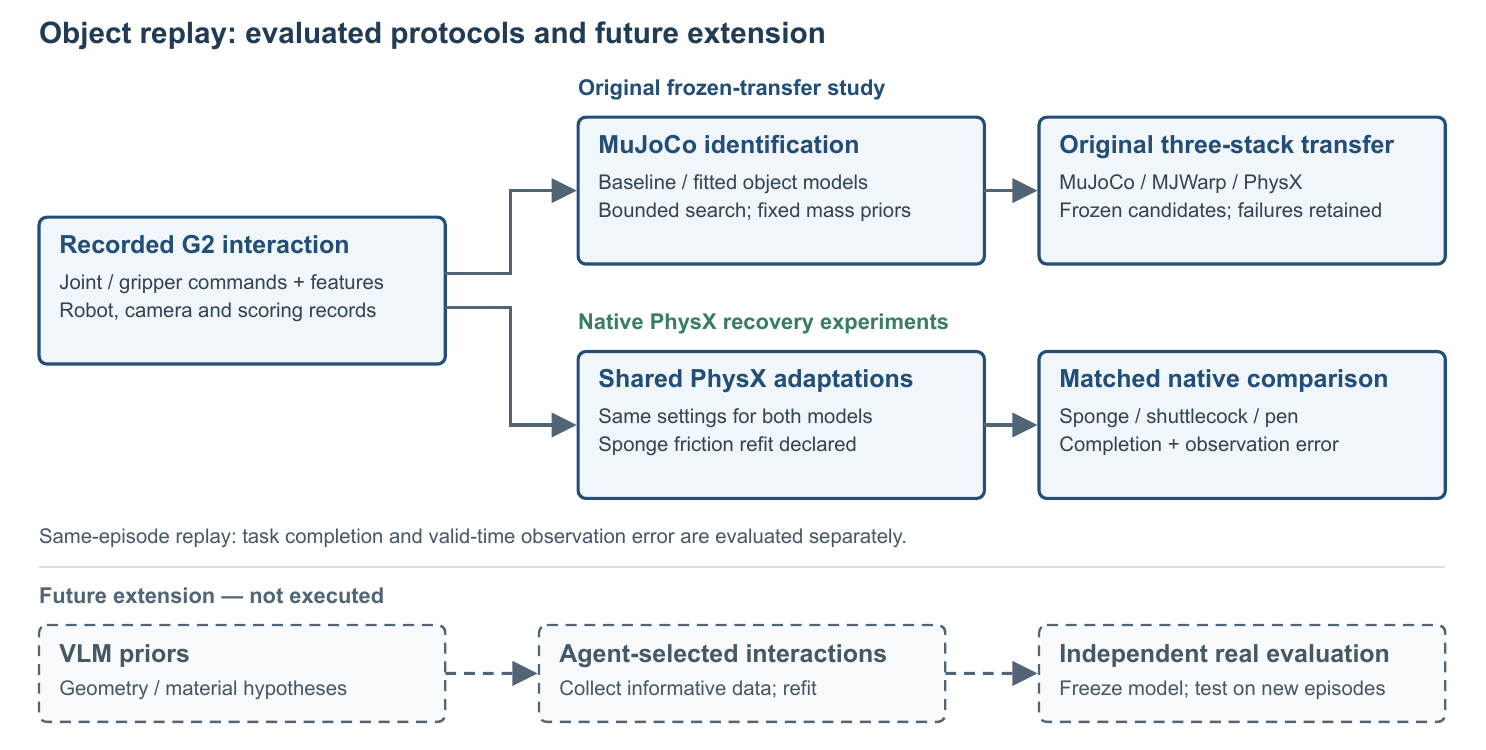}
\caption{Two evaluated object-replay protocols. The source study freezes the MuJoCo candidate for three-stack comparison. The native PhysX experiments apply the same accepted engine settings to both parameter sets and separately declare the sponge friction refit. Both protocols evaluate task completion and observation error. The dashed strip outlines future VLM priors,\allowbreak{} agent-selected interactions and independent real evaluation; these steps have not been executed. Author-created schematic,\allowbreak{} not an additional experiment.}
\label{fig:7}
\phantomsection\label{html:object-evidence-loop}
\end{figure*}

\hyperref[sec:4.2.1]{Section~\ref*{sec:4.2.1}} describes the original frozen-parameter transfer study. Recorded real joint and gripper position commands drive each replay. The later native PhysX comparisons use separate engine adaptations; the sponge additionally uses a declared native friction refit. \hyperref[fig:7]{Figure~7} distinguishes these two protocols.

\textbf{Original cross-engine results.} Kettle lid-centre RMSE decreases from \textbf{39.56 to 25.24 mm} in MuJoCo,\allowbreak{} 40.98 to 25.12 mm in MJWarp,\allowbreak{} and 38.66 to 26.23 mm in PhysX. The gum box improves slightly in MuJoCo \allowbreak{}(12.78 \ensuremath{\rightarrow} 12.23 mm),\allowbreak{} while errors increase in MJWarp \allowbreak{}(13.37 \ensuremath{\rightarrow} 13.58 mm) and PhysX \allowbreak{}(10.99 \ensuremath{\rightarrow} 11.46 mm). The source pen candidate remains rejected. \hyperref[fig:6]{Figure~6} and \hyperref[tab:D3]{Table~D3} retain all fifteen case--stack comparisons,\allowbreak{} including these mixed outcomes.

\textbf{Contact-transfer failures and recovery.} The original sponge transfer increases PhysX error from 25.50 to 524.61 mm and fails to lift the object; the original PhysX shuttlecock and pen roll away before closure. Native recovery applies common finger-contact settings to both sponge models \allowbreak{}(800 N/m stiffness,\allowbreak{} 5 N\ensuremath{\cdot}s/m damping),\allowbreak{} and common angular damping to both handover models \allowbreak{}(20 s\textsuperscript{-}\textsuperscript{1} for shuttlecock,\allowbreak{} 500 s\textsuperscript{-}\textsuperscript{1} for pen). Both parameter sets then complete their respective tasks: sponge 2/2 \ensuremath{\rightarrow} 4/4,\allowbreak{} shuttlecock 2/2 \ensuremath{\rightarrow} 2/2,\allowbreak{} and pen 2/2 \ensuremath{\rightarrow} 2/2 replays. Historical failures and the earlier separate sponge fit remain in D.1--D.3.

\textbf{Identification under matched settings.} Native sponge friction fitting reduces full centre RMSE \textbf{41.54 \ensuremath{\rightarrow} 26.73 mm \allowbreak{}(+35.64\%)},\allowbreak{} although vertical error increases 13.40 \ensuremath{\rightarrow} 19.38 mm. Shuttlecock RGB error increases \textbf{20.47 \ensuremath{\rightarrow} 21.05 px} \allowbreak{}(\ensuremath{-}2.84\% improvement),\allowbreak{} so its baseline remains recommended. Pen RGB error decreases \textbf{9.72 \ensuremath{\rightarrow} 9.31 px \allowbreak{}(+4.20\%)},\allowbreak{} a 0.408 px change below approximately 3 px annotation uncertainty. \hyperref[tab:D3]{Table~D3} gives full,\allowbreak{} training and later-time errors; \hyperref[fig:D1]{Figures~D1} and D2 show the native trajectories and valid-time residuals.

\subsubsection{Findings and scope}
\label{sec:4.2.3}
\phantomsection\label{html:object-findings}

\textbf{Task recovery and identification gain are distinct outcomes.} The matched native comparisons show that feasible contact execution can be restored for both baseline and fitted models. The remaining parameter changes improve one object overall,\allowbreak{} worsen another,\allowbreak{} and produce a pen change smaller than annotation uncertainty. Successful execution alone therefore cannot establish an identification benefit,\allowbreak{} and improved aggregate error does not imply improvement along every coordinate.

\textbf{Scope of the evidence.} These five single-episode studies demonstrate recorded-command object identification and contact diagnostics. Repeated simulations and temporal subsets of those recordings do not establish independent-episode generalization. MuJoCo/MJWarp agreement involves related backends; fixed mass priors,\allowbreak{} approximate geometry,\allowbreak{} effective penetration and strong damping also limit physical-parameter interpretation. The current results do not establish a causal reduction in benchmark task-score discrepancy.

\reportnotePageNine

\FloatBarrier

Complete results: \hyperref[sec:D]{Appendix~\ref*{sec:D}},\allowbreak{} \hyperref[tab:D1]{Tables~D1}--D9 and \hyperref[fig:D1]{Figures~D1}--D9.

\phantomsection\label{html:proposed}
\textbf{Future work.} VLM-based priors and agent-selected interactions are proposed extensions,\allowbreak{} not yet validated in this study.

\phantomsection\label{html:conclusion}
\section{Conclusion}
\label{sec:5}

\textbf{Task-level consistency.} We present Genie Sim Benchmark through paired real--simulation evaluation. This report details selected robot- and object-level identification evidence. Across 22 robot--task pairs spanning 16 task names and four configurations,\allowbreak{} the evaluated Scores exhibit strong correspondence,\allowbreak{} with a descriptive Pearson correlation of \textbf{0.9686} and signed simulation--real differences from \ensuremath{-}0.110 to +0.080. These results demonstrate close task-level correspondence within the evaluated conditions and support the use of simulation as an informative reference for real-robot performance.

\textbf{Physical-response evidence.} Beyond task Scores,\allowbreak{} this report examines the physical responses underlying benchmark interactions. G2 crsB identification and excitation-family closed-loop diagnostics assess coupled robot motion through position and joint-side sensor torque. A separate 32.6503 s teleoperation trajectory,\allowbreak{} excluded from mechanical identification and servo-gain search,\allowbreak{} evaluates external position targets through simulated MC command shaping,\allowbreak{} the calibrated position servo and fixed robot dynamics.\vadjust{\penalty-10000} Across both arms,\allowbreak{} per-joint position RMSE spans 0.00614--0.03522 rad and current-derived torque RMSE spans 0.73545--4.00099 N\ensuremath{\cdot}m. This single-recording validation supports transfer of the combined execution model to the evaluated motion; it does not isolate the contribution of mechanical identification or establish benchmark task-level gains. Five signature-asset studies examine representative grasping,\allowbreak{} carrying,\allowbreak{} contact and handover mechanics. Fixed-parameter replay across MuJoCo,\allowbreak{} MJWarp and PhysX exposes the influence of object parameters and contact representations,\allowbreak{} while native PhysX adaptations recover recorded interactions. Together,\allowbreak{} these studies provide complementary evidence at the execution,\allowbreak{} interaction and task levels.

\textbf{Evaluation value.} A central contribution is the connection between real-data-driven calibration and measurable simulation behavior. Task outcomes establish whether an interaction succeeds; motion,\allowbreak{} torque and object observations characterize how closely its physical response follows reality. Evaluating both distinguishes successful execution from trajectory alignment and makes model differences actionable for simulator development. By organizing paired task evaluation alongside targeted physical diagnostics,\allowbreak{} Genie Sim Benchmark provides an evidence-led approach to assessing and improving simulation fidelity for robot manipulation.

\clearpage
\onecolumn
\setlength{\multicolsep}{5pt}
\begin{multicols}{2}\raggedcolumns

\end{multicols}
\setlength{\multicolsep}{5pt}
\useRomanappendicesfalse
\appendices
\renewcommand{\thesubsection}{\thesection.\arabic{subsection}}
\renewcommand{\thesubsectiondis}{\arabic{subsection}.}
\phantomsection\label{html:appendix}

\section{Task inventory and pairing coverage}
\label{sec:A}
\phantomsection\label{html:task-inventory}

\begin{reporttable}
\centering
\renewcommand{\thetable}{A1}
\caption{The 16 evaluated task names represented in the paired benchmark results.}
\label{tab:A1}
{\fontsize{8.2}{10.1}\selectfont
\setlength{\tabcolsep}{3pt}
\renewcommand{\arraystretch}{1.13}
\begin{tabular}{>{\raggedright\arraybackslash}p{\dimexpr 0.3100\linewidth-2\tabcolsep\relax}>{\raggedright\arraybackslash}p{\dimexpr 0.2500\linewidth-2\tabcolsep\relax}>{\raggedright\arraybackslash}p{\dimexpr 0.4400\linewidth-2\tabcolsep\relax}}
\toprule
\rowcolor{tablehead}\textbf{Source task name} & \textbf{Editorial family} & \textbf{Evidence status} \\
\midrule
Grasp Target Object & Selection \& reasoning & Score available \\
\rowcolor{tablerow}Select by Color & Selection \& reasoning & Score available \\
Recognize Size & Selection \& reasoning & Score available \\
\rowcolor{tablerow}Stack Blocks & Stacking & Score available \\
Fold Towel & Textile manipulation & Score available \\
\rowcolor{tablerow}Fold The Clothes In Half & Textile manipulation & Score available \\
Folding Clothes & Textile manipulation & Score available \\
\rowcolor{tablerow}Place Object Into Box Color & Placement \& sorting & Score available \\
Pack In Supermarket & Placement \& sorting & Score available \\
\rowcolor{tablerow}Sort Fruit & Placement \& sorting & Score available \\
Place Block Into Drawer & Placement \& sorting & Score available \\
\rowcolor{tablerow}Bimanual Chip Handover & Bimanual handover & Score available \\
Apply Common Sense & Selection \& reasoning & Score available \\
\rowcolor{tablerow}Follow Logic \allowbreak{}(OR) & Selection \& reasoning & Score available \\
Place Bottle On Coaster & Placement \& sorting & Score available \\
\rowcolor{tablerow}Fold Cloth & Textile manipulation & Score available \\
\bottomrule
\end{tabular}
}
\end{reporttable}

\reportneedspace{400pt}
\section{Complete Benchmark Evaluation Results}
\label{sec:B}
\phantomsection\label{html:appendix-benchmark}

\begin{reporttable}
\centering
\renewcommand{\thetable}{B1}
\caption{Final benchmark Scores for all 22 robot--task pairs. \ensuremath{\Delta} = Sim \ensuremath{-} Real. Mini-plots show the final gap on a common \ensuremath{-}0.12 to +0.12 scale,\allowbreak{} not uncertainty intervals. OP = Omnipicker; SP = Swiftpicker. Rollouts are the evaluation counts in each environment; real and simulation use equal counts. Scores and gaps in \hyperref[tab:B1]{Tables~B1}--B2 are displayed to two decimals; calculations use full precision.}
\label{tab:B1}
\phantomsection\label{html:benchmark-table}
{\fontsize{8.2}{10.1}\selectfont
\setlength{\tabcolsep}{3pt}
\renewcommand{\arraystretch}{1.13}
\begin{tabular}{>{\raggedright\arraybackslash}p{\dimexpr 0.1200\linewidth-2\tabcolsep\relax}>{\raggedright\arraybackslash}p{\dimexpr 0.3100\linewidth-2\tabcolsep\relax}>{\raggedright\arraybackslash}p{\dimexpr 0.1000\linewidth-2\tabcolsep\relax}>{\raggedright\arraybackslash}p{\dimexpr 0.1000\linewidth-2\tabcolsep\relax}>{\raggedright\arraybackslash}p{\dimexpr 0.2500\linewidth-2\tabcolsep\relax}>{\raggedright\arraybackslash}p{\dimexpr 0.1200\linewidth-2\tabcolsep\relax}}
\toprule
\rowcolor{tablehead}\textbf{Config.} & \textbf{Task} & \textbf{Real} & \textbf{Sim} & \textbf{\ensuremath{\Delta} Score} & \textbf{Rollouts} \\
\midrule
G1 OP & Grasp Target Object & 0.58 & 0.54 & -0.04\hfill\raisebox{-1.4pt}{\includegraphics[width=44pt]{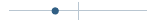}} & 50 \\
\rowcolor{tablerow}G1 OP & Select by Color & 0.73 & 0.75 & +0.02\hfill\raisebox{-1.4pt}{\includegraphics[width=44pt]{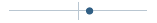}} & 50 \\
G1 OP & Recognize Size & 0.75 & 0.75 & +0.00\hfill\raisebox{-1.4pt}{\includegraphics[width=44pt]{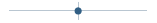}} & 50 \\
\rowcolor{tablerow}G1 OP & Place Object Into Box Color & 0.40 & 0.45 & +0.05\hfill\raisebox{-1.4pt}{\includegraphics[width=44pt]{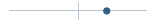}} & 50 \\
G1 OP & Pack In Supermarket & 0.95 & 1.00 & +0.05\hfill\raisebox{-1.4pt}{\includegraphics[width=44pt]{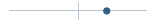}} & 50 \\
\rowcolor{tablerow}G1 OP & Sort Fruit & 1.00 & 0.90 & -0.10\hfill\raisebox{-1.4pt}{\includegraphics[width=44pt]{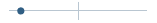}} & 50 \\
G1 OP & Place Block Into Drawer & 0.90 & 0.90 & +0.00\hfill\raisebox{-1.4pt}{\includegraphics[width=44pt]{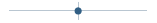}} & 50 \\
\rowcolor{tablerow}G1 OP & Bimanual Chip Handover & 0.71 & 0.70 & -0.01\hfill\raisebox{-1.4pt}{\includegraphics[width=44pt]{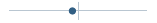}} & 50 \\
G2 SP & Grasp Target Object & 0.83 & 0.82 & -0.02\hfill\raisebox{-1.4pt}{\includegraphics[width=44pt]{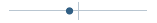}} & 100 \\
\rowcolor{tablerow}G2 SP & Select by Color & 0.86 & 0.76 & -0.10\hfill\raisebox{-1.4pt}{\includegraphics[width=44pt]{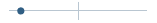}} & 100 \\
G2 SP & Recognize Size & 0.42 & 0.47 & +0.05\hfill\raisebox{-1.4pt}{\includegraphics[width=44pt]{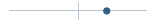}} & 50 \\
\rowcolor{tablerow}G2 SP & Apply Common Sense & 0.49 & 0.46 & -0.03\hfill\raisebox{-1.4pt}{\includegraphics[width=44pt]{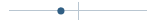}} & 100 \\
G2 SP & Follow Logic \allowbreak{}(OR) & 0.74 & 0.73 & -0.01\hfill\raisebox{-1.4pt}{\includegraphics[width=44pt]{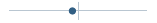}} & 100 \\
\rowcolor{tablerow}G2 SP & Fold Cloth & 0.30 & 0.33 & +0.03\hfill\raisebox{-1.4pt}{\includegraphics[width=44pt]{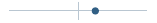}} & 30 \\
G2 Wuji & Place Bottle On Coaster & 0.68 & 0.73 & +0.05\hfill\raisebox{-1.4pt}{\includegraphics[width=44pt]{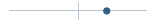}} & 30 \\
\rowcolor{tablerow}Arx X5 & Grasp Target Object & 0.85 & 0.79 & -0.06\hfill\raisebox{-1.4pt}{\includegraphics[width=44pt]{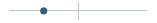}} & 100 \\
Arx X5 & Select by Color & 0.91 & 0.90 & -0.01\hfill\raisebox{-1.4pt}{\includegraphics[width=44pt]{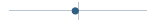}} & 100 \\
\rowcolor{tablerow}Arx X5 & Recognize Size & 0.77 & 0.66 & -0.11\hfill\raisebox{-1.4pt}{\includegraphics[width=44pt]{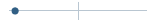}} & 50 \\
Arx X5 & Stack Blocks & 0.24 & 0.24 & +0.00\hfill\raisebox{-1.4pt}{\includegraphics[width=44pt]{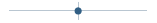}} & 30 \\
\rowcolor{tablerow}Arx X5 & Fold Towel & 0.82 & 0.86 & +0.04\hfill\raisebox{-1.4pt}{\includegraphics[width=44pt]{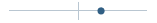}} & 50 \\
Arx X5 & Fold The Clothes In Half & 0.84 & 0.80 & -0.04\hfill\raisebox{-1.4pt}{\includegraphics[width=44pt]{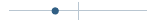}} & 50 \\
\rowcolor{tablerow}Arx X5 & Folding Clothes & 0.72 & 0.80 & +0.08\hfill\raisebox{-1.4pt}{\includegraphics[width=44pt]{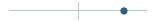}} & 50 \\
\bottomrule
\end{tabular}
}
\end{reporttable}

\phantomsection\label{html:benchmark-post-sysid}

\begin{reporttable}
\centering
\renewcommand{\thetable}{B2}
\caption{All 14 G2/Arx before/after identification or calibration comparisons. The Real reference is the same in the two source columns. Change in |gap| = |Sim \ensuremath{-} Real|after \ensuremath{-} |Sim \ensuremath{-} Real|before; negative denotes a smaller discrepancy. Mini-plots share the \ensuremath{-}0.12 to +0.12 scale of \hyperref[tab:B1]{Table~B1}: open circles show before,\allowbreak{} filled circles after.}
\label{tab:B2}
{\fontsize{8.2}{10.1}\selectfont
\setlength{\tabcolsep}{3pt}
\renewcommand{\arraystretch}{1.13}
\begin{tabular}{>{\raggedright\arraybackslash}p{\dimexpr 0.0900\linewidth-2\tabcolsep\relax}>{\raggedright\arraybackslash}p{\dimexpr 0.2500\linewidth-2\tabcolsep\relax}>{\raggedright\arraybackslash}p{\dimexpr 0.0700\linewidth-2\tabcolsep\relax}>{\raggedright\arraybackslash}p{\dimexpr 0.1700\linewidth-2\tabcolsep\relax}>{\raggedright\arraybackslash}p{\dimexpr 0.2800\linewidth-2\tabcolsep\relax}>{\raggedright\arraybackslash}p{\dimexpr 0.1400\linewidth-2\tabcolsep\relax}}
\toprule
\rowcolor{tablehead}\textbf{Config.} & \textbf{Task} & \textbf{Real} & \textbf{Sim before \ensuremath{\rightarrow} after} & \textbf{Gap before \ensuremath{\rightarrow} after} & \textbf{Change in |gap|} \\
\midrule
G2 SP & Grasp Target Object & 0.83 & 0.81 \ensuremath{\rightarrow} 0.82 & -0.02 \ensuremath{\rightarrow} -0.02\hfill\raisebox{-1.4pt}{\includegraphics[width=44pt]{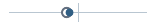}} & -0.01 \\
\rowcolor{tablerow}G2 SP & Select by Color & 0.86 & 0.75 \ensuremath{\rightarrow} 0.76 & -0.11 \ensuremath{\rightarrow} -0.10\hfill\raisebox{-1.4pt}{\includegraphics[width=44pt]{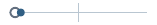}} & -0.01 \\
G2 SP & Recognize Size & 0.42 & 0.47 \ensuremath{\rightarrow} 0.47 & +0.05 \ensuremath{\rightarrow} +0.05\hfill\raisebox{-1.4pt}{\includegraphics[width=44pt]{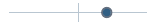}} & +0.00 \\
\rowcolor{tablerow}G2 SP & Apply Common Sense & 0.49 & 0.43 \ensuremath{\rightarrow} 0.46 & -0.06 \ensuremath{\rightarrow} -0.03\hfill\raisebox{-1.4pt}{\includegraphics[width=44pt]{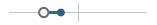}} & -0.03 \\
G2 SP & Follow Logic \allowbreak{}(OR) & 0.74 & 0.73 \ensuremath{\rightarrow} 0.73 & -0.01 \ensuremath{\rightarrow} -0.01\hfill\raisebox{-1.4pt}{\includegraphics[width=44pt]{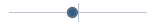}} & +0.00 \\
\rowcolor{tablerow}G2 SP & Fold Cloth & 0.30 & 0.40 \ensuremath{\rightarrow} 0.33 & +0.10 \ensuremath{\rightarrow} +0.03\hfill\raisebox{-1.4pt}{\includegraphics[width=44pt]{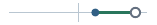}} & -0.07 \\
G2 Wuji & Place Bottle On Coaster & 0.68 & 0.76 \ensuremath{\rightarrow} 0.73 & +0.08 \ensuremath{\rightarrow} +0.05\hfill\raisebox{-1.4pt}{\includegraphics[width=44pt]{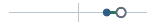}} & -0.03 \\
\rowcolor{tablerow}Arx X5 & Grasp Target Object & 0.85 & 0.79 \ensuremath{\rightarrow} 0.79 & -0.06 \ensuremath{\rightarrow} -0.06\hfill\raisebox{-1.4pt}{\includegraphics[width=44pt]{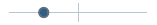}} & +0.00 \\
Arx X5 & Select by Color & 0.91 & 0.90 \ensuremath{\rightarrow} 0.90 & -0.01 \ensuremath{\rightarrow} -0.01\hfill\raisebox{-1.4pt}{\includegraphics[width=44pt]{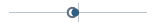}} & -0.01 \\
\rowcolor{tablerow}Arx X5 & Recognize Size & 0.77 & 0.69 \ensuremath{\rightarrow} 0.66 & -0.08 \ensuremath{\rightarrow} -0.11\hfill\raisebox{-1.4pt}{\includegraphics[width=44pt]{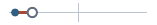}} & +0.03 \\
Arx X5 & Stack Blocks & 0.24 & 0.24 \ensuremath{\rightarrow} 0.24 & +0.00 \ensuremath{\rightarrow} +0.00\hfill\raisebox{-1.4pt}{\includegraphics[width=44pt]{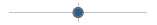}} & +0.00 \\
\rowcolor{tablerow}Arx X5 & Fold Towel & 0.82 & 0.88 \ensuremath{\rightarrow} 0.86 & +0.06 \ensuremath{\rightarrow} +0.04\hfill\raisebox{-1.4pt}{\includegraphics[width=44pt]{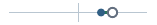}} & -0.02 \\
Arx X5 & Fold The Clothes In Half & 0.84 & 0.80 \ensuremath{\rightarrow} 0.80 & -0.04 \ensuremath{\rightarrow} -0.04\hfill\raisebox{-1.4pt}{\includegraphics[width=44pt]{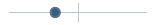}} & +0.00 \\
\rowcolor{tablerow}Arx X5 & Folding Clothes & 0.72 & 0.82 \ensuremath{\rightarrow} 0.80 & +0.10 \ensuremath{\rightarrow} +0.08\hfill\raisebox{-1.4pt}{\includegraphics[width=44pt]{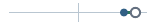}} & -0.02 \\
\bottomrule
\end{tabular}
}
\end{reporttable}
\begin{multicols}{2}\raggedcolumns
Result definitions and reporting convention. \hyperref[tab:B1]{Table~B1} reports final Scores; \hyperref[tab:B2]{Table~B2} separates all 14 G2/Arx before/after conditions. Stage-rate records retain their own definitions and are excluded from the final paired correspondence statistics.

Score is task-specific and distinct from the stage-level rates in \hyperref[fig:E1]{Figure~E1}. \hyperref[fig:1]{Figure~1} displays the 22 final pairs in \hyperref[tab:B1]{Table~B1},\allowbreak{} with all 14 G2/Arx comparisons in \hyperref[tab:B2]{Table~B2}.
\end{multicols}

\reportneedspace{320pt}
\section{G2 crsB identification evidence}
\label{sec:C}
\begin{multicols}{2}\raggedcolumns
\phantomsection\label{html:appendix-g2}

\subsection{Evaluation data and metric conventions}
\end{multicols}
\label{sec:C.1}
\begin{multicols}{2}\raggedcolumns
\phantomsection\label{html:appendix-g2-scope}

Each arm has 18 identification recordings: seven single-joint friction recordings,\allowbreak{} ten sine recordings and one coupled Fourier recording. J1--J3 sine recordings cover two background postures; J4--J7 cover one. The three-stage mechanical-identification method is described in \hyperref[sec:4.1]{Section~\ref*{sec:4.1}} and is not repeated here.

The recorded-data summaries use 3,\allowbreak{}696 scored time rows per arm: 25,\allowbreak{}872 all-joint rows and 9,\allowbreak{}684 driven-joint rows. \hyperref[tab:C1]{Tables~C1}--C2 retain the complete joint/family breakdown,\allowbreak{} including held joints. These are correlated recording diagnostics,\allowbreak{} not independent trials. Centred residuals remove one constant per recording and joint; their RMS and pooled R\textsuperscript{2} are not validation-set scores.

Friction conventions differ across the archived evaluations. Mechanical identification uses tanh\allowbreak{}(v/C\ensuremath{_0}),\allowbreak{} where v is joint speed and C\ensuremath{_0} = 0.001 rad/s. The recorded-data scoring harness uses sign\allowbreak{}(v) for |v| \ensuremath{\geq} 0.02 rad/s and zero otherwise. The joint-side sensor diagnostics use motor-side sign\allowbreak{}(v),\allowbreak{} with sign\allowbreak{}(0) = 0 and no finite dead zone. Constant offsets remain in forward simulation; sensor errors are evaluated without centring.
\end{multicols}
\begin{reporttable}
\centering
\renewcommand{\thetable}{C1}
\caption{Per-joint centred torque residuals; all recorded motions,\allowbreak{} including unexcited rows.}
\label{tab:C1}
{\fontsize{8.2}{10.1}\selectfont
\setlength{\tabcolsep}{3pt}
\renewcommand{\arraystretch}{1.13}
\begin{tabular}{>{\raggedright\arraybackslash}p{\dimexpr 0.1200\linewidth-2\tabcolsep\relax}>{\raggedright\arraybackslash}p{\dimexpr 0.2200\linewidth-2\tabcolsep\relax}>{\raggedright\arraybackslash}p{\dimexpr 0.2200\linewidth-2\tabcolsep\relax}>{\raggedright\arraybackslash}p{\dimexpr 0.2200\linewidth-2\tabcolsep\relax}>{\raggedright\arraybackslash}p{\dimexpr 0.2200\linewidth-2\tabcolsep\relax}}
\toprule
\rowcolor{tablehead}\textbf{Joint} & \textbf{Left RMS \allowbreak{}(N\ensuremath{\cdot}m)} & \textbf{Left R\textsuperscript{2}} & \textbf{Right RMS \allowbreak{}(N\ensuremath{\cdot}m)} & \textbf{Right R\textsuperscript{2}} \\
\midrule
J1 & 1.0177 & 0.9814 & 1.0415 & 0.9698 \\
\rowcolor{tablerow}J2 & 1.0885 & 0.9872 & 1.2345 & 0.9836 \\
J3 & 0.4120 & 0.9783 & 0.4808 & 0.9707 \\
\rowcolor{tablerow}J4 & 0.4367 & 0.9745 & 0.5697 & 0.9563 \\
J5 & 0.3360 & 0.9174 & 0.3608 & 0.9172 \\
\rowcolor{tablerow}J6 & 0.4225 & 0.9410 & 0.3308 & 0.9273 \\
J7 & 0.3518 & 0.9370 & 0.3086 & 0.9473 \\
\bottomrule
\end{tabular}
}
\end{reporttable}
\begin{reporttable}
\centering
\renewcommand{\thetable}{C2}
\caption{Excitation-family breakdown. Joint samples are correlated time-series rows,\allowbreak{} not independent trials.}
\label{tab:C2}
{\fontsize{8.2}{10.1}\selectfont
\setlength{\tabcolsep}{3pt}
\renewcommand{\arraystretch}{1.13}
\begin{tabular}{>{\raggedright\arraybackslash}p{\dimexpr 0.1400\linewidth-2\tabcolsep\relax}>{\raggedright\arraybackslash}p{\dimexpr 0.2000\linewidth-2\tabcolsep\relax}>{\raggedright\arraybackslash}p{\dimexpr 0.1800\linewidth-2\tabcolsep\relax}>{\raggedright\arraybackslash}p{\dimexpr 0.1400\linewidth-2\tabcolsep\relax}>{\raggedright\arraybackslash}p{\dimexpr 0.1600\linewidth-2\tabcolsep\relax}>{\raggedright\arraybackslash}p{\dimexpr 0.1800\linewidth-2\tabcolsep\relax}}
\toprule
\rowcolor{tablehead}\textbf{Excitation} & \textbf{Arm} & \textbf{Joint samples} & \textbf{RMS \allowbreak{}(N\ensuremath{\cdot}m)} & \textbf{R\textsuperscript{2}} & \textbf{Max |error| \allowbreak{}(N\ensuremath{\cdot}m)} \\
\midrule
friction & left & 15386 & 0.6144 & 0.9413 & 6.2807 \\
\rowcolor{tablerow}friction & right & 15386 & 0.5752 & 0.9516 & 5.9243 \\
sine & left & 3500 & 0.5665 & 0.9943 & 3.7197 \\
\rowcolor{tablerow}sine & right & 3500 & 0.6057 & 0.9930 & 4.8681 \\
fourier & left & 6986 & 0.7700 & 0.9874 & 7.6106 \\
\rowcolor{tablerow}fourier & right & 6986 & 0.9689 & 0.9795 & 7.3545 \\
\bottomrule
\end{tabular}
}
\end{reporttable}
\vspace{5pt}
\begin{multicols}{2}\raggedcolumns
\hyperref[fig:C1]{Figures~C1}--C2 retain the left-arm friction and sine recording diagnostics,\allowbreak{} showing source-designated driven joints only. Displayed curves are centred per recording/joint; Own RMS labels use full-row residuals with the identified constant offset. The displayed curves and labels therefore use different offset conventions.
\end{multicols}
\reportnotePageThirteen

\begin{reportfigure}
\centering
\renewcommand{\thefigure}{C1}
\includegraphics[width=\linewidth,height=.69\textheight,keepaspectratio]{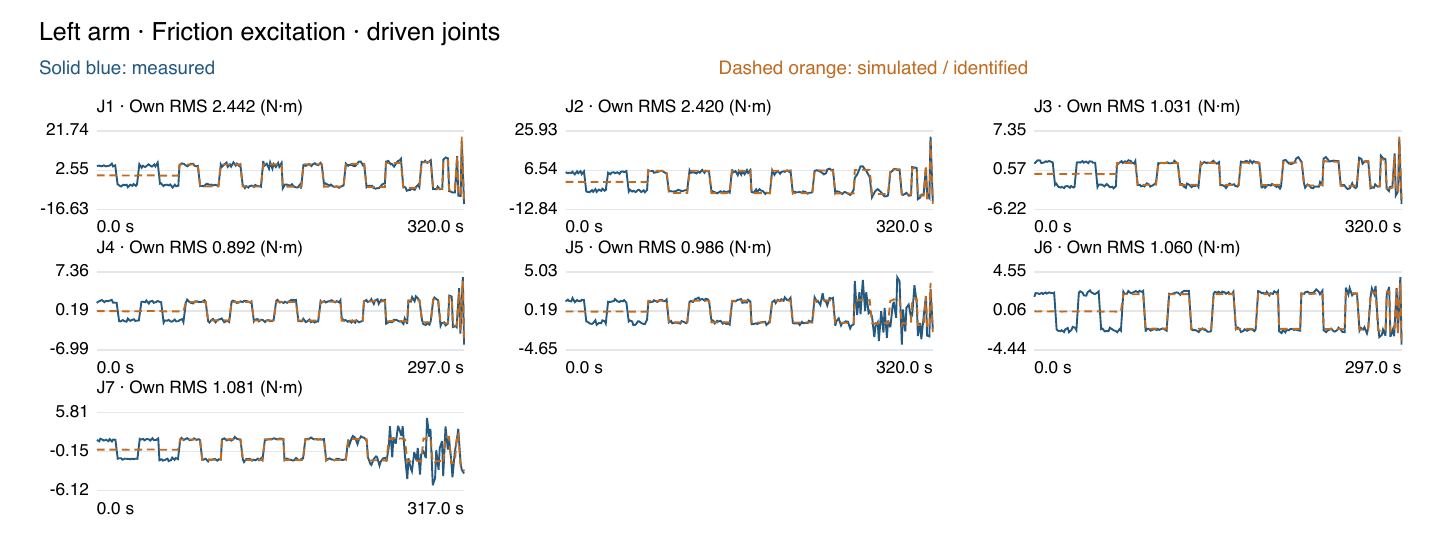}
\caption{Left-arm friction recording diagnostics,\allowbreak{} driven joints only. Curves use recording/joint centring; Own RMS labels use full-row own-offset scores. These are fitting-recording diagnostics.}
\label{fig:C1}
\phantomsection\label{html:g2-panel-2}
\end{reportfigure}
\begin{reportfigure}
\centering
\renewcommand{\thefigure}{C2}
\includegraphics[width=\linewidth,height=.69\textheight,keepaspectratio]{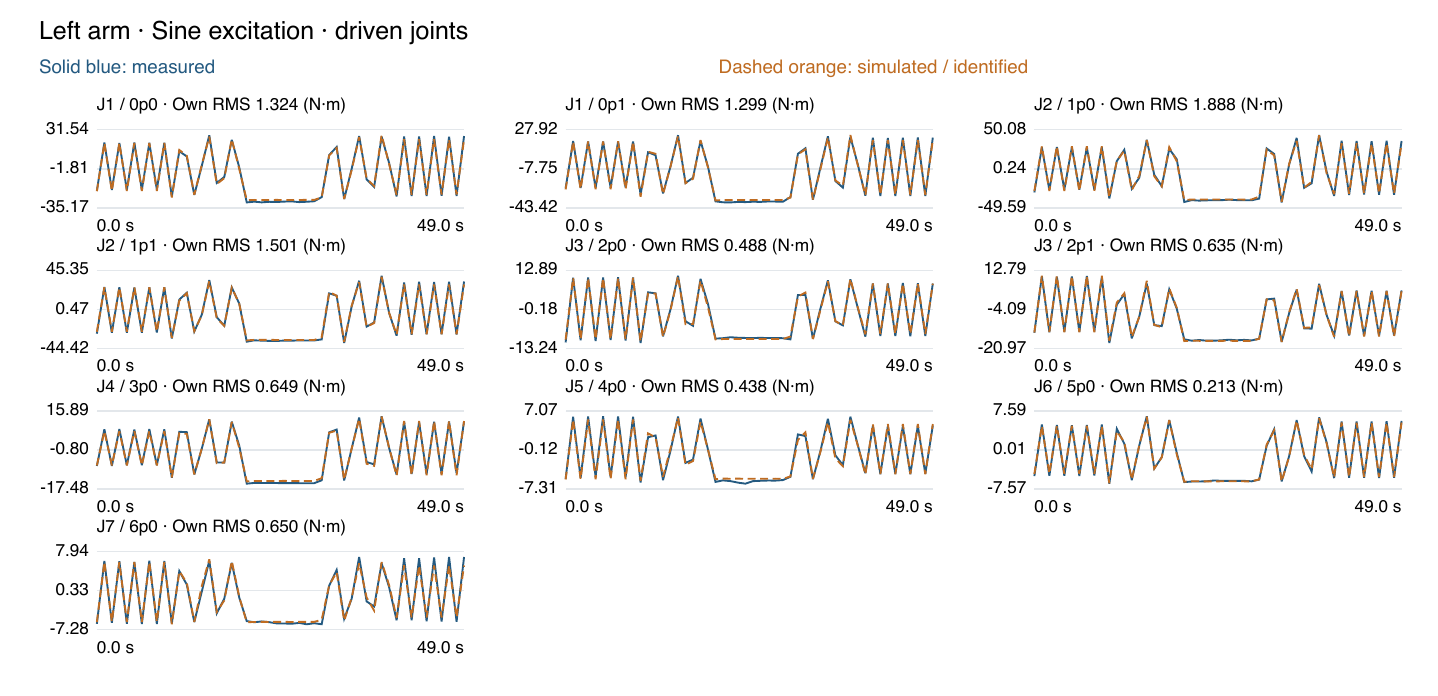}
\caption{Left-arm sine recording diagnostics,\allowbreak{} driven joints only. All background postures and driven joints are retained. Curves use recording/joint centring; Own RMS labels use full-row own-offset scores.}
\label{fig:C2}
\phantomsection\label{html:g2-panel-3}
\end{reportfigure}
\reportnotePageFourteen

\reportneedspace{320pt}
\subsection{Full-recording Fourier reconstruction}

\label{sec:C.2}
\begin{multicols}{2}\raggedcolumns
\phantomsection\label{html:appendix-g2-fourier}

The approximately 100 s Fourier recording is evaluated over its full recorded interval. \hyperref[fig:C3]{Figure~C3} shows all seven driven left-arm joints,\allowbreak{} comparing current-derived torque with inverse-dynamics reconstruction. Curves use recording/joint centring; Own RMS labels retain full-row own-offset scores. This is a same-recording reconstruction diagnostic,\allowbreak{} not an independent holdout test. \hyperref[tab:C1]{Tables~C1}--C2 retain both arms and their complete metric denominators.
\end{multicols}
\begin{reportfigure}
\centering
\renewcommand{\thefigure}{C3}
\includegraphics[width=\linewidth,height=.69\textheight,keepaspectratio]{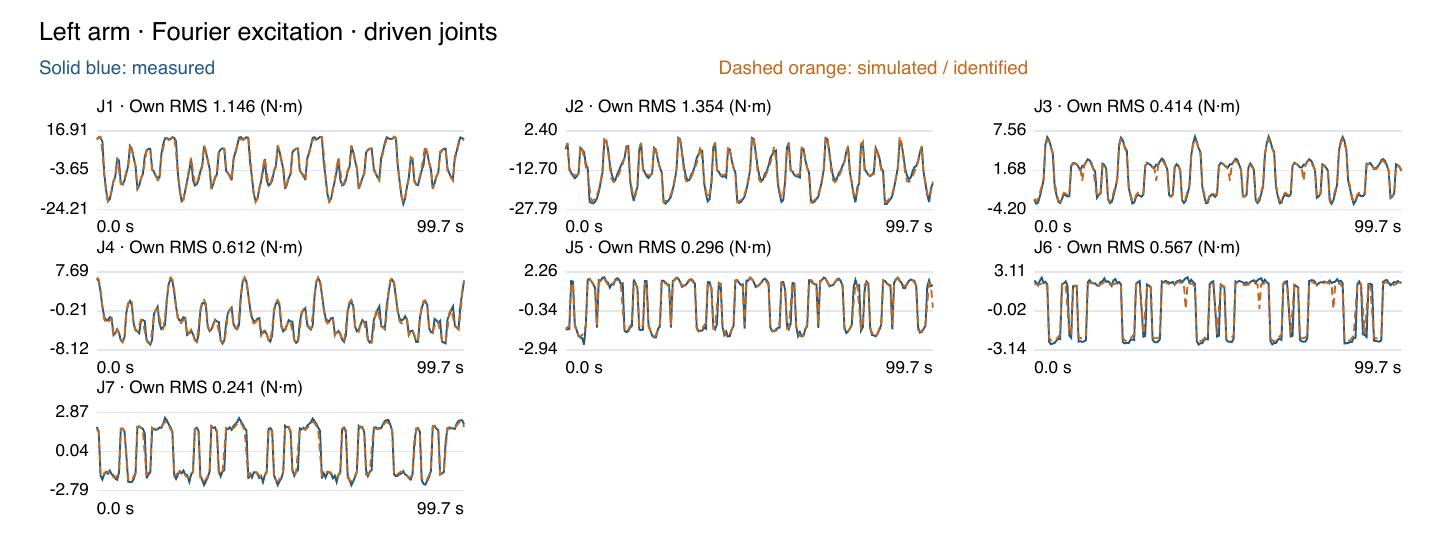}
\caption{Full-recording Fourier torque reconstruction,\allowbreak{} left-arm J1--J7,\allowbreak{} approximately 100 s \allowbreak{}(display samples span 0--99.7 s). Solid blue: current-derived measured torque; dashed orange: identified model. Curves use recording/joint centring; Own RMS labels use full-row own-offset scores. These are same-recording diagnostics.}
\label{fig:C3}
\phantomsection\label{html:g2-panel-1}
\end{reportfigure}
\reportneedspace{320pt}
\subsection{Independent teleoperation validation}

\label{sec:C.3}
\begin{multicols}{2}\raggedcolumns
\phantomsection\label{html:appendix-g2-teleoperation}
\phantomsection\label{html:g2-teleoperation-execution}
\par\smallskip\noindent\colorbox{reviewyellow}{\parbox{\dimexpr\linewidth-2\fboxsep\relax}{\normalsize The teleoperation replay uses external position targets,\allowbreak{} MC command shaping and the joint PD law in \hyperref[sec:4.1.4]{Section~\ref*{sec:4.1.4}}. Both arms use separate identified gains \allowbreak{}(\hyperref[tab:C3]{Table~C3}). The mechanical model,\allowbreak{} armature and torque-bias parameters remain fixed. Native MuJoCo Coulomb frictionloss and linear damping are active; this replay does not evaluate the exact fitted smooth friction function.}}\par\smallskip

\phantomsection\label{html:g2-teleoperation-mc-settings}
\par\smallskip\noindent\colorbox{reviewyellow}{\parbox{\dimexpr\linewidth-2\fboxsep\relax}{\normalsize MC settings. Position--velocity--time \allowbreak{}(PVT) interpolation,\allowbreak{} configured position/velocity/acceleration and coupling limits,\allowbreak{} and a second-order 5 Hz Butterworth low-pass filter are enabled. The optional joint-delay filter and position-lead compensation are disabled. The documented command publication order includes one control-tick staging. Simulation control and physics run at 1 kHz \allowbreak{}(1 ms steps). The MC control-period scale is 1.1; the validation summary records an 11 ms command period. This command-period setting is distinct from the 1 ms simulation step.}}\par\smallskip

\phantomsection\label{html:g2-teleoperation-sampling}
\par\smallskip\noindent\colorbox{reviewyellow}{\parbox{\dimexpr\linewidth-2\fboxsep\relax}{\normalsize Sampling and scoring. Recorded external targets average 90.625 Hz; recorded joint commands and states are approximately 100 Hz. Simulated position and motor torque are linearly interpolated to measured receive timestamps,\allowbreak{} with no further centring or filtering of the published comparison. Torque compares current converted to joint-side units with simulated PD motor torque; it is not the joint-side sensor channel used in the older diagnostics. All 3,\allowbreak{}265 aligned rows are shown,\allowbreak{} while RMSE uses 3,\allowbreak{}165 rows after the first 1 s.}}\par\smallskip

\phantomsection\label{html:g2-teleoperation-implementation-pending}
\par\smallskip\noindent\colorbox{reviewyellow}{\parbox{\dimexpr\linewidth-2\fboxsep\relax}{\normalsize Implementation details pending. Exact limiter thresholds,\allowbreak{} coupling rules and filter recurrences/initial states,\allowbreak{} actuator-limit settings,\allowbreak{} full-robot initialisation and teleoperation hand/payload configuration still require executable configuration records. The available summary and method description do not establish these details.}}\par\smallskip

\hyperref[tab:C3]{Table~C3} reports raw,\allowbreak{} uncentred per-joint position RMSE in radians and simulated motor-side joint torque RMSE against current-derived joint torque in N\ensuremath{\cdot}m. Both arms are retained; no joint-side or end-effector force sensor correction is applied. There are no independent repeated real trials or uncertainty intervals in the supplied results.
\end{multicols}
\reportnotePageFifteen

\par\noindent\begin{minipage}{\textwidth}
\begin{reporttable}
\colorbox{reviewyellow}{\parbox{\dimexpr\linewidth-2\fboxsep\relax}{
\centering
\renewcommand{\thetable}{C3}
\caption{Independent teleoperation replay,\allowbreak{} 32.6503 s: frozen PD gains and all fourteen arm/joint results. Position/torque RMSE units are rad/N\ensuremath{\cdot}m; Kp/Kd units are N\ensuremath{\cdot}m/rad and N\ensuremath{\cdot}m\ensuremath{\cdot}s/rad. Five-decimal RMSE uses 3,\allowbreak{}165 aligned measured-time rows after the first 1 s of warmup,\allowbreak{} without centring or pooling. These gains belong to this teleoperation servo,\allowbreak{} not the older sensor-diagnostic controller.}
\label{tab:C3}
\phantomsection\label{html:g2-teleoperation-gains}
{\fontsize{8.2}{10.1}\selectfont
\setlength{\tabcolsep}{5pt}
\renewcommand{\arraystretch}{1.10}
\begin{tabular}{lrrrrrrrr}
\toprule
Joint & \multicolumn{4}{c}{Left arm} & \multicolumn{4}{c}{Right arm} \\
 & $K_p$ & $K_d$ & $q$ RMSE & $\tau$ RMSE & $K_p$ & $K_d$ & $q$ RMSE & $\tau$ RMSE \\
\midrule
J1 & 925 & 177.5 & 0.01377 & 2.46279 & 1375 & 267.5 & 0.01914 & 4.00099 \\
J2 & 850 & 307.5 & 0.00881 & 2.11839 & 975 & 382.5 & 0.00614 & 1.95315 \\
J3 & 400 & 70 & 0.00945 & 0.73545 & 600 & 92.5 & 0.00933 & 0.97216 \\
J4 & 375 & 85 & 0.01832 & 1.23392 & 675 & 135 & 0.02902 & 1.92842 \\
J5 & 150 & 35 & 0.01819 & 1.28795 & 200 & 45 & 0.02057 & 1.52118 \\
J6 & 200 & 30 & 0.01777 & 1.00963 & 225 & 37.5 & 0.01990 & 1.13422 \\
J7 & 200 & 32.5 & 0.03522 & 1.36333 & 225 & 45 & 0.02067 & 0.87570 \\
\bottomrule
\end{tabular}
}
}}
\end{reporttable}
\begin{reportfigure}
\centering
\renewcommand{\thefigure}{C4}
\includegraphics[width=\linewidth,height=.59\textheight,keepaspectratio]{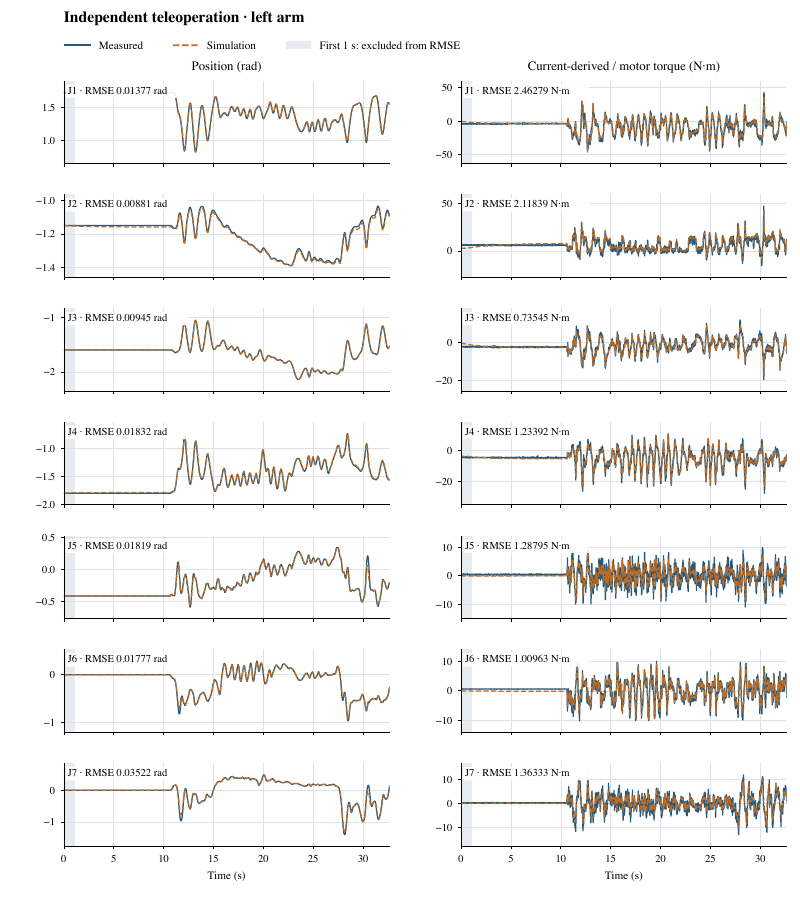}
\caption{Independent teleoperation validation,\allowbreak{} left-arm J1--J7: position and torque over the complete approximately 32.65 s recording. Measured traces are blue/solid; simulation is orange/dashed. Torque compares current-derived measurements with simulated PD motor torque in joint-side units. All 3,\allowbreak{}265 aligned rows are displayed; RMSE excludes the grey first 1 s and uses 3,\allowbreak{}165 rows. Right-arm results remain in \hyperref[tab:C3]{Table~C3}.}
\label{fig:C4}
\phantomsection\label{html:g2-teleoperation-validation}
\end{reportfigure}
\end{minipage}\par
\clearpage
\subsection{Joint-side sensor diagnostics under the original configuration}

\label{sec:C.4}

\begin{multicols}{2}\raggedcolumns
\fontsize{10}{11.2}\selectfont
\phantomsection\label{html:appendix-g2-sensors}

These sensor diagnostics retain their separately frozen,\allowbreak{} gripper-free full-robot configuration. They use joint-side sensor torque rather than the current-derived torque of \hyperref[sec:C.3]{Section~\ref*{sec:C.3}}. Both grippers are removed; remaining degrees of freedom remain live,\allowbreak{} with XML-rest non-arm targets and a compliant base fixture. The original controller uses Kp = 1000 N\ensuremath{\cdot}m/rad and Kd = [1,\allowbreak{}5,\allowbreak{}1,\allowbreak{}1,\allowbreak{}1,\allowbreak{}1,\allowbreak{}1] N\ensuremath{\cdot}m\ensuremath{\cdot}s/rad,\allowbreak{} with damping on simulated velocity and no target-velocity term. \hyperref[tab:D1]{Table~D1} retains these gains alongside the object-replay settings.

At each 1 ms step,\allowbreak{} the controller applies position-error torque minus velocity damping. No explicit drive saturation is applied. Motor-side Coulomb friction and the fixed offset are subtracted once from applied force; viscous friction is installed once as joint damping and armature once as joint inertia. The six live base coordinates use fixture gains of 200000 N/m and 20000 N\ensuremath{\cdot}s/m for translation,\allowbreak{} and 200000 N\ensuremath{\cdot}m/rad and 20000 N\ensuremath{\cdot}m\ensuremath{\cdot}s/rad for rotation,\allowbreak{} relative to the initial base state. These are simulation boundary assumptions.

Torque readback. Before integration,\allowbreak{} simulated link torque is the arm components of qfrc\_\allowbreak{}actuator + qfrc\_\allowbreak{}applied + qfrc\_\allowbreak{}passive minus armature times simulated joint acceleration. These are MuJoCo generalized-force arrays. Sensor-side sign or memory terms are added only to predicted readback and never to plant force or control. Observation-model changes leave full states,\allowbreak{} control and forces bitwise identical. These examined recordings are not a blind independent episode test.
\end{multicols}
\begingroup
\setlength{\reporttablebefore}{10pt}\setlength{\reporttableafter}{5pt}
\begin{reporttable}
\centering
\renewcommand{\thetable}{C4}
\caption{Uncentred prediction errors in the gripper-free full-robot configuration. Ten-second windows,\allowbreak{} 1 ms physics. Frequency-group sensor transfer is separately reported below; its windows differ from this original-window prediction.}
\label{tab:C4}
{\fontsize{8.2}{10.1}\selectfont
\setlength{\tabcolsep}{3pt}
\renewcommand{\arraystretch}{1.13}
\begin{tabular}{>{\raggedright\arraybackslash}p{\dimexpr 0.2000\linewidth-2\tabcolsep\relax}>{\raggedright\arraybackslash}p{\dimexpr 0.0800\linewidth-2\tabcolsep\relax}>{\raggedright\arraybackslash}p{\dimexpr 0.1400\linewidth-2\tabcolsep\relax}>{\raggedright\arraybackslash}p{\dimexpr 0.2000\linewidth-2\tabcolsep\relax}>{\raggedright\arraybackslash}p{\dimexpr 0.1900\linewidth-2\tabcolsep\relax}>{\raggedright\arraybackslash}p{\dimexpr 0.1900\linewidth-2\tabcolsep\relax}}
\toprule
\rowcolor{tablehead}\textbf{Window} & \textbf{Joint} & \textbf{Time rows} & \textbf{Position RMSE \allowbreak{}(deg)} & \textbf{Sensor RMSE \allowbreak{}(Nm)} & \textbf{Sensor RMS \allowbreak{}(Nm)} \\
\midrule
Fourier,\allowbreak{} proximal & J1 & 10000 & 0.5406 & 0.9409 & 6.7965 \\
\rowcolor{tablerow}Fourier,\allowbreak{} proximal & J2 & 10000 & 0.8747 & 1.2271 & 15.2108 \\
Fourier,\allowbreak{} proximal & J3 & 10000 & 0.2083 & 0.3429 & 1.7336 \\
\rowcolor{tablerow}Fourier,\allowbreak{} proximal & J4 & 10000 & 0.1835 & 0.3908 & 3.5386 \\
Fourier,\allowbreak{} wrist & J5 & 10000 & 0.1825 & 0.0994 & 0.1468 \\
\rowcolor{tablerow}Fourier,\allowbreak{} wrist & J6 & 10000 & 0.0785 & 0.2813 & 0.6234 \\
Fourier,\allowbreak{} wrist & J7 & 10000 & 0.0570 & 0.0762 & 0.0892 \\
\rowcolor{tablerow}Friction,\allowbreak{} J2 & J2 & 10000 & 0.3650 & 1.2704 & 0.8130 \\
Sine,\allowbreak{} J1 & J1 & 10000 & 0.5830 & 0.4066 & 0.3600 \\
\bottomrule
\end{tabular}
}
\end{reporttable}
\begin{reporttable}
\centering
\renewcommand{\thetable}{C5}
\caption{Consolidated joint-side sensor diagnostics for wrist,\allowbreak{} speed-group and frequency tests. Sign / memory values are paired RMSEs,\allowbreak{} in that order. Position RMSE uses degrees; torque RMSE,\allowbreak{} mean error and sensor coefficients use N\ensuremath{\cdot}m. Frequency rows report full-window and separate near-zero RMSE. Speed groups and time rows are different denominators; a dash means the source table did not report that metric.}
\label{tab:C5}
\vspace{-8pt}
{\fontsize{8.2}{10.1}\selectfont
\setlength{\tabcolsep}{2.5pt}
\renewcommand{\arraystretch}{1.10}
\begin{minipage}[t]{.485\linewidth}
\textbf{Wrist and speed-group diagnostics}\par\vspace{2pt}
\begin{tabular}{>{\raggedright\arraybackslash}p{\dimexpr.18\linewidth-5pt\relax}>{\raggedright\arraybackslash}p{\dimexpr.19\linewidth-5pt\relax}>{\raggedright\arraybackslash}p{\dimexpr.26\linewidth-5pt\relax}>{\raggedright\arraybackslash}p{\dimexpr.37\linewidth-5pt\relax}}
\toprule
\rowcolor{tablehead}\textbf{Joint / check} & \textbf{Position RMSE} & \textbf{Torque RMSE sign / memory} & \textbf{Additional values / rows} \\
\midrule
\multicolumn{4}{p{\dimexpr\linewidth-5pt\relax}}{\textbf{Wrist,\allowbreak{} 10 s \ensuremath{\cdot} 10000 rows per joint \ensuremath{\cdot} sign / memory observers}} \\
J5 & 0.1825 & 0.0994 / 0.0993 & Mean -0.0830 \\
\rowcolor{tablerow}J6 & 0.0785 & 0.2813 / 0.3277 & Mean +0.2427 \\
J7 & 0.0570 & 0.0762 / 0.0790 & Mean -0.0670 \\
\multicolumn{4}{p{\dimexpr\linewidth-5pt\relax}}{\textbf{Posture-matched wrist sweeps \ensuremath{\cdot} frozen sign observer \ensuremath{\cdot} 7 training / 6 held-out speed groups per joint}} \\
\rowcolor{tablerow}J5 & --- & --- & Coeff. 0.09094 \\
J6 & --- & --- & Coeff. 0.47542 \\
\rowcolor{tablerow}J7 & --- & --- & Coeff. 0.11322 \\
\multicolumn{4}{p{\dimexpr\linewidth-5pt\relax}}{\textbf{Wrist failure check \ensuremath{\cdot} sign / memory observers}} \\
J5 near zero & --- & 0.1274 / 0.1133 & 637 rows; reversal-error cases retained \\
\rowcolor{tablerow}J6 full & --- & 0.2813 / 0.3277 & Memory worse than sign \\
J7 full & --- & 0.0762 / 0.0790 & Memory worse than sign \\
\bottomrule
\end{tabular}
\end{minipage}\hfill
\begin{minipage}[t]{.485\linewidth}
\noindent\textbf{Frequency holdout \ensuremath{\cdot} torque RMSE}\par\smallskip\begin{tabular}{>{\raggedright\arraybackslash}p{\dimexpr 0.1\linewidth-2\tabcolsep\relax}>{\raggedright\arraybackslash}p{\dimexpr 0.24\linewidth-2\tabcolsep\relax}>{\raggedright\arraybackslash}p{\dimexpr 0.12\linewidth-2\tabcolsep\relax}>{\raggedright\arraybackslash}p{\dimexpr 0.17\linewidth-2\tabcolsep\relax}>{\raggedright\arraybackslash}p{\dimexpr 0.17\linewidth-2\tabcolsep\relax}>{\raggedright\arraybackslash}p{\dimexpr 0.2\linewidth-2\tabcolsep\relax}}
\toprule
\rowcolor{tablehead}\textbf{Hz} & \textbf{Observer} & \textbf{Full rows} & \textbf{Near-zero rows} & \textbf{Full RMSE} & \textbf{Near-zero RMSE} \\
\midrule
0.7 & sign & 1000 & 6 & 0.3104 & 0.4697 \\
\rowcolor{tablerow}0.7 & smooth & 1000 & 6 & 0.3072 & 0.3843 \\
0.7 & lag\_\allowbreak{}memory & 1000 & 6 & 0.2712 & 0.2027 \\
\rowcolor{tablerow}0.7 & schmitt\_\allowbreak{}memory & 1000 & 6 & 0.3056 & 0.1879 \\
1.3 & sign & 1000 & 9 & 0.2737 & 0.1330 \\
\rowcolor{tablerow}1.3 & smooth & 1000 & 9 & 0.2751 & 0.0937 \\
1.3 & lag\_\allowbreak{}memory & 1000 & 9 & 0.3625 & 0.0921 \\
\rowcolor{tablerow}1.3 & schmitt\_\allowbreak{}memory & 1000 & 9 & 0.3091 & 0.1330 \\
\bottomrule
\end{tabular}\par\smallskip Position RMSE is not reported for the speed and frequency rows. The wrist-failure rows report torque errors; no additional position RMSE is supplied.
\end{minipage}
}
\end{reporttable}
\endgroup
\begin{multicols}{2}\raggedcolumns
Wrist sensor coefficients use seven training and six held-out speed groups per joint,\allowbreak{} matched within 1 degree across all seven arm angles,\allowbreak{} with no intercept. Near-zero checks use measured |joint speed| < 0.02 rad/s. J5 reversal errors remain; wrist memory worsens full-window J6/J7 error. The J1 lag model improves 0.7 Hz and degrades 1.3 Hz; only six and nine near-zero rows are available,\allowbreak{} respectively. Mounting,\allowbreak{} filtering and zeroing are incompletely documented,\allowbreak{} so these residuals have no uniquely established mechanical cause.
\end{multicols}
\begin{reportfigure}
\centering
\renewcommand{\thefigure}{C5}
\includegraphics[width=\linewidth,height=.69\textheight,keepaspectratio]{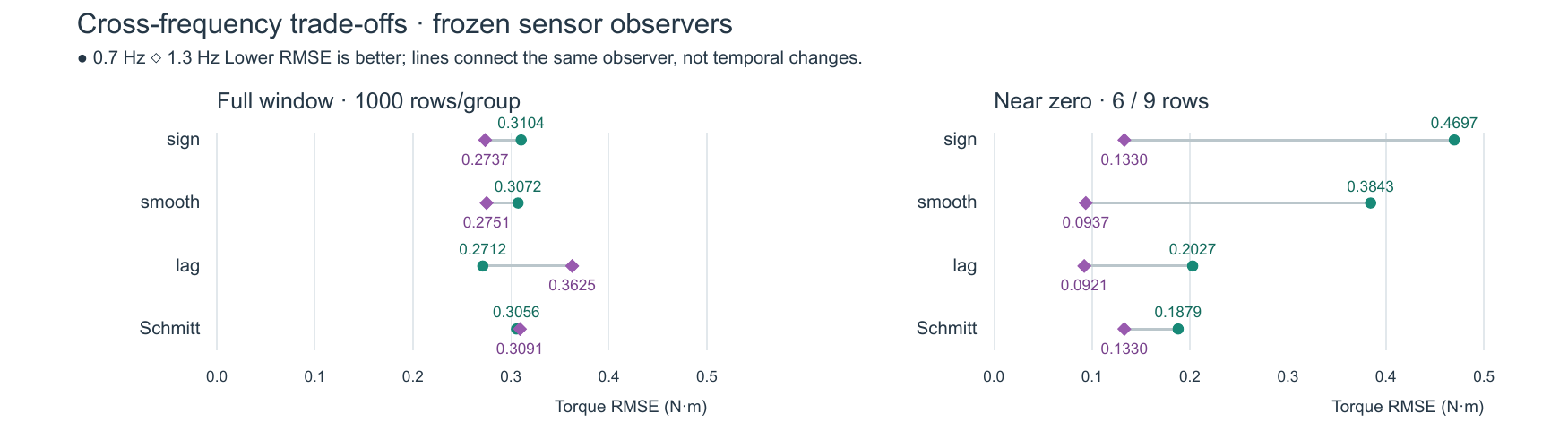}
\caption{Joint-side sensor frequency-transfer diagnostic. All sign,\allowbreak{} smooth,\allowbreak{} lag and Schmitt observer families retain full-window and near-zero metrics,\allowbreak{} including cross-frequency degradation. These are sensor-output models,\allowbreak{} not changes to robot dynamics.}
\label{fig:C5}
\phantomsection\label{html:g2-panel-8}
\end{reportfigure}
\begin{reporttable}
\centering
\renewcommand{\thetable}{C6}
\caption{Adopted simulation PD gains and complete seven-joint Fourier-window results: 10000 native rows per joint,\allowbreak{} 10 s. Mean error is prediction minus joint-side sensor. The gain vector is retained after training-only search,\allowbreak{} not a new performance improvement; the motor group does not imply equal hardware controller gains.}
\label{tab:C6}
{\fontsize{8.2}{10.1}\selectfont
\setlength{\tabcolsep}{3pt}
\renewcommand{\arraystretch}{1.13}
\begin{tabular}{>{\raggedright\arraybackslash}p{\dimexpr 0.1300\linewidth-2\tabcolsep\relax}>{\raggedright\arraybackslash}p{\dimexpr 0.0800\linewidth-2\tabcolsep\relax}>{\raggedright\arraybackslash}p{\dimexpr 0.1400\linewidth-2\tabcolsep\relax}>{\raggedright\arraybackslash}p{\dimexpr 0.1400\linewidth-2\tabcolsep\relax}>{\raggedright\arraybackslash}p{\dimexpr 0.1700\linewidth-2\tabcolsep\relax}>{\raggedright\arraybackslash}p{\dimexpr 0.1700\linewidth-2\tabcolsep\relax}>{\raggedright\arraybackslash}p{\dimexpr 0.1700\linewidth-2\tabcolsep\relax}}
\toprule
\rowcolor{tablehead}\textbf{Motor group} & \textbf{Joint} & \textbf{Kp \allowbreak{}(Nm/rad)} & \textbf{Kd \allowbreak{}(Nm s/rad)} & \textbf{Position RMSE \allowbreak{}(deg)} & \textbf{Sensor RMSE \allowbreak{}(Nm)} & \textbf{Mean error \allowbreak{}(Nm)} \\
\midrule
J1--J2 & J1 & 1000 & 1 & 0.54059 & 0.94089 & +0.54830 \\
\rowcolor{tablerow}J1--J2 & J2 & 1000 & 5 & 0.87469 & 1.22713 & +0.82437 \\
J3--J4 & J3 & 1000 & 1 & 0.20833 & 0.34289 & +0.12997 \\
\rowcolor{tablerow}J3--J4 & J4 & 1000 & 1 & 0.18353 & 0.39084 & +0.04318 \\
J5--J7 & J5 & 1000 & 1 & 0.18246 & 0.09937 & -0.08298 \\
\rowcolor{tablerow}J5--J7 & J6 & 1000 & 1 & 0.07850 & 0.28130 & +0.24271 \\
J5--J7 & J7 & 1000 & 1 & 0.05699 & 0.07619 & -0.06697 \\
\bottomrule
\end{tabular}
}
\end{reporttable}
\begin{multicols}{2}\raggedcolumns
The original gain search used 29 training windows and position-preservation guards. No tested alternative passed all guards; the original gains were retained. \hyperref[tab:C6]{Table~C6} preserves all seven-joint results. This controller belongs to these sensor diagnostics and the separately documented configurations; it is not substituted for the calibrated teleoperation servo.
\end{multicols}

\typeout{REPORTPAGE:D.1:\thepage}
\reportneedspace{285pt}
\section{Object data,\allowbreak{} model choices and identifiability}
\label{sec:D}
\phantomsection\label{html:appendix-robot-config}

\begin{reporttable}
\centering
\renewcommand{\thetable}{D1}
\caption{Separately frozen robot configurations. Object arm arrays are verified in the 14 archived native PhysX mappings for sponge and handovers; original five-case coverage is report-level.}
\label{tab:D1}
{\fontsize{8.2}{10.1}\selectfont
\setlength{\tabcolsep}{3pt}
\renewcommand{\arraystretch}{1.13}
\begin{tabular}{>{\raggedright\arraybackslash}p{\dimexpr 0.2000\linewidth-2\tabcolsep\relax}>{\raggedright\arraybackslash}p{\dimexpr 0.4000\linewidth-2\tabcolsep\relax}>{\raggedright\arraybackslash}p{\dimexpr 0.4000\linewidth-2\tabcolsep\relax}}
\toprule
\rowcolor{tablehead}\textbf{Setting} & \textbf{G2 crsB robot validation} & \textbf{G2--Swiftpicker object replay} \\
\midrule
Robot and load & G2 crsB; both grippers removed; flange frames retained & G2 with both Swiftpicker grippers; fixed within the object study \\
\rowcolor{tablerow}Dynamics / drive source & Identified arm inertials and A/F/D/b; explicit simulation PD \allowbreak{}(\hyperref[sec:C]{Appendix~\ref*{sec:C}}) & G2 simulation priors; accepted gripper settings reported below \\
Arm Kp \allowbreak{}(N\ensuremath{\cdot}m/rad) & Left: [1000,\allowbreak{} 1000,\allowbreak{} 1000,\allowbreak{} 1000,\allowbreak{} 1000,\allowbreak{} 1000,\allowbreak{} 1000] & Each arm: [13500,\allowbreak{} 13500,\allowbreak{} 7200,\allowbreak{} 7200,\allowbreak{} 7200,\allowbreak{} 1800,\allowbreak{} 1800] \\
\rowcolor{tablerow}Arm Kd \allowbreak{}(N\ensuremath{\cdot}m\ensuremath{\cdot}s/rad) & Left: [1,\allowbreak{} 5,\allowbreak{} 1,\allowbreak{} 1,\allowbreak{} 1,\allowbreak{} 1,\allowbreak{} 1] & Each arm: [360,\allowbreak{} 360,\allowbreak{} 138,\allowbreak{} 138,\allowbreak{} 138,\allowbreak{} 74,\allowbreak{} 74] \\
Passive / effort settings & Frozen identified A/F/D/b; no explicit drive saturation in the validation controller & Native armature: [0.05,\allowbreak{} 0.05,\allowbreak{} 0.05,\allowbreak{} 0.05,\allowbreak{} 0.05,\allowbreak{} 0.02,\allowbreak{} 0.02]; native friction tuple \ensuremath{\approx}[0.3,\allowbreak{} 0.3,\allowbreak{} 1]; arm effort limit 10\textsuperscript{7} N\ensuremath{\cdot}m \allowbreak{}(simulation prior) \\
\rowcolor{tablerow}Initialization / integration & Left measured q/v initialized once; other joints at XML rest; compliant base fixture; 1 ms & Recorded initial robot feedback; 22 joint + 2 gripper targets; \ensuremath{\approx}100 Hz zero-order hold; 1 ms \\
Observation & Joint-side torque sensor; wrist direction term only in predicted sensor output & Object RGB/RGB-D and gripper feedback; sponge/handover camera adjustment uses finger geometry and real-feedback FK \\
\rowcolor{tablerow}Gripper Kp / Kd & Grippers absent & 202.026337 / 13.345311 \\
Gripper effort / damping & Grippers absent & 18 N\ensuremath{\cdot}m / 1.253346 \allowbreak{}(source setting) \\
\rowcolor{tablerow}Gripper armature / friction loss & Grippers absent & 0.008019 / 0.072053 \allowbreak{}(source settings) \\
\bottomrule
\end{tabular}
}
\end{reporttable}

\begin{multicols}{2}\raggedcolumns
\phantomsection\label{html:object-configuration-evidence}
Configuration scope. The 14 mappings have identical joint names,\allowbreak{} drive gains,\allowbreak{} armatures,\allowbreak{} native friction tuples and effort limits,\allowbreak{} with parameter readback verified. This establishes fixed settings within those replays,\allowbreak{} not equality with the gripper-free configuration. Model-to-PhysX FK agreement is a numerical mapping check; the rendering camera is separate from the scoring camera. Simulation effort limits are not measured hardware ratings.

\phantomsection\label{html:object-configuration-pending}
\par\smallskip\noindent\colorbox{reviewyellow}{\parbox{\dimexpr\linewidth-2\fboxsep\relax}{\footnotesize Configuration records pending. The original gripper-calibration dataset/version,\allowbreak{} original per-case robot/model bindings \allowbreak{}(including kettle and gum box),\allowbreak{} and scoring-camera,\allowbreak{} finger-geometry and FK versions remain to be archived. Complete run/asset manifests are needed for parameter-level traceability. Fixed use does not exclude robot-model error.}}\par\smallskip
\phantomsection\label{html:appendix-search}
\end{multicols}
\begin{reporttable}
\centering
\renewcommand{\thetable}{D2}
\caption{Original object searches and same-episode windows: original recording time for kettle,\allowbreak{} elapsed time from the retained command start for other cases. The 41-candidate PhysX fit is separate from the later matched adaptation.}
\label{tab:D2}
{\fontsize{8.2}{10.1}\selectfont
\setlength{\tabcolsep}{3pt}
\renewcommand{\arraystretch}{1.13}
\begin{tabular}{>{\raggedright\arraybackslash}p{\dimexpr 0.1800\linewidth-2\tabcolsep\relax}>{\raggedright\arraybackslash}p{\dimexpr 0.1800\linewidth-2\tabcolsep\relax}>{\raggedright\arraybackslash}p{\dimexpr 0.1900\linewidth-2\tabcolsep\relax}>{\raggedright\arraybackslash}p{\dimexpr 0.2300\linewidth-2\tabcolsep\relax}>{\raggedright\arraybackslash}p{\dimexpr 0.2200\linewidth-2\tabcolsep\relax}}
\toprule
\rowcolor{tablehead}\textbf{Case / engine} & \textbf{Search budget} & \textbf{Train / later \allowbreak{}(s)} & \textbf{Adjusted quantities} & \textbf{Objective / decision} \\
\midrule
Kettle \ensuremath{\cdot} MuJoCo & 3 bounded coordinate-grid rounds & 6--19 / 19--24 & Geometry,\allowbreak{} initial yaw,\allowbreak{} COM,\allowbreak{} contact & Lid-position squared error + right-gripper feedback \\
\rowcolor{tablerow}Gum box \ensuremath{\cdot} MuJoCo & 100 candidates & 3.260--7.120 / 7.120--8.620 & Friction,\allowbreak{} contact time,\allowbreak{} impedance & \ensuremath{J_{\mathrm{3D}}} \\
Sponge \ensuremath{\cdot} MuJoCo & 120 candidates & 1.430--4.870 / 4.870--6.370 & Effective contact & \ensuremath{J_{\mathrm{3D}}} + task-feasibility gate \\
\rowcolor{tablerow}Sponge \ensuremath{\cdot} original native PhysX & 41 candidates & 1.430--4.870 / 4.870--6.370 & Stiffness,\allowbreak{} damping,\allowbreak{} friction,\allowbreak{} contact scope & \ensuremath{J_{\mathrm{3D}}} + grasp / release / target-placement gate \\
Shuttlecock \ensuremath{\cdot} MuJoCo & 81 candidates & 1--8.5 / 8.5--end & Sliding / rotational friction,\allowbreak{} contact time,\allowbreak{} impedance & \ensuremath{J_{\mathrm{RGB}}} \\
\rowcolor{tablerow}Pen \ensuremath{\cdot} MuJoCo & 81 contact + 42 radius candidates & 1--6.2 / 6.2--end & Contact parameters and radius & \ensuremath{J_{\mathrm{RGB}}}; source candidate rejected \\
\bottomrule
\end{tabular}
}
\end{reporttable}
\begin{multicols}{2}\raggedcolumns
\textbf{Configuration and search scope.} The configuration table describes the robot and load assumptions; the search table describes what was varied in the object study. Keeping these records together separates fixed execution settings from fitted effective object parameters. The following results retain the original observation units and compare each candidate within its stated replay protocol.
\end{multicols}
\typeout{REPORTPAGE:D.2:\thepage}

\begin{multicols}{2}\raggedcolumns
\phantomsection\label{html:object-search-objectives}
Objective definitions. \begin{equation}
J_{\mathrm{3D}}=\left(\frac{e_{\mathrm{3D}}}{0.01\,\mathrm m}\right)^2
+0.1\left(\frac{e_{\mathrm{hold}}}{0.03\,\mathrm{rad}}\right)^2.
\end{equation} Here \ensuremath{e_{\mathrm{3D}}} is training centre RMSE and ehold is holding-gripper RMSE. \ensuremath{J_{\mathrm{RGB}}=e_{\mathrm{RGB}}+150e_{\mathrm{dual}}},\allowbreak{} with training RGB RMSE in pixels and dual-gripper RMSE in radians; 150 is the reported numerical weight. Later-segment error is excluded from the documented \ensuremath{J_{\mathrm{3D}}} and \ensuremath{J_{\mathrm{RGB}}} objectives; development checks,\allowbreak{} including kettle inspection and full-trajectory feasibility,\allowbreak{} are not independent blind tests. The pen candidate is rejected because its small training gain accompanies worse full/later error.

\phantomsection\label{html:object-task-gates}
Reported completion checks. Kettle: lift >50 mm,\allowbreak{} airborne hand-contact samples >100,\allowbreak{} held tilt >25\ensuremath{^{\circ}},\allowbreak{} and after original t>25 s,\allowbreak{} no hand contact with table-support samples >100. Pick-and-place: lift >30 mm,\allowbreak{} hand normal force >0.1 N,\allowbreak{} unsupported samples >30,\allowbreak{} post-release support samples >30,\allowbreak{} and endpoint inside the target. Handovers: at least 20/3/20/20 samples for left hold / dual contact / right hold / released container support in preset windows,\allowbreak{} lift >25 mm,\allowbreak{} hand normal force >0.02 N,\allowbreak{} with endpoint centre and conservative horizontal envelope inside the opening. These are replay predicates,\allowbreak{} not measured real contact forces.

\phantomsection\label{html:object-search-selection}
Separate native selection. The original 41-candidate sponge search accepts 400 N/m and 1 N\ensuremath{\cdot}s/m. The later matched comparison uses 800 N/m and 5 N\ensuremath{\cdot}s/m for both models with a separate friction refit. Its archived training objective recomputes to 7.97275 \ensuremath{\rightarrow} 6.76392. Another archived 400/1 candidate scores 5.63802,\allowbreak{} so the later accepted setup is not a global minimum of that objective across these records. The two experiments retain separate results in \hyperref[tab:D9]{Table~D9}.

\phantomsection\label{html:object-search-pending}
\par\smallskip\noindent\colorbox{reviewyellow}{\parbox{\dimexpr\linewidth-2\fboxsep\relax}{\footnotesize Search records pending. Exact bounds/grids,\allowbreak{} candidate generation and per-candidate losses/gates,\allowbreak{} acceptance,\allowbreak{} tie-breaking and stopping rules,\allowbreak{} kettle weights,\allowbreak{} holding masks and dual-gripper aggregation remain to be archived. Reported completion thresholds do not specify executable gate bindings,\allowbreak{} all window/region definitions or sample cadence. Local archives contain plotting code and references to search artifacts,\allowbreak{} not the search implementations. The later native selection rationale cannot be inferred from its final settings or later-segment performance.}}\par\smallskip
\phantomsection\label{html:appendix-object-results}
\end{multicols}
\reportnotePageNineteen

\begin{reporttable}
\centering
\renewcommand{\thetable}{D3}
\caption{Object alignment before and after identification. PhysX\textsuperscript{\ensuremath{\dagger}} denotes the updated native comparisons; other rows retain source results. The source pen candidate remains unaccepted.}
\label{tab:D3}
{\fontsize{8.2}{10.1}\selectfont
\setlength{\tabcolsep}{3pt}
\renewcommand{\arraystretch}{1.13}
\begin{tabular}{>{\raggedright\arraybackslash}p{\dimexpr 0.1700\linewidth-2\tabcolsep\relax}>{\raggedright\arraybackslash}p{\dimexpr 0.1600\linewidth-2\tabcolsep\relax}>{\raggedright\arraybackslash}p{\dimexpr 0.0600\linewidth-2\tabcolsep\relax}>{\raggedright\arraybackslash}p{\dimexpr 0.1300\linewidth-2\tabcolsep\relax}>{\raggedright\arraybackslash}p{\dimexpr 0.1300\linewidth-2\tabcolsep\relax}>{\raggedright\arraybackslash}p{\dimexpr 0.1300\linewidth-2\tabcolsep\relax}>{\raggedright\arraybackslash}p{\dimexpr 0.1000\linewidth-2\tabcolsep\relax}>{\raggedright\arraybackslash}p{\dimexpr 0.1200\linewidth-2\tabcolsep\relax}}
\toprule
\rowcolor{tablehead}\textbf{Case} & \textbf{Execution stack} & \textbf{Unit} & \textbf{Full replay: before \ensuremath{\rightarrow} after} & \textbf{Training: before \ensuremath{\rightarrow} after} & \textbf{Later-time: before \ensuremath{\rightarrow} after} & \textbf{Full-replay improvement} & \textbf{Completed replays} \\
\midrule
Kettle motion & MuJoCo & mm & 39.56 \ensuremath{\rightarrow} 25.24 & 46.47 \ensuremath{\rightarrow} 29.15 & 32.65 \ensuremath{\rightarrow} 22.68 & +36.2\% & 1/1 \ensuremath{\rightarrow} 1/1 \\
\rowcolor{tablerow}Kettle motion & Newton / MJWarp & mm & 40.98 \ensuremath{\rightarrow} 25.12 & 46.98 \ensuremath{\rightarrow} 29.05 & 38.63 \ensuremath{\rightarrow} 22.68 & +38.7\% & 3/3 \ensuremath{\rightarrow} 3/3 \\
Kettle motion & PhysX & mm & 38.66 \ensuremath{\rightarrow} 26.23 & 46.54 \ensuremath{\rightarrow} 32.11 & 33.07 \ensuremath{\rightarrow} 24.01 & +32.2\% & 1/1 \ensuremath{\rightarrow} 2/2 \\
\rowcolor{tablerow}Gum-box pick and place & MuJoCo & mm & 12.78 \ensuremath{\rightarrow} 12.23 & 12.44 \ensuremath{\rightarrow} 11.75 & 15.38 \ensuremath{\rightarrow} 14.51 & +4.3\% & 1/1 \ensuremath{\rightarrow} 1/1 \\
Gum-box pick and place & Newton / MJWarp & mm & 13.37 \ensuremath{\rightarrow} 13.58 & 12.92 \ensuremath{\rightarrow} 12.85 & 16.13 \ensuremath{\rightarrow} 16.09 & -1.5\% & 3/3 \ensuremath{\rightarrow} 3/3 \\
\rowcolor{tablerow}Gum-box pick and place & PhysX & mm & 10.99 \ensuremath{\rightarrow} 11.46 & 12.98 \ensuremath{\rightarrow} 12.86 & 13.00 \ensuremath{\rightarrow} 13.41 & -4.2\% & 1/1 \ensuremath{\rightarrow} 1/1 \\
Sponge pick and place & MuJoCo & mm & 34.99 \ensuremath{\rightarrow} 20.84 & 47.85 \ensuremath{\rightarrow} 28.40 & 23.31 \ensuremath{\rightarrow} 20.81 & +40.5\% & 1/1 \ensuremath{\rightarrow} 1/1 \\
\rowcolor{tablerow}Sponge pick and place & Newton / MJWarp & mm & 59.85 \ensuremath{\rightarrow} 21.77 & 61.83 \ensuremath{\rightarrow} 28.27 & 64.90 \ensuremath{\rightarrow} 21.06 & +63.6\% & 3/3 \ensuremath{\rightarrow} 3/3 \\
Sponge pick and place & PhysX\textsuperscript{\ensuremath{\dagger}} & mm & 41.54 \ensuremath{\rightarrow} 26.73 & 27.84 \ensuremath{\rightarrow} 25.89 & 38.58 \ensuremath{\rightarrow} 26.80 & +35.64\% & 2/2 \ensuremath{\rightarrow} 4/4 \\
\rowcolor{tablerow}Shuttlecock handover & MuJoCo & px & 19.71 \ensuremath{\rightarrow} 11.15 & 18.29 \ensuremath{\rightarrow} 9.93 & 25.45 \ensuremath{\rightarrow} 14.83 & +43.4\% & 1/1 \ensuremath{\rightarrow} 1/1 \\
Shuttlecock handover & Newton / MJWarp & px & 19.81 \ensuremath{\rightarrow} 11.34 & 18.41 \ensuremath{\rightarrow} 10.11 & 25.52 \ensuremath{\rightarrow} 15.07 & +42.8\% & 3/3 \ensuremath{\rightarrow} 3/3 \\
\rowcolor{tablerow}Shuttlecock handover & PhysX\textsuperscript{\ensuremath{\dagger}} & px & 20.47 \ensuremath{\rightarrow} 21.05 & 18.88 \ensuremath{\rightarrow} 18.91 & 26.67 \ensuremath{\rightarrow} 28.37 & -2.84\% & 2/2 \ensuremath{\rightarrow} 2/2 \\
Pen handover & MuJoCo & px & 9.30 \ensuremath{\rightarrow} 9.50 & 8.59 \ensuremath{\rightarrow} 8.50 & 11.27 \ensuremath{\rightarrow} 11.83 & -2.2\% & 1/1 \ensuremath{\rightarrow} 1/1 \\
\rowcolor{tablerow}Pen handover & Newton / MJWarp & px & 9.21 \ensuremath{\rightarrow} 9.52 & 8.58 \ensuremath{\rightarrow} 8.51 & 11.08 \ensuremath{\rightarrow} 11.86 & -3.4\% & 3/3 \ensuremath{\rightarrow} 3/3 \\
Whiteboard-pen handover & PhysX\textsuperscript{\ensuremath{\dagger}} & px & 9.72 \ensuremath{\rightarrow} 9.31 & 8.50 \ensuremath{\rightarrow} 8.43 & 12.30 \ensuremath{\rightarrow} 11.47 & +4.20\% & 2/2 \ensuremath{\rightarrow} 2/2 \\
\bottomrule
\end{tabular}
}
\end{reporttable}

\begin{multicols}{2}\raggedcolumns
\textsuperscript{\ensuremath{\dagger}} Both baseline and identified models receive the same accepted PhysX settings within each case. Sponge friction is fitted separately in native PhysX; handovers compare the source parameter sets with shared angular damping. See the configuration notes below and D.1--D.3. Historical failures remain in the diagnostics and source figures; these rows do not claim unchanged-parameter equivalence across engines.
\end{multicols}
\typeout{REPORTPAGE:D.3:\thepage}
\begin{reporttable}
\centering
\renewcommand{\thetable}{D4}
\caption{Source recordings and observation coverage.}
\label{tab:D4}
{\fontsize{8.2}{10.1}\selectfont
\setlength{\tabcolsep}{3pt}
\renewcommand{\arraystretch}{1.13}
\begin{tabular}{>{\raggedright\arraybackslash}p{\dimexpr 0.1500\linewidth-2\tabcolsep\relax}>{\raggedright\arraybackslash}p{\dimexpr 0.1400\linewidth-2\tabcolsep\relax}>{\raggedright\arraybackslash}p{\dimexpr 0.1300\linewidth-2\tabcolsep\relax}>{\raggedright\arraybackslash}p{\dimexpr 0.1400\linewidth-2\tabcolsep\relax}>{\raggedright\arraybackslash}p{\dimexpr 0.2400\linewidth-2\tabcolsep\relax}>{\raggedright\arraybackslash}p{\dimexpr 0.2000\linewidth-2\tabcolsep\relax}}
\toprule
\rowcolor{tablehead}\textbf{Case} & \textbf{Real episode} & \textbf{Command samples} & \textbf{Replay duration} & \textbf{Object observations} & \textbf{Task goal} \\
\midrule
Kettle motion & 2959360 & 2621 & 26.199 s & 260 valid RGB-D lid centres & Grasp,\allowbreak{} lift,\allowbreak{} tilt,\allowbreak{} return and release \\
\rowcolor{tablerow}Gum box & 4244905 & 1270 & 12.690 s & 127 RGB-D box-fit centres & Move to yellow hexagonal tray \\
Sponge & 4247312 & 922 & 9.210 s & 92 RGB-D box-fit centres & Place in target area right of white bottle \\
\rowcolor{tablerow}Yellow shuttlecock & 4246265 & 1320 & 13.190 s & 12 RGB head points; only 3 valid depth points & Left grasp,\allowbreak{} right receive,\allowbreak{} place in red container \\
Whiteboard pen & 4257003 & 1098 & 10.970 s & 11 RGB body centres and 11 valid depth points & Left grasp,\allowbreak{} right receive,\allowbreak{} place in blue container \\
\bottomrule
\end{tabular}
}
\end{reporttable}

\reportnotePageTwenty

\begin{multicols}{2}\raggedcolumns
\paragraph{Model geometry,\allowbreak{} fixed mass priors and limitations}
\end{multicols}
\begin{reporttable}
\centering
\renewcommand{\thetable}{D5}
\caption{Model choices,\allowbreak{} fixed mass priors and limitations.}
\label{tab:D5}
{\fontsize{8.2}{10.1}\selectfont
\setlength{\tabcolsep}{3pt}
\renewcommand{\arraystretch}{1.13}
\begin{tabular}{>{\raggedright\arraybackslash}p{\dimexpr 0.1600\linewidth-2\tabcolsep\relax}>{\raggedright\arraybackslash}p{\dimexpr 0.2700\linewidth-2\tabcolsep\relax}>{\raggedright\arraybackslash}p{\dimexpr 0.2900\linewidth-2\tabcolsep\relax}>{\raggedright\arraybackslash}p{\dimexpr 0.2800\linewidth-2\tabcolsep\relax}}
\toprule
\rowcolor{tablehead}\textbf{Object} & \textbf{Asset / geometry source} & \textbf{Model and fixed mass prior} & \textbf{Main limitations} \\
\midrule
Green-lid kettle & Video-based approximate reconstruction & Cylinder body,\allowbreak{} spout,\allowbreak{} green lid,\allowbreak{} hollow five-segment handle; 0.8 kg & Approximate CAD; exact object match not established; water and kettle form an effective rigid body; liquid COM shifts omitted \\
\rowcolor{tablerow}Gum box & building\_\allowbreak{}blocks/benchmark\_\allowbreak{}building\_\allowbreak{}blocks\_\allowbreak{}090 & 50 mm blue-cube proxy; 0.1 kg & Cube proxy; exact physical-object geometry not established \\
Sponge & bowl\_\allowbreak{}sponge/benchmark\_\allowbreak{}bowl\_\allowbreak{}sponge\_\allowbreak{}001 & Scaled from static RGB-D to approximately 118.4\ensuremath{\times}82.6\ensuremath{\times}31.8 mm; 0.1 kg & Free rigid box with effective soft contact; no volumetric deformation \\
\rowcolor{tablerow}Whiteboard pen & pen/benchmark\_\allowbreak{}pen\_\allowbreak{}006 & 143.2 mm cylinder with black cap appearance; diameter approximately 15 mm; 0.1 kg & Category proxy; exact physical-object geometry not established \\
Shuttlecock & Approximate yellow-shuttlecock reconstruction & Solid head plus 16 convex skirt sectors with openings; length 75 mm,\allowbreak{} skirt diameter 62 mm; 0.005 kg & Exact object match not established; no skirt deformation or aerodynamics; equivalent-cylinder inertia prior \\
\bottomrule
\end{tabular}
}
\end{reporttable}

\begin{multicols}{2}\raggedcolumns
These representative and approximate models support an object-identification workflow across interaction families. They do not establish exact asset identity for every scored benchmark task,\allowbreak{} coverage of all benchmark assets,\allowbreak{} or an independent robot \ensuremath{\times} object ablation.

\paragraph{Identified effective contact settings and mass-sensitivity measurements}
\end{multicols}
\begin{reporttable}
\centering
\renewcommand{\thetable}{D6}
\caption{Original MuJoCo identification parameters. Native PhysX settings are given below and belong to separate experiments.}
\label{tab:D6}
{\fontsize{8.2}{10.1}\selectfont
\setlength{\tabcolsep}{3pt}
\renewcommand{\arraystretch}{1.13}
\begin{tabular}{>{\raggedright\arraybackslash}p{\dimexpr 0.1400\linewidth-2\tabcolsep\relax}>{\raggedright\arraybackslash}p{\dimexpr 0.0900\linewidth-2\tabcolsep\relax}>{\raggedright\arraybackslash}p{\dimexpr 0.1600\linewidth-2\tabcolsep\relax}>{\raggedright\arraybackslash}p{\dimexpr 0.2000\linewidth-2\tabcolsep\relax}>{\raggedright\arraybackslash}p{\dimexpr 0.1600\linewidth-2\tabcolsep\relax}>{\raggedright\arraybackslash}p{\dimexpr 0.2500\linewidth-2\tabcolsep\relax}}
\toprule
\rowcolor{tablehead}\textbf{Case} & \textbf{Fixed mass \allowbreak{}(kg)} & \textbf{Sliding friction: before \ensuremath{\rightarrow} after} & \textbf{MuJoCo contact time \allowbreak{}(s): before \ensuremath{\rightarrow} after} & \textbf{Impedance: before \ensuremath{\rightarrow} after} & \textbf{Other changes} \\
\midrule
Kettle motion & 0.8 & 0.6\ensuremath{\rightarrow}1.2 & 0.004\ensuremath{\rightarrow}0.008 & Model default retained & Geometry,\allowbreak{} COM and initial yaw jointly adjusted \\
\rowcolor{tablerow}Gum box & 0.1 & 0.5\ensuremath{\rightarrow}0.8380273 & 0.004\ensuremath{\rightarrow}0.0039090 & 0.95\ensuremath{\rightarrow}0.98 & Geometry / inertia strategy fixed \\
Sponge & 0.1 & 3.0\ensuremath{\rightarrow}1.5380 & 0.004\ensuremath{\rightarrow}0.018283 & 0.5\ensuremath{\rightarrow}0.95 & Effective soft finger-pad contact; stiffer table support \\
\rowcolor{tablerow}Shuttlecock & 0.005 & 0.5\ensuremath{\rightarrow}0.499165 & 0.004\ensuremath{\rightarrow}0.004626972 & 0.95\ensuremath{\rightarrow}0.9387749 & Torsional friction 0.005\ensuremath{\rightarrow}0.0003694332 m; rolling 0.003\ensuremath{\rightarrow}0.0003632741 m \\
Pen candidate & 0.1 & 0.5\ensuremath{\rightarrow}0.6440305 & 0.004\ensuremath{\rightarrow}0.002277796 & 0.95\ensuremath{\rightarrow}0.9004645 & Radius 7.5\ensuremath{\rightarrow}7.4 mm; torsional 0.005\ensuremath{\rightarrow}0.004163496 m; rolling 0.003\ensuremath{\rightarrow}0.001974794 m \\
\bottomrule
\end{tabular}
}
\end{reporttable}
\typeout{REPORTPAGE:D.4:\thepage}

\begin{multicols}{2}\raggedcolumns
\textbf{Updated native PhysX configuration.} Both baseline and identified models share the accepted engine settings and initial state within each case. Native integration uses CPU TGS,\allowbreak{} 32/1 solver iterations and a 1 ms step; recorded real joint and gripper position commands,\allowbreak{} robot settings,\allowbreak{} camera calibration and scoring times are retained.

\textbf{Sponge:} right-finger contact stiffness 800 N/m and damping 5 N\ensuremath{\cdot}s/m for both models; native object sliding friction 3.0 \ensuremath{\rightarrow} 1.2. \textbf{Shuttlecock:} target angular damping 20 s\textsuperscript{-}\textsuperscript{1} for both models; sliding friction 0.500000 \ensuremath{\rightarrow} 0.499165,\allowbreak{} with fixed geometry and a 5 g mass prior. \textbf{Pen:} target angular damping 500 s\textsuperscript{-}\textsuperscript{1} and the same frozen baseline initial state; sliding friction 0.500000 \ensuremath{\rightarrow} 0.644031,\allowbreak{} radius 7.5 \ensuremath{\rightarrow} 7.4 mm with associated source inertia,\allowbreak{} and a fixed 100 g mass prior. Unmapped MuJoCo contact terms are not counted as applied native differences.

\textbf{Evaluation windows and repetition.} For the updated native comparisons,\allowbreak{} full / training / later-time primary-observation counts are 92 / 34 / 15 for sponge,\allowbreak{} 12 / 8 / 3 for shuttlecock,\allowbreak{} and 11 / 6 / 4 for pen. Relative to replay start,\allowbreak{} sponge training is [1.430,\allowbreak{} 4.870) s and later-time is [4.870,\allowbreak{} 6.370) s \allowbreak{}(boundaries rounded to 1 ms); shuttlecock training is [1,\allowbreak{} 8.5) s and later-time is \ensuremath{\geq} 8.5 s; pen training is [1,\allowbreak{} 6.2) s and later-time is \ensuremath{\geq} 6.2 s,\allowbreak{} within each replay. Full errors include additional valid early or released observations and are not averages of the two subset errors. These temporal subsets were inspected during development and are not blind independent tests. Recorded-state repeat checks find identical trajectories within each handover condition; repetition counts therefore do not support confidence intervals or population success rates.
\end{multicols}
\begin{reporttable}
\centering
\renewcommand{\thetable}{D7}
\caption{Mass sensitivity with fixed contact settings.}
\label{tab:D7}
{\fontsize{8.2}{10.1}\selectfont
\setlength{\tabcolsep}{3pt}
\renewcommand{\arraystretch}{1.13}
\begin{tabular}{>{\raggedright\arraybackslash}p{\dimexpr 0.2300\linewidth-2\tabcolsep\relax}>{\raggedright\arraybackslash}p{\dimexpr 0.1900\linewidth-2\tabcolsep\relax}>{\raggedright\arraybackslash}p{\dimexpr 0.3000\linewidth-2\tabcolsep\relax}>{\raggedright\arraybackslash}p{\dimexpr 0.2800\linewidth-2\tabcolsep\relax}}
\toprule
\rowcolor{tablehead}\textbf{Mass varied with accepted contact settings fixed} & \textbf{Mass \allowbreak{}(kg)} & \textbf{Training 3D RMSE \allowbreak{}(mm)} & \textbf{Interpretation} \\
\midrule
Kettle motion & 0.2 / 0.4 / 0.8 / 1.2 / 1.6 & 30.07 / 29.21 / 29.15 / 29.23 / 29.43 & Similar errors over 0.4--1.6 kg; insufficient mass excitation \\
\rowcolor{tablerow}Gum box & 0.01 / 0.03 / 0.1 / 0.3 & 14.2 / 11.6 / 11.8 / 12.8 & All complete; 0.1 kg is not uniquely identified \\
Sponge & 0.01 / 0.03 / 0.1 / 0.3 & 421.9 / 396.2 / 28.4 / 30.3 & First two settings fail; parameter coupling does not establish unique mass identification \\
\bottomrule
\end{tabular}
}
\end{reporttable}

\begin{multicols}{2}\raggedcolumns
Contact time and impedance are MuJoCo settings,\allowbreak{} not independent stiffness measurements. Rotational-friction terms follow MuJoCo's length-unit convention. Pen settings are a rejected candidate.

\subsection{Handover diagnostics and task stages}
\label{sec:D.1}

\phantomsection\label{html:dual-arm}

Shuttlecock and pen left-to-right replays expose holding,\allowbreak{} simultaneous contact and receiver-grasp stages relevant to the bimanual benchmark family. These are representative handover-interaction probes,\allowbreak{} evaluated separately from robot dynamics identification. MuJoCo/MJWarp complete the recorded sequences. Original unadapted PhysX objects roll away before grasp; each task has historical completion counts of 0/1 \ensuremath{\rightarrow} 0/2. The updated native pairs complete both tasks under shared angular damping. This separates contact-stage recovery from changes in observation error.
\end{multicols}
\begin{reporttable}
\centering
\renewcommand{\thetable}{D8}
\caption{Handover completion counts; baseline \ensuremath{\rightarrow} identified / candidate. PhysX\textsuperscript{\ensuremath{\dagger}} uses shared adapted settings; source failures are retained above.}
\label{tab:D8}
{\fontsize{8.2}{10.1}\selectfont
\setlength{\tabcolsep}{3pt}
\renewcommand{\arraystretch}{1.13}
\begin{tabular}{>{\raggedright\arraybackslash}p{\dimexpr 0.1700\linewidth-2\tabcolsep\relax}>{\raggedright\arraybackslash}p{\dimexpr 0.2300\linewidth-2\tabcolsep\relax}>{\raggedright\arraybackslash}p{\dimexpr 0.2000\linewidth-2\tabcolsep\relax}>{\raggedright\arraybackslash}p{\dimexpr 0.2000\linewidth-2\tabcolsep\relax}>{\raggedright\arraybackslash}p{\dimexpr 0.2000\linewidth-2\tabcolsep\relax}}
\toprule
\rowcolor{tablehead}\textbf{Case} & \textbf{Observations} & \textbf{MuJoCo} & \textbf{MJWarp} & \textbf{PhysX\textsuperscript{\ensuremath{\dagger}}} \\
\midrule
Shuttlecock & 12 RGB / 3 depth points & 1/1 \ensuremath{\rightarrow} 1/1 & 3/3 \ensuremath{\rightarrow} 3/3 & 2/2 \ensuremath{\rightarrow} 2/2 \\
\rowcolor{tablerow}Whiteboard pen & 11 RGB / 11 depth points & 1/1 \ensuremath{\rightarrow} 1/1 & 3/3 \ensuremath{\rightarrow} 3/3 & 2/2 \ensuremath{\rightarrow} 2/2 \\
\bottomrule
\end{tabular}
}
\end{reporttable}
\begin{multicols}{2}\raggedcolumns
\subsection{A separate rotational-resistance diagnostic}
\label{sec:D.2}

The rotational-resistance ablation reproduces pre-grasp rolling in MuJoCo,\allowbreak{} supporting a contact-representation explanation for the shuttlecock and pen failures. It does not rule out other model and solver differences. The identification study did not flatten the table to hide the failure or claim an independent PhysX identification for these two objects.

\textbf{Updated native rotational damping.} Applying target angular damping to both object parameter sets prevents pre-grasp roll-away without changing the recorded commands,\allowbreak{} task geometry or success predicates. Historical full RGB errors were 97.38 \ensuremath{\rightarrow} 97.36 px for the shuttlecock and 203.52 \ensuremath{\rightarrow} 251.12 px for the pen; these failures remain archived. With shared 20 s\textsuperscript{-}\textsuperscript{1} damping,\allowbreak{} both shuttlecock models complete the handover and container placement,\allowbreak{} but full RGB error rises 20.47 \ensuremath{\rightarrow} 21.05 px and later-time error rises 26.67 \ensuremath{\rightarrow} 28.37 px. There is no matched-setting identification gain; the baseline remains recommended.

\end{multicols}
\typeout{REPORTPAGE:D.5:\thepage}
\begin{multicols}{2}\raggedcolumns
With shared 500 s\textsuperscript{-}\textsuperscript{1} damping,\allowbreak{} both pen models also complete the task. Full RGB error falls 9.72 \ensuremath{\rightarrow} 9.31 px,\allowbreak{} only 0.408 px; this is below approximately 3 px annotation uncertainty. Auxiliary full 3D error is nearly unchanged at 19.39 \ensuremath{\rightarrow} 19.35 mm,\allowbreak{} while training 3D error worsens 17.15 \ensuremath{\rightarrow} 18.01 mm. Strong damping is an effective engine adaptation rather than measured rolling friction. A fixed 100 g mass prior,\allowbreak{} unvalidated contact forces and approximately 0.8--0.9 mm final conservative clearance limit physical conclusions.

Each updated handover pair completes 2/2 \ensuremath{\rightarrow} 2/2 replays of one real episode. Later-time segments were inspected during recovery; they are not blind independent tests. MuJoCo/MJWarp values are archived source results,\allowbreak{} not newly integrated runs. Task recovery and identification gains are reported separately.
\end{multicols}

\begin{multicols}{2}\raggedcolumns
\subsection{Historical transfer failure and native PhysX recovery}
\label{sec:D.3}

\phantomsection\label{html:sponge-diagnosis-full}

\textbf{Historical question and observation.} Can an effective contact model fitted in MuJoCo preserve a sponge grasp after fixed transfer? The G2 source recording contains 92 RGB-D centre observations over 9.210 s,\allowbreak{} from one real episode \allowbreak{}(4247312). After a 120-candidate MuJoCo search,\allowbreak{} full-replay centre RMSE falls from 34.99 to 20.84 mm in MuJoCo and 59.85 to 21.77 mm in MJWarp. PhysX instead rises from 25.50 to \textbf{524.61 mm}; completed replays change from 1/1 to 0/1,\allowbreak{} with failure to lift. These are replay counts,\allowbreak{} not population success rates.

\textbf{Working hypothesis.} The fitted effective compliant contact is not represented equivalently in the PhysX transfer. The source path does not express MuJoCo solref/solimp equivalently. This supports investigating contact representation and parameter mapping; it does not uniquely identify the cause. Geometry,\allowbreak{} initialization,\allowbreak{} robot/observation calibration and numerical differences remain potential contributors.

\textbf{Earlier correction attempt --- a separate experiment.} A native PhysX round searches 41 candidates with effective compliant finger-pad contact,\allowbreak{} while holding the accepted robot settings fixed. Its baseline already completes 3/3 replays and its fitted model completes 4/4. Therefore this round cannot be described as rescuing the failed fixed-transfer baseline.
\end{multicols}
\begin{reportfigure}
\centering
\renewcommand{\thefigure}{D1}
\includegraphics[width=\linewidth,height=.53\textheight,keepaspectratio]{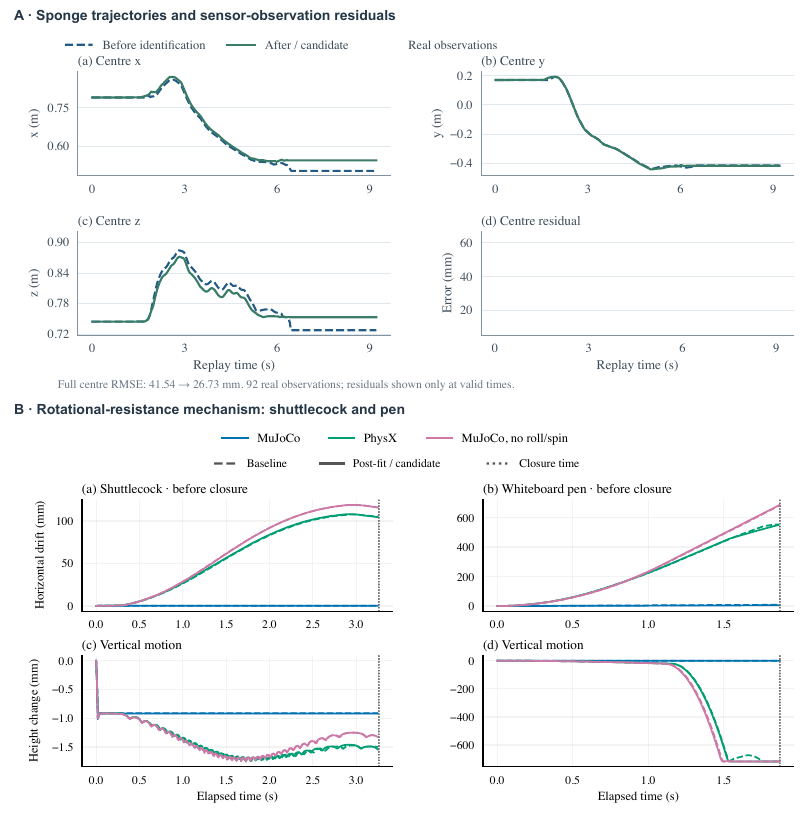}
\caption{Sponge motion under shared native contact settings. Complete simulated x/y/z centre trajectories are curves; 92 valid real observations and Euclidean centre residuals retain their sampled times. Blue dashed / crosses: baseline; teal solid / circles: fitted model; charcoal: real observations. Units are metres for coordinates and millimetres for residuals. No dense real trajectory or independent validation interval is inferred.  Historical experiment. Removing MuJoCo rotational resistance reproduces pre-grasp rolling for the shuttlecock and pen; this supports a mechanism without excluding other causes.}
\label{fig:D1}
\phantomsection\label{html:physx-sponge-comparison}
\end{reportfigure}
\typeout{REPORTPAGE:D.6:\thepage}
\begin{reporttable}
\centering
\renewcommand{\thetable}{D9}
\caption{Native PhysX sponge comparisons,\allowbreak{} separated by experiment. Matched 800/5 denotes shared contact stiffness 800 N/m and damping 5 N\ensuremath{\cdot}s/m.}
\label{tab:D9}
{\fontsize{8.2}{10.1}\selectfont
\setlength{\tabcolsep}{3pt}
\renewcommand{\arraystretch}{1.13}
\begin{tabular}{>{\raggedright\arraybackslash}p{\dimexpr 0.5800\linewidth-2\tabcolsep\relax}>{\raggedright\arraybackslash}p{\dimexpr 0.1500\linewidth-2\tabcolsep\relax}>{\raggedright\arraybackslash}p{\dimexpr 0.1500\linewidth-2\tabcolsep\relax}>{\raggedright\arraybackslash}p{\dimexpr 0.1200\linewidth-2\tabcolsep\relax}}
\toprule
\rowcolor{tablehead}\textbf{Experiment and metric} & \textbf{Before} & \textbf{After} & \textbf{Unit} \\
\midrule
Earlier native fit: full centre RMSE & 25.5031 & 24.8778 & mm \\
\rowcolor{tablerow}Earlier native fit: training centre RMSE & 26.3395 & 23.7108 & mm \\
Earlier native fit: later-time centre RMSE & 27.2964 & 29.5705 & mm \\
\rowcolor{tablerow}Earlier native fit: holding-gripper RMSE & 0.8221 & 0.0120 & rad \\
Earlier native fit: completed replays & 3/3 & 4/4 & count \\
\rowcolor{tablerow}Matched 800/5: full centre RMSE & 41.54 & 26.73 & mm \\
Matched 800/5: training centre RMSE & 27.84 & 25.89 & mm \\
\rowcolor{tablerow}Matched 800/5: later-time centre RMSE & 38.58 & 26.80 & mm \\
Matched 800/5: full-replay improvement & --- & +35.64 & \% \\
\rowcolor{tablerow}Matched 800/5: completed replays & 2/2 & 4/4 & count \\
\bottomrule
\end{tabular}
}
\end{reporttable}

\begin{multicols}{2}\raggedcolumns
\textbf{What the earlier correction establishes.} Holding-gripper RMSE improves from 0.8221 to 0.0120 rad,\allowbreak{} while full-replay centre error changes only from 25.5031 to 24.8778 mm. Later-time error worsens from 27.2964 to 29.5705 mm. Reported finger-pad penetration changes from about 0.10 to 35.92 mm. The model can approximate compression through effective contact; it does not validate real volumetric deformation,\allowbreak{} generalization or measured material stiffness.

\textbf{Updated matched-contact comparison.} Both models complete grasp,\allowbreak{} lift,\allowbreak{} transport and release onto the specified support. Under the same 800 N/m stiffness and 5 N\ensuremath{\cdot}s/m damping,\allowbreak{} native friction refitting from 3.0 to 1.2 reduces full centre RMSE by 35.64\%; training and later-time errors also decrease \allowbreak{}(\hyperref[tab:D9]{Table~D9}). The full-trajectory gain is driven mainly by horizontal x alignment \allowbreak{}(38.31 \ensuremath{\rightarrow} 16.79 mm axis RMSE); vertical z error worsens from 13.40 to 19.38 mm,\allowbreak{} so lift-height agreement does not improve. Maximum effective hand penetration remains 35.82 \ensuremath{\rightarrow} 32.79 mm,\allowbreak{} so successful rigid-contact motion does not validate volumetric deformation or measured material stiffness. Common engine recovery is not included in the identification improvement. The earlier baseline belongs to a different contact configuration and is not substituted into this matched pair.
\end{multicols}

\begin{reportfigure}
\centering
\renewcommand{\thefigure}{D2}
\includegraphics[width=.95\linewidth,height=.69\textheight,keepaspectratio]{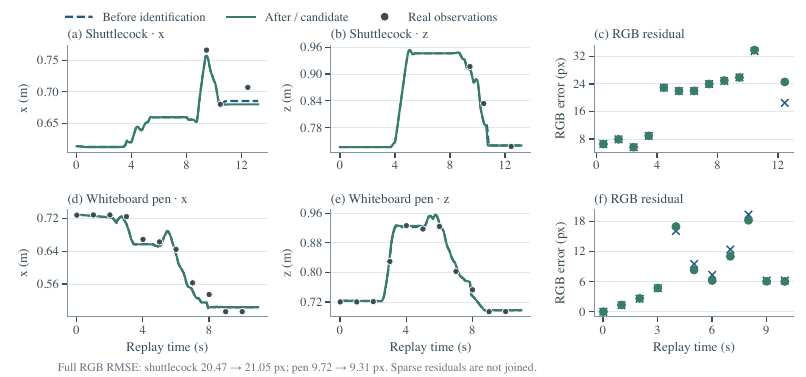}
\caption{Handover recovery and alignment diagnostics: shuttlecock \allowbreak{}(top) and whiteboard pen \allowbreak{}(bottom) under shared native angular damping. Task completion is recovered in both conditions; the shuttlecock error increases and the pen gain is below annotation uncertainty. Complete tracked-feature horizontal x and vertical z trajectories are curves; valid real depth points remain scatter. The right column shows Euclidean RGB residuals only at valid annotation times: 12 for the shuttlecock and 11 for the pen. Full x/y/z data are retained in the editable snapshots. Blue dashed / crosses: baseline; teal solid / circles: after / candidate; charcoal: real observations. The shuttlecock has only three valid depth points. Sparse residuals are not joined.}
\label{fig:D2}
\phantomsection\label{html:physx-handover-comparison}
\end{reportfigure}
\typeout{REPORTPAGE:D.7:\thepage}

\begin{multicols}{2}\raggedcolumns
\phantomsection\label{html:object-study-supplement}
\subsection{Original identification and fixed-parameter transfer}
\label{sec:D.4}

\phantomsection\label{html:original-identification}

Reading \hyperref[fig:D3]{Figures~D3} and D5--D7. Each cell retains before \ensuremath{\rightarrow} after means,\allowbreak{} signed error reduction and repeat ranges; logarithmic RMSE axes are shared across engines within each object row,\allowbreak{} with units shown per row. Crosses mark incomplete tasks; pen after-results use an unaccepted candidate. Repeat ranges are all-repeat min--max,\allowbreak{} not confidence intervals. Reduction = 100 \ensuremath{\times} \allowbreak{}(1 \ensuremath{-} after/before); millimetre and pixel errors are not pooled.
\end{multicols}
\reportnotePageTwentyFour

\begin{reportfigure}
\centering
\renewcommand{\thefigure}{D3}
\includegraphics[width=.94\linewidth,height=.69\textheight,keepaspectratio]{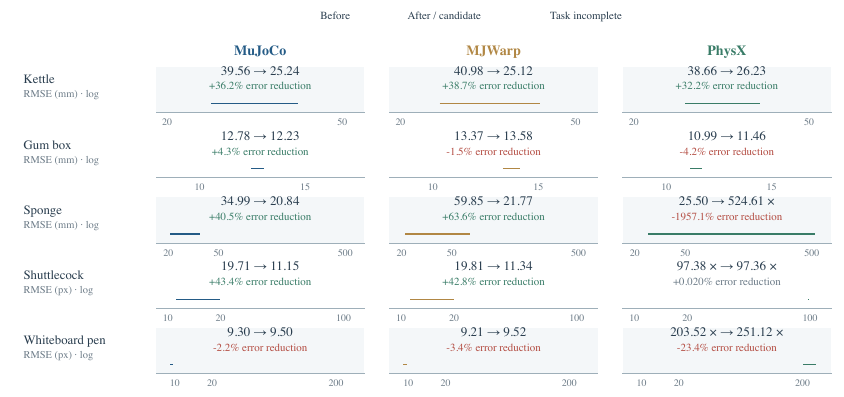}
\caption{Original identification: object-observation RMSE across three stacks,\allowbreak{} including repeat ranges and incomplete replays.}
\label{fig:D3}
\phantomsection\label{html:raw001-figure-1}
\end{reportfigure}
\begin{reportfigure}
\centering
\renewcommand{\thefigure}{D4}
{\fontsize{8.2}{10.1}\selectfont
\setlength{\tabcolsep}{3pt}
\renewcommand{\arraystretch}{1.13}
\begin{tabular}{>{\raggedright\arraybackslash}p{\dimexpr 0.2500\linewidth-2\tabcolsep\relax}>{\raggedright\arraybackslash}p{\dimexpr 0.2500\linewidth-2\tabcolsep\relax}>{\raggedright\arraybackslash}p{\dimexpr 0.2500\linewidth-2\tabcolsep\relax}>{\raggedright\arraybackslash}p{\dimexpr 0.2500\linewidth-2\tabcolsep\relax}}
\toprule
\rowcolor{tablehead}\textbf{Case} & \textbf{MuJoCo: before \ensuremath{\rightarrow} after} & \textbf{MJWarp: before \ensuremath{\rightarrow} after} & \textbf{PhysX: before \ensuremath{\rightarrow} after} \\
\midrule
Kettle & 1/1 \ensuremath{\rightarrow} 1/1 & 3/3 \ensuremath{\rightarrow} 3/3 & 1/1 \ensuremath{\rightarrow} 2/2 \\
\rowcolor{tablerow}Gum box & 1/1 \ensuremath{\rightarrow} 1/1 & 3/3 \ensuremath{\rightarrow} 3/3 & 1/1 \ensuremath{\rightarrow} 1/1 \\
Sponge & 1/1 \ensuremath{\rightarrow} 1/1 & 3/3 \ensuremath{\rightarrow} 3/3 & 1/1 \ensuremath{\rightarrow} 0/1 \\
\rowcolor{tablerow}Shuttlecock & 1/1 \ensuremath{\rightarrow} 1/1 & 3/3 \ensuremath{\rightarrow} 3/3 & 0/1 \ensuremath{\rightarrow} 0/2 \\
Pen candidate \allowbreak{}(not accepted) & 1/1 \ensuremath{\rightarrow} 1/1 & 3/3 \ensuremath{\rightarrow} 3/3 & 0/1 \ensuremath{\rightarrow} 0/2 \\
\bottomrule
\end{tabular}
}

\caption{Original fixed-transfer completion counts,\allowbreak{} before \ensuremath{\rightarrow} after. All 30 source fractions are retained; the pen uses an unaccepted candidate. Command repeats are not independent real trials; the separate PhysX refit is excluded.}
\label{fig:D4}
\phantomsection\label{html:raw001-figure-2}
\end{reportfigure}
\begin{reportfigure}
\centering
\renewcommand{\thefigure}{D5}
\includegraphics[width=.94\linewidth,height=.69\textheight,keepaspectratio]{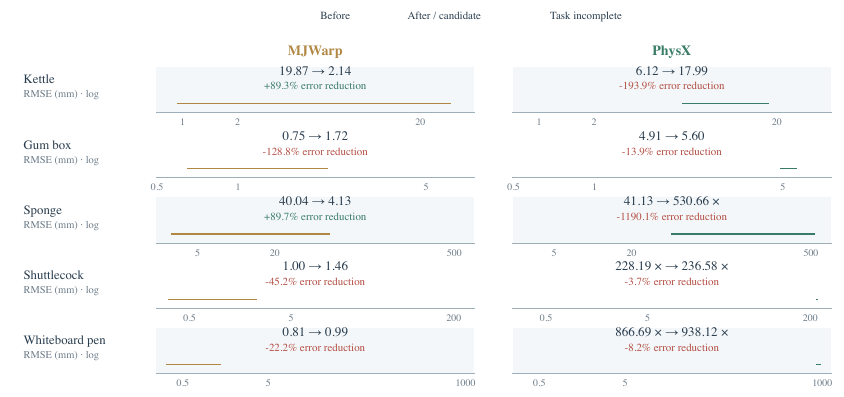}
\caption{Same-parameter trajectory differences relative to MuJoCo: cross-stack agreement,\allowbreak{} rather than real-observation accuracy.}
\label{fig:D5}
\phantomsection\label{html:raw001-figure-3}
\end{reportfigure}
\typeout{REPORTPAGE:D.8:\thepage}

\subsection{Later-time evaluation and gripper response}

\label{sec:D.5}
\phantomsection\label{html:later-time-evaluation}

\begin{reportfigure}
\centering
\renewcommand{\thefigure}{D6}
\includegraphics[width=\linewidth,height=.69\textheight,keepaspectratio]{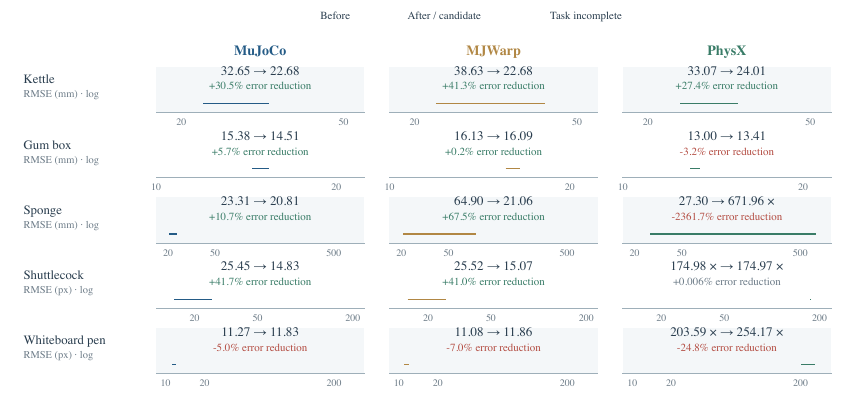}
\caption{Later-time observation errors from the same recorded episodes; these temporal checks are not independent-episode tests.}
\label{fig:D6}
\phantomsection\label{html:raw001-figure-4}
\end{reportfigure}
\begin{reportfigure}
\centering
\renewcommand{\thefigure}{D7}
\includegraphics[width=\linewidth,height=.69\textheight,keepaspectratio]{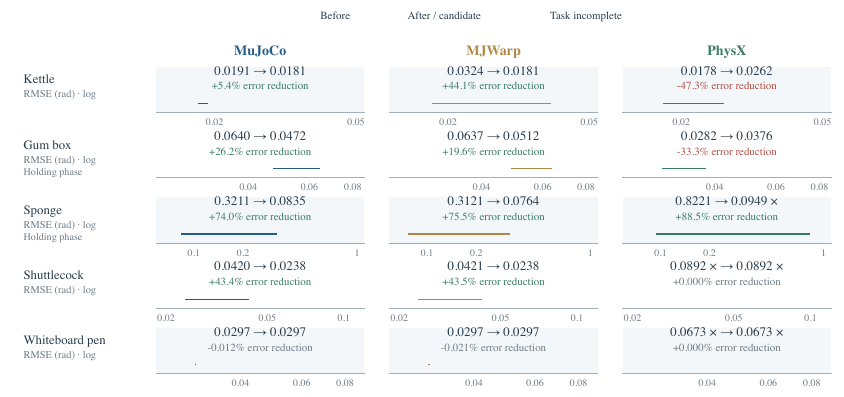}
\caption{Gripper-feedback errors across the original replays. Gum box and sponge use the holding phase; other cases use full replay. A missed grasp may still produce a small angle error.}
\label{fig:D7}
\phantomsection\label{html:raw001-figure-5}
\end{reportfigure}
\begin{multicols}{2}\raggedcolumns
\textbf{Two complementary error views.} The upper matrix reports object-observation error later in each recorded episode; the lower matrix reports gripper feedback over its stated phase. Their units, time windows and denominators differ. A small gripper-angle error can occur even when object contact fails, so it is read alongside the retained completion counts and object trajectories rather than as a substitute for them. These panels also preserve cases where the candidate increases error.
\end{multicols}
\reportnotePageTwentySix

\par\noindent\begin{minipage}{\textwidth}
\setlength{\reportfigurebefore}{2pt}\setlength{\reportfigureafter}{6pt}
\begin{minipage}[t]{.34\linewidth}
\subsection{Separate native PhysX re-identification}

\label{sec:D.6}
\phantomsection\label{html:separate-physx-refit}

\begin{reportfigure}
\centering
\renewcommand{\thefigure}{D8}
\includegraphics[width=\linewidth,height=.43\textheight,keepaspectratio]{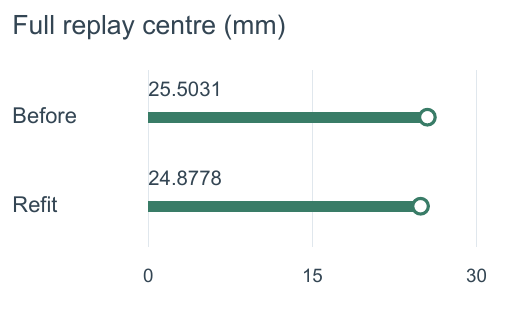}\par\vspace{2pt}
\includegraphics[width=\linewidth,height=.43\textheight,keepaspectratio]{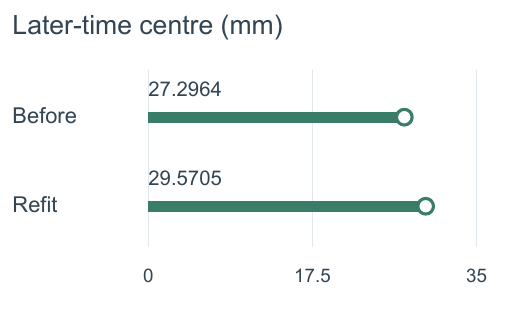}\par\vspace{2pt}
\includegraphics[width=\linewidth,height=.43\textheight,keepaspectratio]{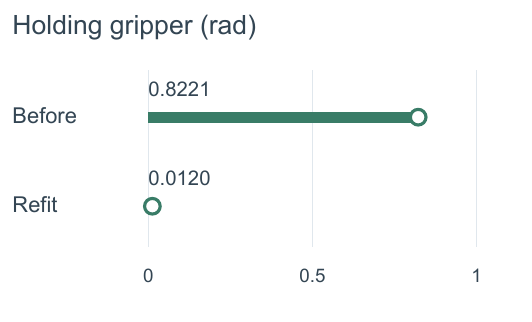}
\caption{Separate native PhysX sponge refit: full/later-time centre RMSE and holding-gripper RMSE. Completed replays: baseline 3/3,\allowbreak{} refit 4/4; linear axes start at zero. Later-time error increases. The training-centre metric remains in the supplementary table and was not an original panel.}
\label{fig:D8}
\phantomsection\label{html:raw001-figure-6}
\end{reportfigure}

\end{minipage}\hfill
\begin{minipage}[t]{.63\linewidth}
\subsection{Recorded observations and trajectories}

\label{sec:D.7}
\phantomsection\label{html:recorded-object-trajectories}

\newsavebox{\trajectorynatural}
\sbox{\trajectorynatural}{\includegraphics{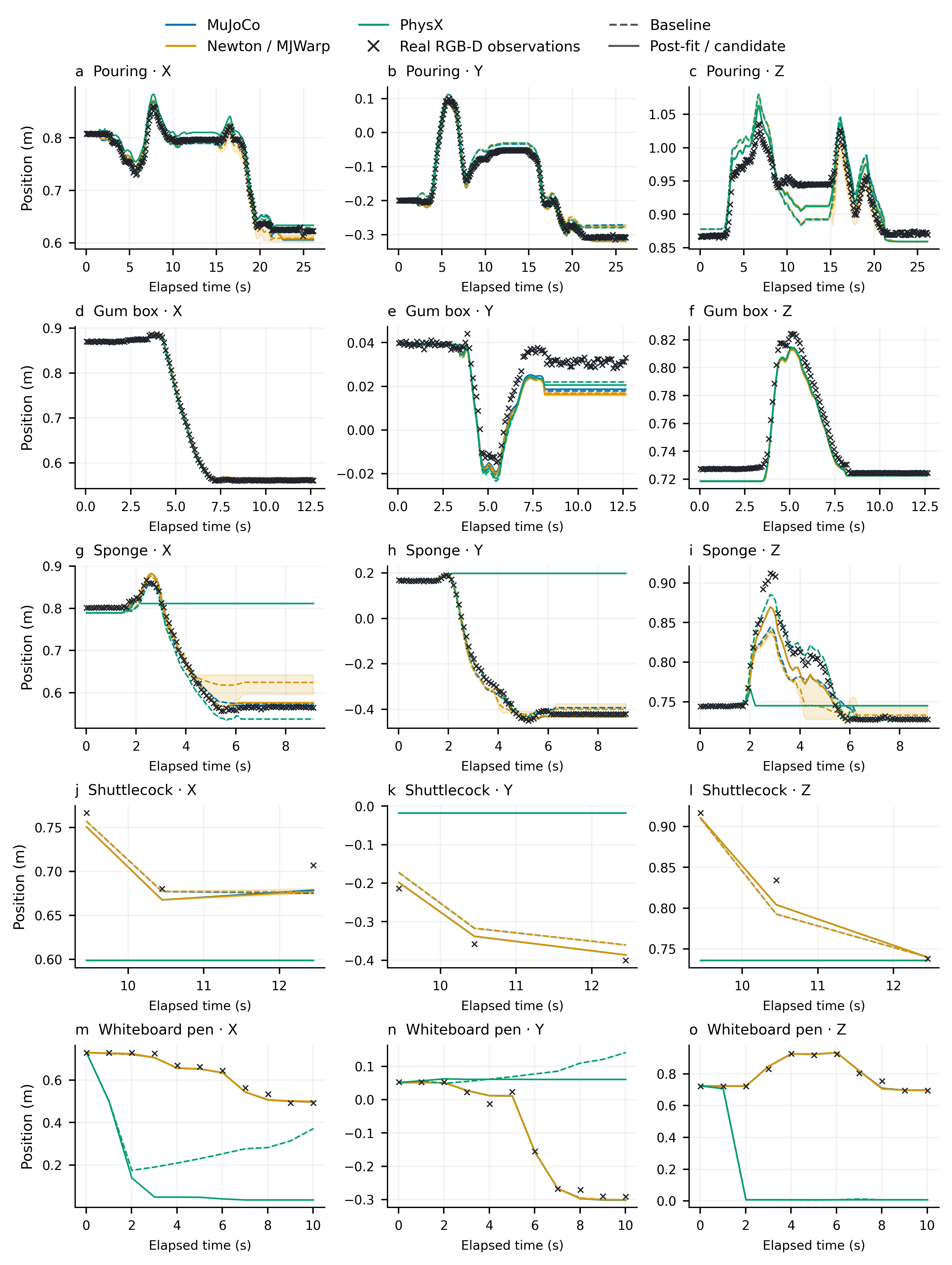}}
\newlength{\trajectorybottom}
\newlength{\trajectorytop}
\noindent\begin{minipage}{\linewidth}\centering\offinterlineskip
\setlength{\trajectorybottom}{0.9740092640\ht\trajectorynatural}
\setlength{\trajectorytop}{0.0149253731\ht\trajectorynatural}
\hbox{\includegraphics[trim={0pt \trajectorybottom{} 0pt \trajectorytop},clip,width=\linewidth]{figures/figure-07.png}}
\vspace{.35pt}
\setlength{\trajectorybottom}{0.9572825528\ht\trajectorynatural}
\setlength{\trajectorytop}{0.0313947504\ht\trajectorynatural}
\hbox{\includegraphics[trim={0pt \trajectorybottom{} 0pt \trajectorytop},clip,width=\linewidth]{figures/figure-07.png}}
\vspace{.35pt}
\setlength{\trajectorybottom}{0.9374678332\ht\trajectorynatural}
\setlength{\trajectorytop}{0.0514668039\ht\trajectorynatural}
\hbox{\includegraphics[trim={0pt \trajectorybottom{} 0pt \trajectorytop},clip,width=\linewidth]{figures/figure-07.png}}
\vspace{.35pt}
\setlength{\trajectorybottom}{0.7982501287\ht\trajectorynatural}
\setlength{\trajectorytop}{0.0681935152\ht\trajectorynatural}
\hbox{\includegraphics[trim={0pt \trajectorybottom{} 0pt \trajectorytop},clip,width=\linewidth]{figures/figure-07.png}}
\vspace{.35pt}
\setlength{\trajectorybottom}{0.7848687597\ht\trajectorynatural}
\setlength{\trajectorytop}{0.2068965517\ht\trajectorynatural}
\hbox{\includegraphics[trim={0pt \trajectorybottom{} 0pt \trajectorytop},clip,width=\linewidth]{figures/figure-07.png}}
\vspace{.35pt}
\setlength{\trajectorybottom}{0.7676273803\ht\trajectorynatural}
\setlength{\trajectorytop}{0.2225939269\ht\trajectorynatural}
\hbox{\includegraphics[trim={0pt \trajectorybottom{} 0pt \trajectorytop},clip,width=\linewidth]{figures/figure-07.png}}
\vspace{.35pt}
\setlength{\trajectorybottom}{0.7498713330\ht\trajectorynatural}
\setlength{\trajectorytop}{0.2411219763\ht\trajectorynatural}
\hbox{\includegraphics[trim={0pt \trajectorybottom{} 0pt \trajectorytop},clip,width=\linewidth]{figures/figure-07.png}}
\vspace{.35pt}
\setlength{\trajectorybottom}{0.6088522903\ht\trajectorynatural}
\setlength{\trajectorytop}{0.2555326814\ht\trajectorynatural}
\hbox{\includegraphics[trim={0pt \trajectorybottom{} 0pt \trajectorytop},clip,width=\linewidth]{figures/figure-07.png}}
\vspace{.35pt}
\setlength{\trajectorybottom}{0.5954709213\ht\trajectorynatural}
\setlength{\trajectorytop}{0.3962943901\ht\trajectorynatural}
\hbox{\includegraphics[trim={0pt \trajectorybottom{} 0pt \trajectorytop},clip,width=\linewidth]{figures/figure-07.png}}
\vspace{.35pt}
\setlength{\trajectorybottom}{0.5782295419\ht\trajectorynatural}
\setlength{\trajectorytop}{0.4119917653\ht\trajectorynatural}
\hbox{\includegraphics[trim={0pt \trajectorybottom{} 0pt \trajectorytop},clip,width=\linewidth]{figures/figure-07.png}}
\vspace{.35pt}
\setlength{\trajectorybottom}{0.5586721565\ht\trajectorynatural}
\setlength{\trajectorytop}{0.4302624807\ht\trajectorynatural}
\hbox{\includegraphics[trim={0pt \trajectorybottom{} 0pt \trajectorytop},clip,width=\linewidth]{figures/figure-07.png}}
\vspace{.35pt}
\setlength{\trajectorybottom}{0.4194544519\ht\trajectorynatural}
\setlength{\trajectorytop}{0.4436438497\ht\trajectorynatural}
\hbox{\includegraphics[trim={0pt \trajectorybottom{} 0pt \trajectorytop},clip,width=\linewidth]{figures/figure-07.png}}
\vspace{.35pt}
\setlength{\trajectorybottom}{0.4063304169\ht\trajectorynatural}
\setlength{\trajectorytop}{0.5856922285\ht\trajectorynatural}
\hbox{\includegraphics[trim={0pt \trajectorybottom{} 0pt \trajectorytop},clip,width=\linewidth]{figures/figure-07.png}}
\vspace{.35pt}
\setlength{\trajectorybottom}{0.3888317036\ht\trajectorynatural}
\setlength{\trajectorytop}{0.6013896037\ht\trajectorynatural}
\hbox{\includegraphics[trim={0pt \trajectorybottom{} 0pt \trajectorytop},clip,width=\linewidth]{figures/figure-07.png}}
\vspace{.35pt}
\setlength{\trajectorybottom}{0.3692743181\ht\trajectorynatural}
\setlength{\trajectorytop}{0.6196603191\ht\trajectorynatural}
\hbox{\includegraphics[trim={0pt \trajectorybottom{} 0pt \trajectorytop},clip,width=\linewidth]{figures/figure-07.png}}
\vspace{.35pt}
\setlength{\trajectorybottom}{0.2300566135\ht\trajectorynatural}
\setlength{\trajectorytop}{0.6335563562\ht\trajectorynatural}
\hbox{\includegraphics[trim={0pt \trajectorybottom{} 0pt \trajectorytop},clip,width=\linewidth]{figures/figure-07.png}}
\vspace{.35pt}
\setlength{\trajectorybottom}{0.2169325785\ht\trajectorynatural}
\setlength{\trajectorytop}{0.7750900669\ht\trajectorynatural}
\hbox{\includegraphics[trim={0pt \trajectorybottom{} 0pt \trajectorytop},clip,width=\linewidth]{figures/figure-07.png}}
\vspace{.35pt}
\setlength{\trajectorybottom}{0.1994338652\ht\trajectorynatural}
\setlength{\trajectorytop}{0.7907874421\ht\trajectorynatural}
\hbox{\includegraphics[trim={0pt \trajectorybottom{} 0pt \trajectorytop},clip,width=\linewidth]{figures/figure-07.png}}
\vspace{.35pt}
\setlength{\trajectorybottom}{0.1798764797\ht\trajectorynatural}
\setlength{\trajectorytop}{0.8090581575\ht\trajectorynatural}
\hbox{\includegraphics[trim={0pt \trajectorybottom{} 0pt \trajectorytop},clip,width=\linewidth]{figures/figure-07.png}}
\vspace{.35pt}
\setlength{\trajectorybottom}{0.0406587751\ht\trajectorynatural}
\setlength{\trajectorytop}{0.8257848688\ht\trajectorynatural}
\hbox{\includegraphics[trim={0pt \trajectorybottom{} 0pt \trajectorytop},clip,width=\linewidth]{figures/figure-07.png}}
\vspace{.35pt}
\setlength{\trajectorybottom}{0.0275347401\ht\trajectorynatural}
\setlength{\trajectorytop}{0.9644879053\ht\trajectorynatural}
\hbox{\includegraphics[trim={0pt \trajectorybottom{} 0pt \trajectorytop},clip,width=\linewidth]{figures/figure-07.png}}
\vspace{.35pt}
\setlength{\trajectorybottom}{0.0100360268\ht\trajectorynatural}
\setlength{\trajectorytop}{0.9801852805\ht\trajectorynatural}
\hbox{\includegraphics[trim={0pt \trajectorybottom{} 0pt \trajectorytop},clip,width=\linewidth]{figures/figure-07.png}}
\end{minipage}
\end{minipage}
\begin{reportfigure}
\renewcommand{\thefigure}{D9}
\caption{Recorded 3D feature observations and corresponding simulated trajectories. Only valid observations are plotted; the shuttlecock has three valid depth observations.}
\label{fig:D9}
\phantomsection\label{html:raw001-figure-7}

\end{reportfigure}

\phantomsection\label{html:supplementary-scoring}

\onecolumn
\vspace{-9pt}
\section{Supplementary Real2Sim evaluation results}
\label{sec:E}
This appendix reports stage-level success rates and supplementary task Scores for Real2Sim evaluation. Complete benchmark results are reported in \hyperref[sec:B]{Appendix~\ref*{sec:B}}. Table values are displayed to two decimal places; calculations use full precision.
\begin{reportfigure}
\centering
\renewcommand{\thefigure}{E1}
{\fontsize{8.2}{10.1}\selectfont
\setlength{\tabcolsep}{3pt}
\renewcommand{\arraystretch}{1.05}
\begin{tabular}{>{\raggedright\arraybackslash}p{\dimexpr 0.1667\linewidth-2\tabcolsep\relax}>{\raggedright\arraybackslash}p{\dimexpr 0.1667\linewidth-2\tabcolsep\relax}>{\raggedright\arraybackslash}p{\dimexpr 0.1667\linewidth-2\tabcolsep\relax}>{\raggedright\arraybackslash}p{\dimexpr 0.1667\linewidth-2\tabcolsep\relax}>{\raggedright\arraybackslash}p{\dimexpr 0.1667\linewidth-2\tabcolsep\relax}>{\raggedright\arraybackslash}p{\dimexpr 0.1667\linewidth-2\tabcolsep\relax}}
\toprule
\rowcolor{tablehead}\textbf{Source task} & \textbf{Capability} & \textbf{Real follow} & \textbf{Real grasp} & \textbf{Sim follow} & \textbf{Sim grasp} \\
\midrule
Select by Color & Color & 0.87 & 0.84 & 0.84 & 0.79 \\
\rowcolor{tablerow}Grasp Target Object & Semantics & 1.00 & 0.66 & 0.83 & 0.68 \\
Recognize Size & Size & 0.47 & 0.37 & 0.52 & 0.42 \\
\rowcolor{tablerow}Follow Logic \allowbreak{}(OR) & Logic & 0.84 & 0.63 & 0.79 & 0.66 \\
Apply Common Sense & Common sense & 0.54 & 0.44 & 0.50 & 0.42 \\
\bottomrule
\end{tabular}
}

\caption{G2 stage-level instruction-following and grasping success rates across five capability categories. Each cell shows the final reported value. These stage rates are distinct from the task-level benchmark Score in \hyperref[tab:B1]{Table~B1}. The source provides neither evaluation counts nor exact phase definitions.}
\label{fig:E1}
\phantomsection\label{html:raw002-figure-2}
\end{reportfigure}
\begin{reportfigure}
\centering
\renewcommand{\thefigure}{E2}
{\fontsize{8.2}{10.1}\selectfont
\setlength{\tabcolsep}{3pt}
\renewcommand{\arraystretch}{1.05}
\begin{tabular}{>{\raggedright\arraybackslash}p{\dimexpr 0.2500\linewidth-2\tabcolsep\relax}>{\raggedright\arraybackslash}p{\dimexpr 0.2500\linewidth-2\tabcolsep\relax}>{\raggedright\arraybackslash}p{\dimexpr 0.2500\linewidth-2\tabcolsep\relax}>{\raggedright\arraybackslash}p{\dimexpr 0.2500\linewidth-2\tabcolsep\relax}}
\toprule
\rowcolor{tablehead}\textbf{Task} & \textbf{Capability} & \textbf{Real Score} & \textbf{Sim Score} \\
\midrule
Select by Color & Color & 0.91 & 0.90 \\
\rowcolor{tablerow}Recognize Size & Size & 0.77 & 0.66 \\
Grasp Target Object & Semantics & 0.85 & 0.79 \\
\rowcolor{tablerow}Stack Blocks & Precision & 0.24 & 0.24 \\
\bottomrule
\end{tabular}
}

\caption{Arx X5 final task-level Scores for four representative capabilities,\allowbreak{} using the final results table's after-identification columns. Full before/after comparisons for all seven tasks appear in \hyperref[tab:B2]{Table~B2}. Score retains its task-specific definition and is distinct from stage-level success rate.}
\label{fig:E2}
\phantomsection\label{html:raw002-figure-3}
\end{reportfigure}

\end{minipage}\par

\end{document}